\documentclass{article} 
\usepackage{iclr2027_conference,times}

\usepackage{amsmath,amsfonts,bm}

\def\eqref#1{equation~\ref{#1}}
\def\Eqref#1{Equation~\ref{#1}}

\def\1{\bm{1}}

\DeclareMathAlphabet{\mathsfit}{\encodingdefault}{\sfdefault}{m}{sl}
\SetMathAlphabet{\mathsfit}{bold}{\encodingdefault}{\sfdefault}{bx}{n}

\newcommand{\R}{\mathbb{R}}

\usepackage{hyperref}
\usepackage{url}
\usepackage{amssymb}
\usepackage[T1]{fontenc}
\usepackage{xcolor}
\usepackage{booktabs}
\usepackage{multirow}
\usepackage{graphicx}
\usepackage{algorithm}
\usepackage{algpseudocode}
\usepackage{amsthm}
\usepackage{enumerate}
\usepackage{array}
\usepackage{enumitem}
\usepackage{wrapfig}
\usepackage{float}
\usepackage{comment}
\usepackage{array}
\usepackage{url}
\usepackage{caption}
\usepackage[table]{xcolor}
\newcolumntype{?}{!{\vrule width 0.5pt}}

\newtheorem{proposition}{Proposition}
\newtheorem{lemma}{Lemma}
\newtheorem{theorem}{Theorem}
\newtheorem{remark}{Remark}
\newtheorem{corollary}{Corollary}
\newtheorem{example}{Example}
\DeclareMathOperator{\MST}{MST}
\DeclareMathOperator{\pers}{Pers}
\DeclareMathOperator{\TotPers}{TotPers}
\newcommand{\cL}{\mathcal{L}}
\newcommand{\md}{\mathrm{MD}}
\newcommand{\uhat}{\hat u}
\newcommand{\Sb}{S^{b}}
\newcommand{\Sw}{S^{w}}

\newcommand{\ourmethod}{\textsc{TopoSteer}}

\definecolor{educolor}{RGB}{143,20,180} 

\definecolor{fixcolor}{RGB}{0,122,92} 

\usepackage{hyperref}
\usepackage{url}

\title{TopoSteer: Topological Layer Selection for Activation Steering in LLMs}

\author{%
  Tuc Nguyen\textsuperscript{1} \quad
  Marcos Guti\'errez-del-Pozo\textsuperscript{2} \quad
  Eduardo Paluzo-Hidalgo\textsuperscript{3,4} \\
  \textbf{Yifan Hu}\textsuperscript{5} \quad
  \textbf{Adaku Uchendu}\textsuperscript{6} \quad
  \textbf{Thai Le}\textsuperscript{1} \\[0.5em]
  \normalfont\normalsize
  \textsuperscript{1}Indiana University, USA \\
  \textsuperscript{2}Institute of Mathematics Antonio de Castro Brzezicki, University of Sevilla, Spain \\
  \textsuperscript{3}Department of Applied Mathematics I, University of Sevilla, Spain \\
  \textsuperscript{4}Graduate School of Mathematical Sciences, The University of Tokyo, Japan \\
  \textsuperscript{5}Northeastern University, USA \\
  \textsuperscript{6}MIT Lincoln Laboratory, USA \\[0.3em]
}

\iclrfinalcopy 
\begin{document}
\maketitle
\begin{abstract}
    Activation steering controls large language models at inference time by adding a vector to their internal representations. Most work asks how to build that vector, and far less attention has gone to where to add it, even though steering at a poorly chosen layer can be worse than not steering at all. Prior selectors choose a layer without ever steering the model to test it: they score how separable the target behavior is at that layer, and how consistently the individual examples point along the steering direction. We argue that what matters is not the separability of the two activation clouds, nor the alignment of individual displacements with their mean, but the extent of the cloud of per-example displacements that mean-difference steering averages. We propose a topology-based layer selection method, called \ourmethod{}, which scores a layer by the total $H_0$ persistence of its displacement cloud, computed as the length of the cloud's Euclidean minimum spanning tree.  We show that the score focuses on the displacement cloud's relative geometry, while class separation affects it only through standardization. We also provide a theoretical explanation for our method by proving that, as example-level variation increases, \ourmethod{} and discriminability move in opposite directions, so the two criteria can rank layers in opposite ways. Across multiple language models and steering tasks, \ourmethod{} provides a principled alternative to discriminability- and separability-based layer selection and yields more effective steering without requiring held-out validation.
\end{abstract}

\section{Introduction}
Large language models (LLMs) perform strongly across diverse tasks and are now deployed as commercial chat products whose real-world usage is long-form, multilingual, and highly varied across platforms~\citep{yan2025sharechat}, so deployment increasingly demands controllable, reliable, and aligned behavior at inference time. While fine-tuning and reinforcement learning can improve alignment, they are costly, task-specific, and irreversible. Activation steering has therefore emerged as a lightweight alternative that controls LLMs behavior at inference time by intervening directly on internal representations, without modifying model parameters~\citep{azizi2025activation}. The dominant recipe is mean-difference (MD) steering, which computes a steering vector from contrastive examples and injects it into selected layers, and which has shown effectiveness for AI alignment, model reasoning, and safety mitigation~\citep{rimsky-etal-2024-steering, nguyen2026atlas, nguyen2026beyond}. However, MD's steering effectiveness depends strongly on where the intervention is applied~\citep{rimsky2024steering, sun2025layernavigator}.

\begin{figure}[t]
    \centering
    \includegraphics[width=0.95\linewidth]{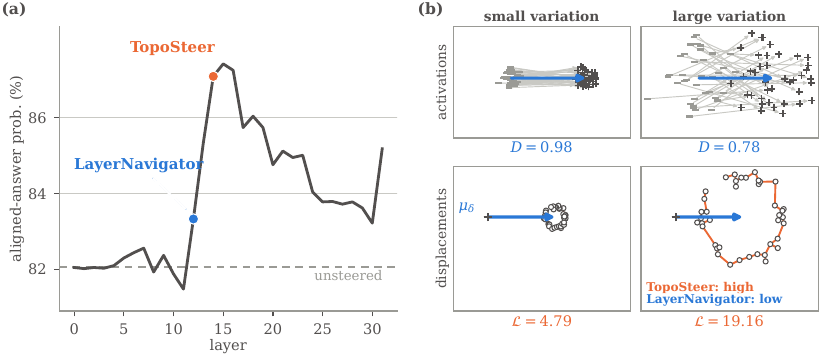}
    \caption{\textbf{Layer choice decides the outcome, and LayerNavigator and \ourmethod{} choose differently.}
        (a) Single-layer steering performance against depth (conscientiousness, Llama-8B) with the
        unsteered baseline, and the layer each method ranks highest ($K{=}1$).
        (b) Why they differ: two layers from the family of Theorem~\ref{thm:decoupling}, with the same class means and the same steering vector, differing only in the scale of example-level variation. Top: activations, each contrastive pair joined by its displacement $\delta_l^i$ (grey arrows), with the mean difference $\mu_\delta$ in blue. With large variation, the two classes overlap, so discriminability $D$ drops, while the displacements keep a common direction but vary widely in length. Bottom: the same displacements drawn from a common origin form the displacement cloud $\Delta_l$; connectivity length $\cL$ is the length of its spanning tree (orange), which by Corollary~\ref{cor:inv} does not see the offset $\mu_\delta$ at all. The large-variation layer is therefore scored higher by \ourmethod{} and lower by LayerNavigator, whose full score $D+C$ also drops. All values shown are computed from the data drawn; the construction is the one of Appendix~\ref{app:toy}.}
    \label{fig:teaser}
\end{figure}

This strong layer dependence makes validation-free layer selection an important practical problem: given the contrastive examples used to construct the steering vector, can we predict effective intervention layers without exhaustively evaluating steering at every candidate layer? Figure~\ref{fig:teaser}(a) illustrates the challenge using MD as an example. The effectiveness of MD steering varies substantially across injection layers: on Llama-8B-Instruct, steering at early layers leaves performance near the unsteered baseline, injecting at layer $15$ improves it, and at other layers steering reduces performance below the baseline. Exhaustively selecting $K$ of $L$ layers requires evaluating $\binom{L}{K}$ possible layer subsets on held-out data, making such a brute-force search impractical. Recent work therefore scores layers using only the extraction data. LayerNavigator~\citep{sun2025layernavigator} ranks layers by $S_l = D_l + C_l$, combining a discriminability term $D_l$, defined as a Fisher-type ratio of between-class to total scatter along the steering direction, with $C_l$, which captures the alignment of individual displacements with the steering direction.

The two terms capture complementary but incomplete aspects of the geometry relevant to MD steering. MD does not use the two activation clouds separately. At layer $l$, let $a_l(x) \in \mathbb{R}^d$ denote the residual-stream activation of input $x$. We define the per-example displacement as $\delta_l^i := a_l(x_i^+) - a_l(x_i^-)$ (in the standardized representation of Sec.~\ref{sec:setup}), and the displacement cloud as $\Delta_l := \{\delta_l^i\}_{i=1}^N$. Because MD steering forms its intervention by averaging these per-example displacements, whether a single vector represents the available signal is a property of the displacement cloud $\Delta_l$. The discriminability term $D_l$ never looks at that cloud: it is built from class means and within-class scatter, and describes how {separable} the two classes are rather than how the individual examples move. The consistency term does look at it, but only at directions; as an average of cosine similarities, it is unchanged if each displacement is rescaled independently. It therefore measures how tightly the displacements concentrate about the mean direction, and discards their extent: whether the cloud is a tight clump around a single vector, or a broad set of example-specific ones. This extent, overlooked by both terms, is what we propose to score in this paper.

To this end, we propose \ourmethod{}, which scores each layer $l$ by the total degree-zero persistence of its displacement cloud
\begin{equation}
    s_l \;:=\; \TotPers_0(\Delta_l) \;=\; \sum_{e\in\MST(\Delta_l)} w(e) \;=:\; \cL(\Delta_l),
    \label{eq:teaser_score}
\end{equation}
evaluated in the standardized representation of Sec.~\ref{sec:setup}. The second equality is the classical identification of finite $H_0$ death scales with Euclidean minimum-spanning-tree edge weights~\citep{7efdf403-c7cc-3f04-98a8-023222e4fc84}, which we use as the computational form of the score. The score requires no held-out steering runs, no auxiliary training, and no forward passes beyond those already performed to extract the MD's steering vectors.

Our analysis shows that \ourmethod{} measures a quantity neither existing term captures. Against discriminability, the relation is an exact reversal: along the axis that scales example-level variation at fixed class means, the two scores move in opposite directions, so the criteria can rank layers oppositely (Theorem~\ref{thm:decoupling}). Against consistency, it is complementarity: connectivity length is invariant to the mean displacement and $1$-homogeneous, while consistency is an average of cosines and therefore reads only directions, discarding the extent that $\cL$ measures. On real activations, we further identify an \emph{answer-polarity} axis induced by the contrastive format rather than the target behavior. This axis splits the displacement cloud into two lobes without contributing between-class information, yet it affects the two criteria through different geometric mechanisms: its effect on discriminability depends on its alignment with the steering direction, whereas its effect on $\cL$ is governed by the spanning-tree bridge and the merging of the two lobes. We analyze these effects in Sec.~\ref{sec:polarity}.

Our main contributions are as follows. We identify the displacement cloud as the steering-relevant object for MD layer selection and propose \ourmethod{}, a validation-free topological selector based on its total $H_0$ persistence. We theoretically characterize how its connectivity differs from discriminability and directional consistency, and empirically show that it improves layer selection across multiple models and steering tasks without held-out search.

\section{Related Work}
\paragraph{Activation Steering and Layer Selection.}
Activation steering computes a direction associated with a concept or behavior and adds it to residual-stream representations at inference time~\citep{zou2023representation, azizi2025activation}. A common approach constructs this direction from differences between paired contrastive activations, including contrastive activation addition~\citep{rimsky2024steering}, refusal-direction interventions~\citep{arditi2024refusal}, and related inference-time interventions~\citep{li2023inference}; the model parameters remain unchanged and the intervention can be applied dynamically~\citep{nguyen2026atlas, nguyen2026beyond}. Our contribution is orthogonal to the choice of direction: we take the extractor as given and ask where to apply it. Existing selection strategies rely on probing accuracy, linear separability, activation differences, or heuristics such as intervening in the middle third of the network.  LayerNavigator~\citep{sun2025layernavigator} is the closest point of comparison, combining a Fisher-type discriminability score $D$, computed from class means and within-class scatter, with a directional-consistency term $C$, the mean cosine between the steering vector and the individual displacements. \ourmethod{} scores the extent of those displacements, which neither term captures. \ourmethod{} evaluates the geometry of the per-example displacement vectors that MD steering actually averages, rather than the separability of the two activation classes.

\paragraph{Persistent Homology of Neural Representations.} 
Persistent homology gives a multiscale description of a point cloud~\citep{edelsbrunner2008persistent}, and has been used to study neural representations and their geometry~\citep{rieck2019neural, pmlr-v108-gabrielsson20a, JMLR:v21:20-345, 10.1145/3820356.3820362}. \citet{fay2026the} analyze LLMs' activation spaces and show that adversarial inputs induce topological compression, in which diverse small-scale features collapse into fewer dominant large-scale structures. Related work on transformer geometry has also documented systematic layerwise variation in intrinsic dimension, anisotropy, and representational expansion and compression with depth~\citep{gardinazzi2024persistent}.
Recent work has moved toward adaptive steering, selecting intervention layers or steering strengths based on the input. W2S~\citep{gadgil2026steer} learns a predictor for the
input-specific optimal intervention layer, using candidate-layer evaluations to construct the supervision signal. Dynamically Scaled Activation Steering~\citep{ferrando2025dynamically}
instead learns input- and layer-dependent scaling factors that determine when and how strongly to steer. \citet{nelwan2026deployable} extend adaptive selection to multiple layers, training a prompt-based ranker from per-instance layer-selection targets and an adaptive gate for the number of layers. These approaches require either candidate-layer evaluation or additional training to learn an input-dependent selection policy, whereas \ourmethod{} selects layers directly from the contrastive extraction data, without held-out steering evaluations or selector training. We therefore use persistent homology not as a classifier or training objective, but as a validation-free diagnostic of the geometry of paired displacements for selecting intervention layers.

\section{\ourmethod{}: Topological Layer Selection}
\label{sec:toposteer}
This section defines the method: the object it scores (Sec.~\ref{sec:setup}), the score and its
exact invariances (Sec.~\ref{sec:score}), and the selection procedure, its cost, the diagnostic that guards its main failure mode, and the two regimes in which the selected layers are steered
(Sec.~\ref{sec:toposteer_algorithm}). Why this criterion differs from class-separability selectors, and what the displacement cloud looks like in a trained model, are deferred to
Sec.~\ref{sec:information_encoding_geometry}, after the empirical comparison.

\subsection{The Displacement Cloud}
\label{sec:setup}

Let $l$ be a fixed layer of width $d$ and let $\{(x_i^+,x_i^-)\}_{i=1}^{N}$ be paired contrastive inputs. Write $a_l(x_i^c)\in\mathbb{R}^d$, $c\in\{+,-\}$, for the residual-stream activation at the final prompt token. MD vectors are computed from the raw activations as follows: 
$v_l^{\mathrm{raw}} := \frac{1}{N}\sum_{i=1}^{N}\bigl[a_l(x_i^+)-a_l(x_i^-)\bigr]$
and injects $\alpha v_l^{\mathrm{raw}}$ at the chosen layers. For selection, we need displacement geometries that are comparable across layers rather than dominated by coordinate-wise scale, we standardize the representation with 
\begin{equation}
    \tilde a_l(x_i^c)
    = \frac{a_l(x_i^c)-\mu_l}{\sigma_l},
    \label{eq:zscore}
\end{equation}
where $\mu_l,\sigma_l\in\mathbb{R}^d$ are the coordinate-wise mean and standard deviation over all $2N$ activations at layer $l$. The \emph{displacement cloud} is
\begin{equation}
    \delta_l^i
    :=
    \tilde a_l(x_i^+)-\tilde a_l(x_i^-),
    \qquad
    \Delta_l
    :=
    \{\delta_l^i\}_{i=1}^{N},
    \label{eq:displacement_cloud}
\end{equation}
with mean $\bar{\delta}_l = \tilde\mu_l^+-\tilde\mu_l^-=:\mu_{\delta,l}$ is
the mean-difference direction in the analyzed space. Writing
$\varepsilon_l(x_{i}^c) := \tilde a_l(x_i^c)-\tilde\mu_l^c$ for the within-class residuals and
$\xi_l^i := \varepsilon_l(x_{i}^+)-\varepsilon_l(x_{i}^-)$ for the residual displacements, each displacement decomposes as
\begin{equation}
    \delta_l^i
    =
    \mu_{\delta,l}+\xi_l^i,
    \qquad
    \sum_{i=1}^{N}\xi_l^i=0,
    \qquad
    \Xi_l:=\{\xi_l^i\}_{i=1}^{N}.
    \label{eq:displacement_decomposition}
\end{equation}
The contrastive construction gives $\Delta_l$ a second, structural decomposition that the score turns out to be insensitive to; we defer it to Sec.~\ref{sec:polarity}, where it is what separates the two criteria.

\subsection{Scoring a Layer}
\label{sec:score}
We score a layer by the connectivity of its displacement cloud, measured by the total degree-zero persistence $\TotPers(X)$ of a finite point cloud $X$, the sum of the finite death scales of its Vietoris--Rips $H_0$ barcode (Appendix~\ref{sec:tda}). Under the Euclidean filtration, these death scales are the edge weights of the Euclidean minimum spanning tree~\citep{7efdf403-c7cc-3f04-98a8-023222e4fc84}, which we use as the computational form of the score,
\begin{equation}
    \TotPers(X)
    =
    \sum_{e\in\MST(X)} w(e)
    =:
    \cL(X),
    \label{eq:mst_identity_main}
\end{equation}
restated as Proposition~\ref{prop:mst} in Appendix~\ref{app:mst}.
It can be computed from the pairwise distance matrix in $O(n^2d)$ time, and admits a direct interpretation as the connectivity length of the displacement cloud.
\ourmethod{} assigns layer $l$ the score:
\begin{equation}
    s_l
    :=
    \TotPers\left(\Delta_l\right)
    =
    \cL(\Delta_l).
    \label{eq:toposteer_layer_score}
\end{equation}

Two exact properties of $\cL$ determine what the score can and cannot see
(Proposition~\ref{prop:inv}). It is translation invariant, and it is $1$-homogeneous:
$\cL(X+u)=\cL(X)$ for any $u\in\mathbb{R}^d$ and $\cL(cX)=c\,\cL(X)$ for $c>0$. Applying translation invariance to the decomposition~\eqref{eq:displacement_decomposition} gives the property we rely on throughout.

\begin{corollary}[Common-mean invariance]
    \label{cor:inv}
    Under the Euclidean metric,
    $\cL(\Delta_l) = \cL(\Delta_l-\mu_{\delta,l}) = \cL(\Xi_l)$.
    The score is therefore invariant to the common mean displacement \emph{of the cloud it is given}, and
        depends only on the relative arrangement of the example-level residual displacements. It does not
        follow that the pipeline is insensitive to between-class separation: the standardization
        \eqref{eq:zscore} is estimated from the same activations, and separation enters through it.
        Proposition~\ref{prop:sepscale} makes the dependence exact.
\end{corollary}

\subsection{Layer Selection and Steering Protocols}
\label{sec:toposteer_algorithm}

\paragraph{Layer Selection.} Given a layer budget $K$, \ourmethod{} selects the $K$ highest-scoring layers, $S_K$ per task. At inference time, each selected layer receives its corresponding raw MD vector,
\begin{equation}
    a_{l,t}^{\prime}
    = a_{l,t}
    +
    \alpha v_l^{\mathrm{raw}},
    \qquad
    l\in S_K,
    \quad
    t\in\mathcal T,
    \label{eq:multilayer_steering}
\end{equation}
where $\alpha$ is the steering strength and $\mathcal T$ denotes the intervention token positions. The standardized displacement cloud $\Delta_l$ is used only for layer scoring, while the raw MD vector $v_l^{\mathrm{raw}}$ is used for intervention; \ourmethod{} therefore changes only the layer-selection rule. Algorithm~\ref{alg:toposteer} summarizes the procedure. The same cached activations produce both the steering vectors and the layer scores, so \ourmethod{} requires no held-out steering evaluations, no
auxiliary selector training, and no forward passes beyond those already needed for MD extraction; its
additional cost is an offline geometric computation over the extraction data, $O(N^2d)$ per layer
(Appendix~\ref{sec:toposteer_complexity}). A large connectivity length can arise from edge weights spread across the cloud or from a few long edges bridging separated modes or isolated outliers; in the second case $\cL$ measures separation between modes rather than the arrangement within them. Algorithm~\ref{alg:toposteer} therefore also returns the \emph{bridge share}
\begin{equation}
    b_l := \max_{e \in \MST(\Delta_l)} w(e)\,/\,\cL(\Delta_l),
    \label{eq:bridge}
\end{equation}
the fraction of the score carried by its longest edge, which flags whether the score is bulk-driven or carried by a single connection (Appendix~\ref{app:toy}).

\paragraph{Simultaneous and Sequential Steering.} 
\begin{wrapfigure}[16]{r}{0.55\linewidth}
\vspace{-25pt}
\begin{minipage}{\linewidth}
\begin{algorithm}[H]
\caption{\ourmethod{} Layer Selection}
\label{alg:toposteer}
\footnotesize
\begin{algorithmic}[1]
\Require $\mathcal{P}=\{(x_i^+,x_i^-)\}_{i=1}^{N}$; model $M$ with $L$ layers; steering budget $K$
\Ensure selected layers $S_K$; $\{v_l^{\mathrm{raw}}\}_{l\in S_K}$; score $\{s_l\}_{l=1}^{L}$
\State Run $M$ on $\mathcal{P}$ and cache $\{a_l(x_i^c)\}_{l\in[L],\,i\in[N],\,c\in\{+,-\}}$
\For{$l=1,\ldots,L$}
\State $v_l^{\mathrm{raw}} \gets \frac{1}{N}\sum_{i=1}^{N}\bigl[a_l(x_i^+)-a_l(x_i^-)\bigr]$
\State $(\mu_l,\sigma_l) \gets$
standardize $\{a_l(x_i^c)\}_{i,c}$ by~\eqref{eq:zscore}
\State Construct $\Delta_l$ by~\eqref{eq:displacement_cloud}
\State $T_l \gets \MST(\Delta_l)$
\State $s_l \gets \sum_{e\in T_l} w(e)$
\State $b_l \gets \max_{e\in T_l} w(e)\,/\,s_l$
\EndFor
\State $S_K \gets \operatorname{TopK}_{l\in[L]} s_l$
\State \Return $S_K$, $\{v_l^{\mathrm{raw}}\}_{l\in S_K}$, $\{s_l\}_{l=1}^{L}$, $\{b_l\}_{l\in S_K}$
\end{algorithmic}
\end{algorithm}
\end{minipage}
\end{wrapfigure}

We consider two intervention protocols. In the \emph{simultaneous} regime, all $K$ selected layers are intervened at once (\eqref{eq:multilayer_steering}) using a single extraction pass. It is the protocol used by prior validation-free selectors, and it assumes that scores computed on the unsteered model still describe the model once the first injection is in place, which may not be true. We therefore also evaluate a \emph{sequential} regime, applying the $K$ injections one at a time. At round $k$ we re-run the extraction pass through the model already carrying rounds $1,\ldots,k-1$, recompute the MD vectors and the layer scores from those steered activations, and inject at the new argmax. 
Each round takes the argmax over the layers not yet selected, so the output is an ordered tuple $(l^{(1)},\ldots,l^{(K)})$ of distinct layers that need not equal the simultaneous top-$K$; at $K=1$ the two regimes coincide. The regime costs $K$ extraction passes instead of one and is a protocol rather than a selector.

\section{Experiments}
\label{sec:experiments_alignment}
\paragraph{Setup.}
We evaluate \ourmethod{} on six binary persona-steering tasks from the Anthropic Persona Dataset~\citep{perez-etal-2023-discovering}. We use approximately $700$ paired examples to extract the MD vectors and compute layer scores, and $100$ held-out examples for evaluation. Experiments cover five instruction-tuned models: \textsc{Llama-3-3B-Instruct} (Llama-3B), \textsc{Llama-3-8B-Instruct} (Llama-8B)~\citep{grattafiori2024llama}, \textsc{Qwen2.5-3B-Instruct} (Qwen-3B), \textsc{Qwen2.5-7B-Instruct} (Qwen-7B), and \textsc{Qwen2.5-32B-Instruct} (Qwen-32B)~\citep{team2024qwen2}. Activations are extracted at the final prompt token. \ourmethod{} scores the standardized displacement clouds of Eq.~\ref{eq:displacement_cloud}, while the raw MD vectors are used for intervention. Unless otherwise stated, $\alpha=1.0$, $K\in\{1,2,3\}$, both intervention regimes in Sec.~\ref{sec:toposteer_algorithm} are considered, and all filtrations use Euclidean distance. 

\paragraph{Baselines.} We compare against LayerNavigator~\citep{sun2025layernavigator}, layer selection by activation discriminativeness, and random selection. We also report a variant of \ourmethod{} in which each standardized displacement $\Delta_l^{(i)}$ is normalized to unit length,
$\tilde{\Delta}_l^{(i)} = \Delta_l^{(i)} / \lVert \Delta_l^{(i)} \rVert_2$, before computing the layer score; the score thus depends only on the directions of the displacements, not on their magnitudes.
We refer to Appendix~\ref{app:expdetails} for additional detail.

\paragraph{Main Results.}

\definecolor{highlightgreen}{RGB}{220,239,225}
\begin{table*}[t]
    \centering
    \setlength{\aboverulesep}{0pt}
    \setlength{\belowrulesep}{0pt}
    \setlength{\extrarowheight}{.75ex}
    \renewcommand{\tabcolsep}{2pt}
    \footnotesize
    \resizebox{\textwidth}{!}{
    \begin{tabular}{cc c ccccc ?ccccc ?ccccc}
        \toprule
        & & \multirow{2}{*}{\textbf{Base}} &
        \multicolumn{5}{c?}{$\mathbf{K=1}$} &
        \multicolumn{5}{c?}{$\mathbf{K=2}$} &
        \multicolumn{5}{c}{$\mathbf{K=3}$}
        \\
        \cmidrule(lr){4-8}
        \cmidrule(lr){9-13}
        \cmidrule(lr){14-18}

        & & &
        Disc & Rand & LN & Ours & Ours$_{\ell_2}$ &
        Disc & Rand & LN & Ours & Ours$_{\ell_2}$ &
        Disc & Rand & LN & Ours & Ours$_{\ell_2}$
        \\
        \midrule

        \multirow{5}{*}{\rotatebox[origin=c]{90}{\textbf{Simultaneous}}}
        & Llama-3B & 66.73
        & \underline{72.79} & 70.46 & \textbf{73.79} & \underline{72.79} & \underline{72.79}
        & \underline{79.71} & 75.40 & \textbf{81.38} & \textbf{81.38} & \textbf{81.38}
        & \underline{77.03} & 73.44 & 75.12 & \cellcolor{highlightgreen}{\textbf{80.49}} & 77.40
        \\
        
        & Llama-8B & 77.62
        & 79.73 & 80.04 & 81.26 & \cellcolor{highlightgreen}{\textbf{83.44}} & \underline{82.70}
        & 81.70 & 83.83 & \underline{84.87} & 84.26 & \cellcolor{highlightgreen}{\textbf{87.09}}
        & 84.11 & 85.79 & \textbf{86.43} & \underline{85.91} & 84.73
        \\

        & Qwen-3B & 64.26
        & 61.12 & 63.88 & 64.30 & \underline{64.86} & \cellcolor{highlightgreen}{\textbf{71.77}}
        & 59.63 & 61.96 & \underline{64.16} & 60.90 & \cellcolor{highlightgreen}{\textbf{77.63}}
        & 56.41 & 61.55 & \underline{66.20} & 61.07 & \cellcolor{highlightgreen}{\textbf{79.71}}
        \\
        
        & Qwen-7B & 85.24
        & \textbf{85.91} & 85.58 & 85.60 & \underline{85.84} & 85.72
        & 85.85 & \underline{85.86} & \textbf{86.10} & 85.85 & 85.62
        & \underline{85.63} & 85.19 & 84.97 & \textbf{85.88} & 85.40
        \\

        & Qwen-32B & 82.68
        & 82.78 & 83.10 & \textbf{83.42} & \underline{83.24} & 83.23
        & 82.51 & 82.97 & 83.08 & \textbf{83.91} & \underline{83.69}
        & 81.92 & 82.94 & 81.75 & \textbf{83.66} & \underline{83.64}
        \\
        
        \midrule

        \multirow{5}{*}{\rotatebox[origin=c]{90}{\textbf{Sequential}}}
        & Llama-3B & 66.73
        & \underline{72.79} & 70.46 & \textbf{73.79} & \underline{72.79} & \underline{72.79}
        & 78.74 & 73.20 & \textbf{80.61} & \underline{80.41} & \underline{80.41}
        & \underline{77.12} & 76.01 & 75.17 & \textbf{78.13} & 73.06
        \\

        & Llama-8B & 77.62
        & 79.73 & 80.04 & 81.26 & \cellcolor{highlightgreen}{\textbf{83.44}} & \underline{82.70}
        & 80.10 & 81.49 & \underline{84.28} & 84.11 & \cellcolor{highlightgreen}{\textbf{86.49}}
        & 80.83 & 83.10 & 85.24 & \underline{85.58} & \textbf{85.65}
        \\

        & Qwen-3B & 64.26
        & 61.12 & 63.88 & 64.30 & \underline{64.86} & \cellcolor{highlightgreen}{\textbf{71.77}}
        & 57.12 & 63.48 & \underline{64.88} & 61.63 & \cellcolor{highlightgreen}{\textbf{76.72}}
        & 56.86 & 63.43 & \underline{66.48} & 61.22 & \cellcolor{highlightgreen}{\textbf{79.19}}
        \\

        & Qwen-7B & 85.24
        & \textbf{85.91} & 85.58 & 85.60 & \underline{85.84} & 85.72
        & \underline{85.86} & 85.80 & \textbf{86.19} & 85.78 & 85.51
        & 85.75 & \textbf{86.82} & \underline{86.30} & 85.66 & 85.38
        \\

        & Qwen-32B & 82.68
        & 82.78 & 83.10 & \textbf{83.42} & \underline{83.24} & 83.23
        & 82.49 & 82.86 & 82.95 & \underline{83.53} & \textbf{83.54}
        & 81.96 & 82.73 & 82.87 & \underline{83.33} & \textbf{83.50}
        \\
        \bottomrule
    \end{tabular}
    }
    \caption{
        Aligned-answer probability (\%, $\uparrow$), averaged over the six persona-steering tasks. \textbf{Base} is the unsteered model. For each layer budget $K\in\{1,2,3\}$ we compare discriminability-only (Disc), random selection, LayerNavigator (LN), and \ourmethod{} under
        two scoring clouds: the standardized displacement cloud (Ours) and its unit-sphere projection (Ours$_{\ell_2}$). \textbf{Bold} marks the best selector per setting and \underline{underline} the second best; Base is a reference and is excluded from both. {\colorbox{highlightgreen}{Green cells}} mark \ourmethod{} results exceeding the strongest non-\ourmethod{} selector by at least $2.0$ percentage points. The upper block reports simultaneous steering; the lower block reports greedy sequential selection with re-extraction, where each step adds a layer not chosen in an earlier step.
    }
    \label{tab:main}
\end{table*}

Table~\ref{tab:main} summarizes alignment performance averaged over the six persona-steering tasks. In the simultaneous setting, \ourmethod{} outperforms LayerNavigator in $10$ of $15$ model--budget comparisons, ties in one, and is lower in four. The largest gains occur on Qwen-3B, where
\ourmethod{} improves alignment from $64.30\%$ to $71.77\%$, from $64.16\%$ to $77.63\%$, and from $66.21\%$ to $79.71\%$ for $K=1,2,3$, respectively. On Llama-8B, it also improves alignment at $K=1$ and $K=2$ by $2.18$ and $1.22$ percentage points, respectively. The gains are smaller and more mixed on Qwen-7B and Qwen-32B, showing that the benefit of displacement-based selection
depends on both the model and layer budget. Table~\ref{tab:alignment_layer_selection} reports per-task results and additional baselines. These results also reveal that increasing the number of selected layers does not necessarily improve alignment, motivating a separate analysis of the
effect of the layer budget and its cost to output quality. We therefore report both alignment and generation perplexity in Table~\ref{tab:alignment_layer_selection_without_l2_diff}.

\textbf{Sequential Steering.}
Table~\ref{tab:main} compares simultaneous and greedy sequential steering at matched layer budgets and steering strength, with per-task results in Table~\ref{tab:sequential_detail}. Sequential selection changes the results more noticeably than in our earlier aggregate comparison, but the
relative ranking remains model-dependent. For Llama-3B, sequential steering favors LayerNavigator at $K=2$ ($80.61$ vs.\ $80.41$) but favors \ourmethod{} at $K=3$ ($78.13$ vs.\ $75.17$). On Llama-8B, \ourmethod{} slightly exceeds LayerNavigator at $K=3$ ($85.58$ vs.\ $85.24$). On Qwen-3B, the unit-sphere variant of \ourmethod{} remains substantially stronger at both $K=2$ and $K=3$ ($76.72$ and $79.19$), while the standardized variant trails LayerNavigator. For Qwen-7B and Qwen-32B, the differences are smaller. Overall, sequential re-selection can change the preferred method as additional layers are introduced, but its effect depends strongly on both the model and the scoring variant.

\paragraph{Model Utility After Steering.}
\begin{wrapfigure}{r}{0.45\linewidth}
\vspace{-9pt}
\centering
\includegraphics[width=\linewidth]{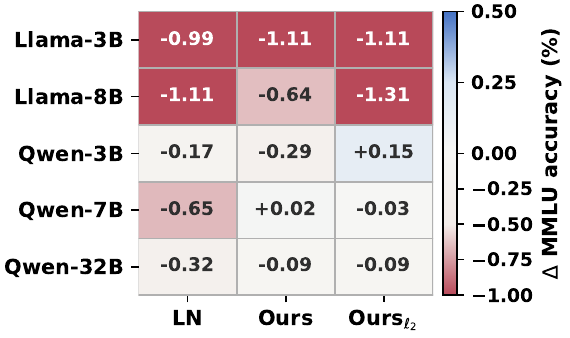}
\vspace{-20pt}
\caption{Change in MMLU accuracy relative to the unsteered model.}
\label{fig:tradeoff_runtime}
\end{wrapfigure}

Perplexity captures output fluency but does not directly measure whether steering preserves general model capabilities. We therefore evaluate the steered models on MMLU~\citep{hendrycks2021measuring}. Figure~\ref{fig:tradeoff_runtime} shows the change in MMLU accuracy relative to the unsteered model, averaged over $K=\{1,2,3\}$. Across models, the changes are small for all three steering variants, with \ourmethod{} remaining close to the unsteered baseline and showing near-zero average change on Qwen-7B and Qwen-32B. This suggests that the alignment gains from \ourmethod{} generally come with limited impact on downstream task performance. Full per-$K$ results and comparisons are reported in Appendix~\ref{mmlu_utility}.

\begin{figure*}[t]
    \centering
    \raisebox{0pt}[0pt][0pt]{%
    \begin{minipage}[t]{0.54\textwidth}
        \vspace{-91pt}
        \centering
        \setlength{\aboverulesep}{0pt}
        \setlength{\belowrulesep}{0pt}
        \setlength{\extrarowheight}{.5ex}
        \renewcommand{\tabcolsep}{1pt}
        \scriptsize
        \begin{tabular}{l ccc ccc c}
            \toprule
            & \multicolumn{3}{c}{\ourmethod}
            & \multicolumn{3}{c}{W2S-Multi} & \\
            \cmidrule(lr){2-4}
            \cmidrule(lr){5-8}
            & Prob ($\uparrow$) & PPL ($\downarrow$) & Time ($\downarrow$)
            & Prob ($\uparrow$) & PPL ($\downarrow$) & Time ($\downarrow$)
            & $\Delta$Prob \\
            \midrule
            Llama-3B & \textbf{78.22} & 26.10 & \textbf{7.7}
            & 78.06 & 26.08 & 14.9 & $-0.16$ \\
            Llama-8B & 84.53 & 11.22 & \textbf{10.7}
            & \textbf{85.83} & 11.24 & 27.4 & $+1.30$ \\
            Qwen-3B & 62.28 & 7.00 & \textbf{3.3}
            & \textbf{68.58} & 8.44 & 6.1 & $+6.30$ \\
            Qwen-7B & 85.86 & 18.72 & \textbf{4.6}
            & \textbf{89.03} & 19.97 & 7.7 & $+3.17$ \\
            Qwen-32B & 83.60 & 5.84 & \textbf{15.9}
            & \textbf{84.75} & 6.92 & 23.6 & $+1.15$ \\
            \midrule
            Average & 78.90 & 13.78 & \textbf{8.4}
            & \textbf{81.25} & 14.53 & 15.9 & $+2.35$ \\
            \bottomrule
        \end{tabular}

        \captionof{table}{
            Alignment probability (\%) and perplexity averaged over six persona tasks and
            $K\in\{1,2,3\}$, and total running time in minutes for all six tasks for
            \ourmethod{}  and W2S-Multi.
        }
        \label{tab:toposteer_vs_w2s_avg}
    \end{minipage}
    }%
    \hfill
    \begin{minipage}[t]{0.42\textwidth}
        \centering
        \includegraphics[
            width=\linewidth
        ]{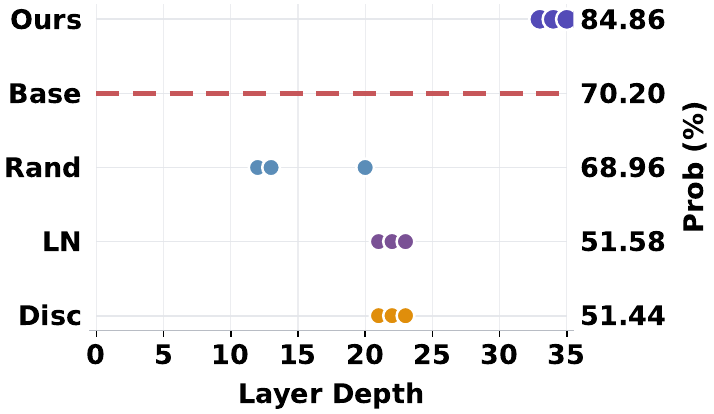}

        \captionof{figure}{Layers selected and resulting alignment on Qwen2.5-3B (Conscientiousness, $k=3$) for LayerNavigator and $\text{Ours}_{\ell_2}$.}
        \label{fig:layer_selection}
    \end{minipage}
\end{figure*}

\paragraph{Which Layers \ourmethod{} Selects?}
Figure~\ref{fig:layer_selection} illustrates how different selection criteria can lead to substantially different intervention layers. On Qwen-3B for \texttt{conscientiousness} at $K=3$, Disc and LayerNavigator select layers $21$--$23$, while \ourmethod{}$_{\ell_2}$ selects layers $33$--$35$. This difference is accompanied by a large change in aligned-answer probability, from $51.58\%$ for LayerNavigator to $84.86\%$ for \ourmethod{}$_{\ell_2}$. The result illustrates that the choice of layer-selection criterion can
matter as much as the number of layers selected. More generally, the two criteria rank layers according to different geometric properties of the representation, so their selected layer sets need not overlap. This case study provides an empirical view of the layer-selection shift; the full selected-layer sequences across models and tasks are reported in
Appendix~\ref{additional_layer_selection}.

\textbf{Comparison with Evaluation-Based Selection.}
We additionally compare \ourmethod{} with W2S-Multi~\citep{nelwan2026deployable}, a per-instance multi-layer method that learns layer selection from evaluation-based steerability
signals. Table \ref{tab:toposteer_vs_w2s_avg} shows the alignment performance on the same five models across six persona tasks. Overall, W2S-Multi achieves higher average alignment ($81.25\%$ vs.\ $78.90\%$) at a slightly higher perplexity ($14.53$ vs.\ $13.78$), but requires substantially more computation: $15.9$ minutes versus $8.4$ minutes for the complete evaluation.
The additional cost arises from evaluating candidate layers on labelled validation prompts: its offline selection cost is $\mathcal{O}(L N_v C_f)$, where $L$ is the number of candidate layers, $N_v$ the number of validation prompts, and $C_f$ the cost of a steered forward pass. In contrast, \ourmethod{} computes its layer score directly from the contrastive extraction data, without held-out steering evaluations or selector training. We provide setup and implementation details in Appendix~\ref{evaluation_based_selection_method}.

\section{Connectivity versus Discriminability}
\label{sec:information_encoding_geometry}
The experiments show that connectivity often ranks layers better than separability in Table~\ref{tab:main}. We therefore ask why the two criteria differ: Sec.~\ref{sec:sepscale} identifies how separation affects the score; Sec.~\ref{sec:ln} shows that the criteria reverse their ordering along a controlled geometric axis; and Sec.~\ref{sec:polarity} shows that they respond differently to the answer-polarity axis in real activations.

\subsection{What Standardization Couples the Score To}
\label{sec:sepscale}
Corollary~\ref{cor:inv} says the score ignores the mean displacement of the cloud it is given. Separation still reaches it, through the standardization~\eqref{eq:zscore}: for balanced classes the pooled variance of coordinate $k$ is $\sigma_{l,k}^{2}=\bar W_{l,k}+\tfrac14\bigl(\mu^{\mathrm{raw}}_{\delta,l,k}\bigr)^{2}$, where $\bar W_{l,k}$ is the pooled within-class variance of the raw activations and $\mu^{\mathrm{raw}}_{\delta,l}$ the raw mean difference.

\begin{proposition}[Standardization couples the score to class separation]
    \label{prop:sepscale}
    Write $\rho_{l,k}:=(\mu^{\mathrm{raw}}_{\delta,l,k})^{2}/(4\bar W_{l,k})$ for the coordinate-wise separation-to-noise ratio, and $H_l:=\{\eta^{i}_{l}\}_{i=1}^{N}$ with $\eta^{i}_{l,k}:=\xi^{\mathrm{raw},i}_{l,k}/\bar W_{l,k}^{1/2}$ for the within-class-standardized residual displacements. Then
    \[
    \cL(\Delta_l)
    \;=\;
    \cL\bigl(H_l\Lambda_l\bigr),
    \qquad
    \Lambda_l:=\operatorname{diag}\bigl((1+\rho_{l,k})^{-1/2}\bigr)_{k=1}^{d},
    \]
    so $\cL(\Delta_l)$ is non-increasing in every $\rho_{l,k}$, strictly decreasing in $\rho_{l,k}$ whenever coordinate $k$ is non-constant on $\Xi_l$, and equal to $\cL(H_l)/\sqrt{1+\rho_l}$ when the separation is spread evenly, $\rho_{l,k}\equiv\rho_l$ (proof in Appendix~\ref{app:sepablation}).
\end{proposition}

Separation therefore enters $\cL$ only as a divisor. $D_l$ is invariant to a common rescaling of all coordinates but not to the coordinate-wise one of \eqref{eq:zscore}, so both criteria move under the shared standardization, and part of any measured anti-correlation between them is attributable to it rather than to displacement geometry. Appendix~\ref{app:sepablation} gives the separation-free form $\cL(H_l)$ of the score, which the same cached activations yield at no extra cost.

\subsection{Decoupling from LayerNavigator's Discriminability}
\label{sec:ln}
LayerNavigator scores a layer with a discriminability term $D_{l}$, a Fisher-type ratio of projected
between-class to total scatter, together with a directional-consistency term. Consider the family that fixes the
class means and scales the within-class residuals by $\sigma>0$, giving displacements
$\Delta_l(\sigma)=\{\mu_{\delta,l}+\sigma\xi_l^i\}_{i=1}^{N}$, with the standardization held fixed. Writing $B_l$ and $W_l$ for the
projected between- and within-class terms, LayerNavigator's discriminability reduces to
$D_l(\sigma)=B_l/(B_l+\sigma^{2}W_l)$ (Appendix~\ref{app:ln}).

\begin{theorem}[Separability--connectivity decoupling]
    \label{thm:decoupling}
    Assume $\cL(\Xi_l) > 0$ and $B_l, W_l > 0$. Then, for every $\sigma > 0$, total $H_0$ persistence is
    strictly increasing in $\sigma$, whereas $D_l$ is strictly decreasing. The two criteria therefore
    induce opposite orderings on the family $\{\Delta(\sigma)\}_{\sigma>0}$.
\end{theorem}
\vspace{-10pt}
\begin{proof}[Proof sketch]
    By Corollary~\ref{cor:inv} and $1$-homogeneity, $\TotPers(\Delta(\sigma))=\sigma\cL(\Xi_l)$, so
    $\partial_\sigma\TotPers = \cL(\Xi_l) > 0$. Differentiating $D_l(\sigma)$ gives
    $\partial_\sigma D_l = -2B_l\sigma W_l/(B_l+\sigma^2W_l)^2 < 0$. The full proof and the reduction of $D_l(\sigma)$ are in Appendix~\ref{app:ln}.
\end{proof}
\vspace{-5pt}
The theorem concerns a one-parameter family at a fixed layer and the discriminability component alone: real layers may also differ along directions on which the two scores agree, which is why the selectors sometimes coincide, and the consistency component and the full published score enter through the comparison of Sec.~\ref{sec:experiments_alignment}. If the standardization is instead re-estimated along the family, $\cL$ remains strictly increasing but saturates at the separation-free score $\cL(H_l)$ of Proposition~\ref{prop:sepscale}, and $D_l$ keeps the reduction above whenever the re-standardization is a common scalar (Remark~\ref{rem:restd}). A two-dimensional instance is Figure~\ref{fig:teaser}(b), constructed in Appendix~\ref{app:toy}, which also shows the two ways a cloud can reach a high score, with length spread over many short edges or concentrated in a single bridge, the case the bridge share~\eqref{eq:bridge} detects.

\subsection{Sensitivity to the Answer-Polarity Axis}
\label{sec:polarity}
The contrastive format imposes a second split on $\Delta_l$, an \emph{answer-polarity axis}, which we characterize and then remove, as a diagnostic, to see how each criterion responds. Formal statements and proofs are in Appendix~\ref{app:anatomy}.

\begin{figure}[t]
			\centering
			\includegraphics[width=\textwidth]{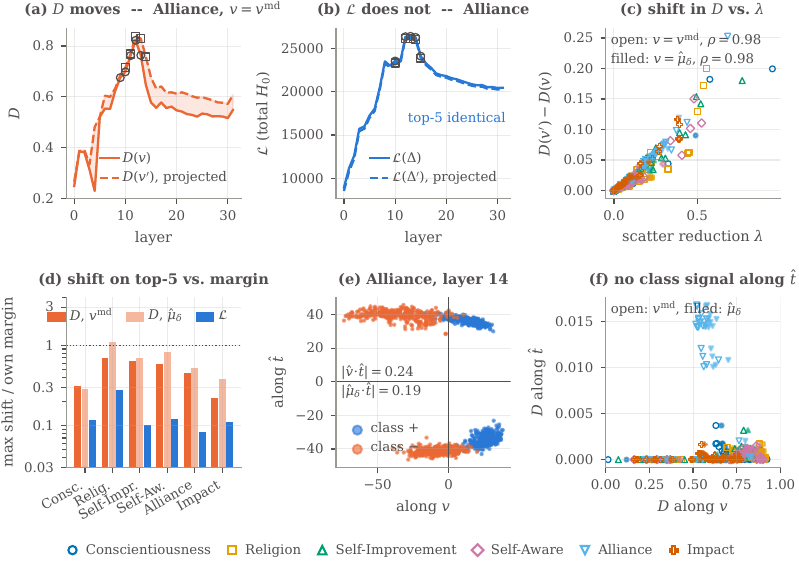}
			\caption{\textbf{Sensitivity to the answer-polarity axis}; Alliance Building shown in (a), (b) and (e).
				(a) $D$ along $v^{\mathrm{raw}}$ before and after the axis is projected out, with each curve's top-$5$ layers marked. (b) The same for $\cL$, whose top-$5$ set is unchanged. (c) The shift in $D$ against the scatter reduction $\lambda$ of Proposition~\ref{prop:leak}, all six tasks and layers. (d) The largest shift among each criterion's own top-$5$ layers, divided by its rank-$1$-to-rank-$5$ gap.
				(e) The activations at layer $14$ in the plane of the steering direction and the polarity axis: the axis separates the lobes but not the classes. (f) Discrimnability along the polarity axis against
				discriminability along the steering direction, all tasks and depths. The polarity sign of each pair is read off the cloud as the side of the longest spanning-tree edge, which splits every cloud into $350/350$ with the same partition at all $32$ layers.}
			\label{fig:leak_full}
		\end{figure}

\textbf{The answer-polarity axis.}
\Eqref{eq:displacement_decomposition} splits each displacement into a common part and a residual. The way contrastive pairs are built imposes a second split. Each example consists of one prompt and two candidate responses, one matching the target behavior and one not, and the activations
$\tilde a_l(x_i^+)$ and $\tilde a_l(x_i^-)$ are taken after appending the matching and non-matching response, respectively. The prompt is the same in both, so it cancels in the difference, and what remains is the difference between the two response tokens plus a behavior term. In the persona datasets, the two responses are \emph{Yes} and \emph{No}. If the behavior-matching answer for example $i$ is \emph{Yes}, the token part of $\delta_l^i$ is $z_{\mathrm{yes}}-z_{\mathrm{no}}$; if it
is \emph{No}, the token part is the same vector with the opposite sign. Writing $t_l$ for half that difference, every displacement carries either $+t_l$ or $-t_l$: $\delta_l^i \;=\; \pm\,t_l \;+\; b_l^i$
where $b_l^i$ is the behavior term. The sign records how the dataset was written for that example, not anything about the behavior being steered. We estimate $t_l$ as the half-difference of the displacement means of the \emph{Yes}- and \emph{No}-matching examples, and write $\hat t_l$ for the corresponding unit vector. Because the two polarities are balanced, the $\pm t_l$ contributions cancel in the mean, so MD steering is blind to the axis. The cloud is not: the two signs produce two lobes separated primarily along $\hat t_l$ (Figure~\ref{fig:anatomy}). The axis carries essentially no class information: $D_l$ evaluated along $\hat t_l$ is close to zero across tasks and layers (Figure~\ref{fig:leak_full}(f)), as expected, since the sign records which response token is designated as behavior-matching, not the behavioral class. To isolate its effect, we project it out with $P_{\hat t_l}=I-\hat t_l\hat t_l^\top$; the projected cloud is a diagnostic only and is never used for steering.

\textbf{Discriminability under projection.} Let $v=\mu_{\delta,l}$ and $v'=P_{\hat t_l}v$, with unit vectors $\hat v$ and $\hat v'$. Projecting the activations and evaluating $D_l$ along $v'$ gives the same value as evaluating the original scatter matrices along $v'$ (Lemma~\ref{lem:projid}), so the projection acts on $D_l$ as a change of direction, not of data. Let $\varepsilon:=\langle\hat v,\hat t_l\rangle$ measure the alignment of the polarity axis with the steering direction, and $\lambda:=1-W(\hat v')/W(\hat v)$ the fractional reduction in within-class scatter from $\hat v$ to $\hat v'$. Then $B(\hat v')=(1-\varepsilon^{2})B(\hat v)$ exactly, and
\begin{equation}
    D(v') > D(v)
    \quad\Longleftrightarrow\quad
    \lambda > \varepsilon^2
    \label{eq:leak_sign}
\end{equation}
(Proposition~\ref{prop:leak}): removing the axis improves discriminability only when the scatter it removes outweighs the between-class signal lost by tilting the steering direction. The shift vanishes when $\varepsilon=0$, and when $\hat t_l$ is an eigenvector of $S^w_l$ it is positive only if the axis carries more than $1+(1-\varepsilon^{2})^{-1}\approx2$ times the within-class scatter of the residual steering direction.

\textbf{Connectivity under projection.}
For $\cL$ the projection acts on the geometry itself. Let $\Delta'_l:=P_{\hat t_l}\Delta_l$. Orthogonal projection cannot increase pairwise distances, so $\cL(\Delta'_l)\le\cL(\Delta_l)$, and the deficit is bounded by the components along $\hat t_l$ of the edges of the spanning tree of the projected cloud (Proposition~\ref{prop:polinv}). In a two-lobe cloud it has two parts: the bridge component along $\hat t_l$, and the superposition of the two lobes, which changes the tree from spanning two sets of about $N/2$ points to spanning $N$ points in the residual space, a relative change of about $\log 2/D_{\mathrm{int}}$ for lobes of intrinsic dimension $D_{\mathrm{int}}$ (Remark~\ref{rem:asym}).

\textbf{Comparison.}
The two criteria thus reach the projected-out direction through different mechanisms: $\cL$ through its contribution to the spanning tree, $D_l$ through the competition between $\varepsilon$ and $\lambda$ in \eqref{eq:leak_sign}. Figure~\ref{fig:leak_full} reports the measurements: panel (c) shows the shift in $D$ against $\lambda$, and panel (d) puts both shifts on a common scale, dividing each by the gap between the criterion's own rank-$1$ and rank-$5$ layer scores. The projection is a diagnostic, not a fix: it does not reconcile the two rankings, moving LayerNavigator's ordering towards \ourmethod{}'s on one task and away from it on the others. The claim is narrower: a criterion's value should not be dominated by a direction with zero between-class variance.

\paragraph{Additional Analysis.}
Due to space limit, we provide other supporting experiments and analyses in the Appendix. Appendix~\ref{app:tda_ablation} ablates the topological descriptor, filtration distance and layer budget used to rank layers, Appendix~\ref{sec_mst_vs_total_dist} compares connectivity length with the sum of distances to the mean and shows when the bridge share is needed, Appendix~\ref{app:toy} numerically verifies the lobe-merging term of Remark~\ref{rem:asym}, Appendix~\ref{app:mst} proves the stability of total persistence and connectivity length, and Appendix~\ref{app:ln} relates LayerNavigator's within-class scatter to the spread of the displacement cloud.

\section{Conclusion}
We introduced \ourmethod{}, a validation-free layer selector for mean-difference activation steering that uses the geometry of per-example displacement clouds rather than class separability. The score is total $H_0$ persistence, equivalently the Euclidean minimum spanning tree length, and requires neither held-out steering evaluations nor auxiliary training. Across six persona tasks and five models, \ourmethod{} improves alignment in many settings at comparable perplexity cost, supported by theoretical and geometric analyses contrasting connectivity and discriminability.

\textbf{Limitations.} The method assumes matched positive and negative examples and currently considers only mean-difference steering, leaving alternative extractors for future work. Multi-layer steering can degrade perplexity on some models, while Theorem~\ref{thm:decoupling} applies only to a fixed layer and its across-layer implications remain empirical; Appendix~\ref{sec_mst_vs_total_dist} further shows that the guarantees can miss mode splits concentrated in a few coordinates.  A promising direction is stylistic control, where author-style imitation or obfuscation is often achieved through prompting ~\citep{nguyen2025unraveling}; because such stylistic signals may be diffuse and layer-dependent, validation-free layer selection could be particularly useful in this setting.

\newpage
\subsection*{AI use statement}
In this work, we used generative AI tools for language editing and code assistance. Specifically, AI tools were used to correct grammar, improve clarity, polish manuscript text, and refine code for data visualization and related implementation tasks. We did not use generative AI tools to conduct experiments, generate or modify experimental results, construct datasets, or formulate scientific claims. All AI-assisted text and code were reviewed by the authors; generated code was manually inspected, executed, and verified for correctness. We take full responsibility for the final content of this work, including all text, claims, analyses, results, and artifacts produced with the aid of generative AI.

\subsection*{Ethics statement}
Activation steering can be used to improve model behavior, such as increasing helpfulness or reducing undesirable responses, but it can also be misused to manipulate model outputs or bypass intended safeguards. This work studies controlled persona-style steering tasks and does not train or release a new model. Because the method operates at inference time, it could be applied dynamically without changing model weights. Any release of code or steering vectors should therefore include documentation of intended use, limitations, and potential misuse risks. Future deployment of steering methods should include safety evaluations to ensure that behavioral interventions do not degrade harmlessness, truthfulness, or robustness.

\subsection*{Reproducibility statement}
The score is defined in Sec.~\ref{sec:score} and computed by Algorithm~\ref{alg:toposteer}; all formal statements are proved in Appendices~\ref{app:mst}--\ref{app:anatomy} with their assumptions stated. Tasks, models, baselines, metrics, steering strength, and layer budgets are given in Appendix~\ref{app:expdetails}, per-task results in Appendix~\ref{app:additional}, and the synthetic constructions behind every toy figure in Appendix~\ref{app:toy} and Appendix~\ref{sec_mst_vs_total_dist}.

\subsubsection*{Acknowledgments}
E.P.-H. is funded by the European Union's Horizon 2020 research and innovation programme under the Marie Sk\l{}odowska-Curie grant agreement No.~101153039 (CHALKS).

\bibliography{iclr2027_conference}

\begin{thebibliography}{34}
\providecommand{\natexlab}[1]{#1}
\providecommand{\url}[1]{\texttt{#1}}
\expandafter\ifx\csname urlstyle\endcsname\relax
  \providecommand{\doi}[1]{doi: #1}\else
  \providecommand{\doi}{doi: \begingroup \urlstyle{rm}\Url}\fi

\bibitem[Arditi et~al.(2024)Arditi, Obeso, Syed, Paleka, Rimsky, Gurnee, and Nanda]{arditi2024refusal}
Andy Arditi, Oscar~Balcells Obeso, Aaquib Syed, Daniel Paleka, Nina Rimsky, Wes Gurnee, and Neel Nanda.
\newblock Refusal in language models is mediated by a single direction.
\newblock In \emph{The Thirty-eighth Annual Conference on Neural Information Processing Systems}, 2024.
\newblock URL \url{https://openreview.net/forum?id=pH3XAQME6c}.

\bibitem[Edelsbrunner \& Harer(2008)Edelsbrunner and Harer]{edelsbrunner2008persistent}
Herbert Edelsbrunner and John Harer.
\newblock Persistent homology: A survey.
\newblock \emph{Contemporary Mathematics}, 2008.
\newblock URL \url{https://pub.ista.ac.at/~edels/Papers/2008-02-PersistentHomology.pdf}.

\bibitem[Fay et~al.(2026)Fay, Garc{\'\i}a-Redondo, Wang, Dubossarsky, and Monod]{fay2026the}
Aideen Fay, In{\'e}s Garc{\'\i}a-Redondo, Qiquan Wang, Haim Dubossarsky, and Anthea Monod.
\newblock The shape of adversarial influence: Characterizing {LLM} latent spaces with persistent homology.
\newblock In \emph{The Fourteenth International Conference on Learning Representations}, 2026.
\newblock URL \url{https://openreview.net/forum?id=v2PglvLLKT}.

\bibitem[Ferrando et~al.(2025)Ferrando, Suau, Gonz{\`a}lez, and Rodriguez]{ferrando2025dynamically}
Alex Ferrando, Xavier Suau, Jordi Gonz{\`a}lez, and Pau Rodriguez.
\newblock Dynamically scaled activation steering.
\newblock \emph{arXiv preprint arXiv:2512.03661}, 2025.
\newblock URL \url{https://arxiv.org/abs/2512.03661}.

\bibitem[Gabrielsson et~al.(2020)Gabrielsson, Nelson, Dwaraknath, and Skraba]{pmlr-v108-gabrielsson20a}
Rickard~Br\"uel Gabrielsson, Bradley~J. Nelson, Anjan Dwaraknath, and Primoz Skraba.
\newblock A topology layer for machine learning.
\newblock In Silvia Chiappa and Roberto Calandra (eds.), \emph{Proceedings of the Twenty Third International Conference on Artificial Intelligence and Statistics}, volume 108 of \emph{Proceedings of Machine Learning Research}, pp.\  1553--1563. PMLR, 26--28 Aug 2020.
\newblock URL \url{https://proceedings.mlr.press/v108/gabrielsson20a.html}.

\bibitem[Gadgil et~al.(2026)Gadgil, Lin, and Lee]{gadgil2026steer}
Soham Gadgil, Chris Lin, and Su-In Lee.
\newblock Where to steer: Input-dependent layer selection for steering improves llm alignment.
\newblock \emph{arXiv preprint arXiv:2604.03867}, 2026.
\newblock URL \url{https://arxiv.org/abs/2604.03867}.

\bibitem[Gardinazzi et~al.(2025)Gardinazzi, Viswanathan, Panerai, ansuini, Cazzaniga, and Biagetti]{gardinazzi2024persistent}
Yuri Gardinazzi, Karthik Viswanathan, Giada Panerai, Alessio ansuini, Alberto Cazzaniga, and Matteo Biagetti.
\newblock Persistent topological features in large language models.
\newblock In \emph{Forty-second International Conference on Machine Learning}, 2025.
\newblock URL \url{https://openreview.net/forum?id=qAHnSkHvsm}.

\bibitem[Gower \& Ross(1969)Gower and Ross]{7efdf403-c7cc-3f04-98a8-023222e4fc84}
J.~C. Gower and G.~J.~S. Ross.
\newblock Minimum spanning trees and single linkage cluster analysis.
\newblock \emph{Journal of the Royal Statistical Society. Series C (Applied Statistics)}, 18\penalty0 (1):\penalty0 54--64, 1969.
\newblock ISSN 00359254, 14679876.
\newblock URL \url{http://www.jstor.org/stable/2346439}.

\bibitem[Grattafiori et~al.(2024)Grattafiori, Dubey, Jauhri, Pandey, Kadian, Al-Dahle, Letman, Mathur, Schelten, Vaughan, et~al.]{grattafiori2024llama}
Aaron Grattafiori, Abhimanyu Dubey, Abhinav Jauhri, Abhinav Pandey, Abhishek Kadian, Ahmad Al-Dahle, Aiesha Letman, Akhil Mathur, Alan Schelten, Alex Vaughan, et~al.
\newblock The llama 3 herd of models.
\newblock \emph{arXiv}, 2024.
\newblock URL \url{https://arxiv.org/abs/2407.21783}.

\bibitem[Hendrycks et~al.(2021)Hendrycks, Burns, Basart, Zou, Mazeika, Song, and Steinhardt]{hendrycks2021measuring}
Dan Hendrycks, Collin Burns, Steven Basart, Andy Zou, Mantas Mazeika, Dawn Song, and Jacob Steinhardt.
\newblock Measuring massive multitask language understanding.
\newblock In \emph{International Conference on Learning Representations}, 2021.
\newblock URL \url{https://openreview.net/forum?id=d7KBjmI3GmQ}.

\bibitem[Houlsby et~al.(2019)Houlsby, Giurgiu, Jastrzebski, Morrone, De~Laroussilhe, Gesmundo, Attariyan, and Gelly]{houlsby2019parameter}
Neil Houlsby, Andrei Giurgiu, Stanislaw Jastrzebski, Bruna Morrone, Quentin De~Laroussilhe, Andrea Gesmundo, Mona Attariyan, and Sylvain Gelly.
\newblock Parameter-efficient transfer learning for nlp.
\newblock In \emph{International conference on machine learning}, pp.\  2790--2799. PMLR, 2019.

\bibitem[Hu et~al.(2022)Hu, yelong shen, Wallis, Allen-Zhu, Li, Wang, Wang, and Chen]{hu2022lora}
Edward~J Hu, yelong shen, Phillip Wallis, Zeyuan Allen-Zhu, Yuanzhi Li, Shean Wang, Lu~Wang, and Weizhu Chen.
\newblock Lo{RA}: Low-rank adaptation of large language models.
\newblock In \emph{International Conference on Learning Representations}, 2022.
\newblock URL \url{https://openreview.net/forum?id=nZeVKeeFYf9}.

\bibitem[Kong et~al.(2024)Kong, Wang, Mu, Du, Zhuang, Zhou, Song, Zhang, Wang, and Zhang]{kong2024aligning}
Lingkai Kong, Haorui Wang, Wenhao Mu, Yuanqi Du, Yuchen Zhuang, Yifei Zhou, Yue Song, Rongzhi Zhang, Kai Wang, and Chao Zhang.
\newblock Aligning large language models with representation editing: A control perspective.
\newblock \emph{The Thirty-eighth Annual Conference on Neural Information Processing Systems}, 37, 2024.
\newblock \url{https://openreview.net/forum?id=yTTomSJsSW}.

\bibitem[Li et~al.(2023)Li, Patel, Vi{\'e}gas, Pfister, and Wattenberg]{li2023inference}
Kenneth Li, Oam Patel, Fernanda Vi{\'e}gas, Hanspeter Pfister, and Martin Wattenberg.
\newblock Inference-time intervention: Eliciting truthful answers from a language model.
\newblock \emph{Advances in neural information processing systems}, 2023.
\newblock URL \url{https://openreview.net/forum?id=aLLuYpn83y}.

\bibitem[Marks \& Tegmark(2024)Marks and Tegmark]{marks2024geometry}
Samuel Marks and Max Tegmark.
\newblock The geometry of truth: Emergent linear structure in large language model representations of true/false datasets.
\newblock In \emph{First Conference on Language Modeling}, 2024.
\newblock URL \url{https://openreview.net/forum?id=aajyHYjjsk}.

\bibitem[Naitzat et~al.(2020)Naitzat, Zhitnikov, and Lim]{JMLR:v21:20-345}
Gregory Naitzat, Andrey Zhitnikov, and Lek-Heng Lim.
\newblock Topology of deep neural networks.
\newblock \emph{Journal of Machine Learning Research}, 21\penalty0 (184):\penalty0 1--40, 2020.
\newblock URL \url{http://jmlr.org/papers/v21/20-345.html}.

\bibitem[Nelwan \& Wicaksono(2026)Nelwan and Wicaksono]{nelwan2026deployable}
Muhammad Faishal~Adly Nelwan and Alfan~Farizki Wicaksono.
\newblock Deployable per-instance multi-layer activation steering for large language models.
\newblock \emph{arXiv preprint arXiv:2608.08829}, 2026.
\newblock URL \url{https://arxiv.org/abs/2608.08829}.

\bibitem[Nguyen \& Le(2024)Nguyen and Le]{nguyen-le-2024-generalizability}
Tuc Nguyen and Thai Le.
\newblock Generalizability of mixture of domain-specific adapters from the lens of signed weight directions and its application to effective model pruning.
\newblock In Lun-Wei Ku, Andre Martins, and Vivek Srikumar (eds.), \emph{Proceedings of the 62nd Annual Meeting of the {A}ssociation for {C}omputational {L}inguistics (Volume 1: Long Papers)}, pp.\  12956--12973, Bangkok, Thailand, August 2024. Association for Computational Linguistics.
\newblock \doi{10.18653/v1/2024.acl-long.700}.
\newblock URL \url{https://aclanthology.org/2024.acl-long.700/}.

\bibitem[Nguyen \& Le(2026{\natexlab{a}})Nguyen and Le]{nguyen2026atlas}
Tuc Nguyen and Thai Le.
\newblock Atlas: Adaptive test-time latent steering with external verifiers for enhancing llms reasoning.
\newblock \emph{arXiv preprint arXiv:2601.03093}, 2026{\natexlab{a}}.
\newblock URL \url{https://arxiv.org/abs/2601.03093v4}.

\bibitem[Nguyen \& Le(2026{\natexlab{b}})Nguyen and Le]{nguyen2026beyond}
Tuc Nguyen and Thai Le.
\newblock Beyond linear activation steering: Invertible latent transformations for controlling llm behavior.
\newblock \emph{The Fortieth Annual Conference on Neural Information Processing Systems}, 2026{\natexlab{b}}.
\newblock URL \url{https://arxiv.org/abs/2606.08454}.

\bibitem[Nguyen et~al.(2025)Nguyen, Hu, and Le]{nguyen2025unraveling}
Tuc Nguyen, Yifan Hu, and Thai Le.
\newblock Unraveling interwoven roles of large language models in authorship privacy: Obfuscation, mimicking, and verification.
\newblock In \emph{Proceedings of the 2025 Conference on Empirical Methods in Natural Language Processing}, pp.\  14914--14930, 2025.
\newblock URL \url{https://aclanthology.org/2025.emnlp-main.753.pdf}.

\bibitem[Perez et~al.(2023)Perez, Ringer, Lukosiute, Nguyen, Chen, Heiner, Pettit, Olsson, Kundu, Kadavath, Jones, Chen, Mann, Israel, Seethor, McKinnon, Olah, Yan, Amodei, Amodei, Drain, Li, Tran-Johnson, Khundadze, Kernion, Landis, Kerr, Mueller, Hyun, Landau, Ndousse, Goldberg, Lovitt, Lucas, Sellitto, Zhang, Kingsland, Elhage, Joseph, Mercado, DasSarma, Rausch, Larson, McCandlish, Johnston, Kravec, El~Showk, Lanham, Telleen-Lawton, Brown, Henighan, Hume, Bai, Hatfield-Dodds, Clark, Bowman, Askell, Grosse, Hernandez, Ganguli, Hubinger, Schiefer, and Kaplan]{perez-etal-2023-discovering}
Ethan Perez, Sam Ringer, Kamile Lukosiute, Karina Nguyen, Edwin Chen, Scott Heiner, Craig Pettit, Catherine Olsson, Sandipan Kundu, Saurav Kadavath, Andy Jones, Anna Chen, Benjamin Mann, Brian Israel, Bryan Seethor, Cameron McKinnon, Christopher Olah, Da~Yan, Daniela Amodei, Dario Amodei, Dawn Drain, Dustin Li, Eli Tran-Johnson, Guro Khundadze, Jackson Kernion, James Landis, Jamie Kerr, Jared Mueller, Jeeyoon Hyun, Joshua Landau, Kamal Ndousse, Landon Goldberg, Liane Lovitt, Martin Lucas, Michael Sellitto, Miranda Zhang, Neerav Kingsland, Nelson Elhage, Nicholas Joseph, Noemi Mercado, Nova DasSarma, Oliver Rausch, Robin Larson, Sam McCandlish, Scott Johnston, Shauna Kravec, Sheer El~Showk, Tamera Lanham, Timothy Telleen-Lawton, Tom Brown, Tom Henighan, Tristan Hume, Yuntao Bai, Zac Hatfield-Dodds, Jack Clark, Samuel~R. Bowman, Amanda Askell, Roger Grosse, Danny Hernandez, Deep Ganguli, Evan Hubinger, Nicholas Schiefer, and Jared Kaplan.
\newblock Discovering language model behaviors with model-written evaluations.
\newblock In Anna Rogers, Jordan Boyd-Graber, and Naoaki Okazaki (eds.), \emph{Findings of the Association for Computational Linguistics: ACL 2023}, pp.\  13387--13434, Toronto, Canada, July 2023. Association for Computational Linguistics.
\newblock \doi{10.18653/v1/2023.findings-acl.847}.
\newblock URL \url{https://aclanthology.org/2023.findings-acl.847/}.

\bibitem[Rieck et~al.(2019)Rieck, Togninalli, Bock, Moor, Horn, Gumbsch, and Borgwardt]{rieck2019neural}
Bastian Rieck, Matteo Togninalli, Christian Bock, Michael Moor, Max Horn, Thomas Gumbsch, and Karsten Borgwardt.
\newblock Neural persistence: A complexity measure for deep neural networks using algebraic topology.
\newblock In \emph{International Conference on Learning Representations}, 2019.
\newblock URL \url{https://openreview.net/forum?id=ByxkijC5FQ}.

\bibitem[Rimsky et~al.(2024{\natexlab{a}})Rimsky, Gabrieli, Schulz, Tong, Hubinger, and Turner]{rimsky-etal-2024-steering}
Nina Rimsky, Nick Gabrieli, Julian Schulz, Meg Tong, Evan Hubinger, and Alexander Turner.
\newblock Steering llama 2 via contrastive activation addition.
\newblock In Lun-Wei Ku, Andre Martins, and Vivek Srikumar (eds.), \emph{Proceedings of the 62nd Annual Meeting of the Association for Computational Linguistics (Volume 1: Long Papers)}, pp.\  15504--15522, Bangkok, Thailand, August 2024{\natexlab{a}}. Association for Computational Linguistics.
\newblock \doi{10.18653/v1/2024.acl-long.828}.
\newblock URL \url{https://aclanthology.org/2024.acl-long.828/}.

\bibitem[Rimsky et~al.(2024{\natexlab{b}})Rimsky, Gabrieli, Schulz, Tong, Hubinger, and Turner]{rimsky2024steering}
Nina Rimsky, Nick Gabrieli, Julian Schulz, Meg Tong, Evan Hubinger, and Alexander Turner.
\newblock Steering llama 2 via contrastive activation addition.
\newblock In Lun-Wei Ku, Andre Martins, and Vivek Srikumar (eds.), \emph{Proceedings of the 62nd Annual Meeting of the Association for Computational Linguistics (Volume 1: Long Papers)}, pp.\  15504--15522, Bangkok, Thailand, August 2024{\natexlab{b}}. Association for Computational Linguistics.
\newblock \doi{10.18653/v1/2024.acl-long.828}.
\newblock URL \url{https://aclanthology.org/2024.acl-long.828/}.

\bibitem[Rodriguez et~al.(2025)Rodriguez, Blaas, Klein, Zappella, Apostoloff, Suau, et~al.]{rodriguez2025controlling}
Pau Rodriguez, Arno Blaas, Michal Klein, Luca Zappella, Nicholas Apostoloff, Xavier Suau, et~al.
\newblock Controlling language and diffusion models by transporting activations.
\newblock In \emph{International Conference on Learning Representations}, volume 2025, pp.\  89812--89855, 2025.

\bibitem[Sun et~al.(2025)Sun, Peng, Dai, Bai, and Cao]{sun2025layernavigator}
Hao Sun, Huailiang Peng, Qiong Dai, Xu~Bai, and Yanan Cao.
\newblock Layernavigator: Finding promising intervention layers for efficient activation steering in large language models.
\newblock In \emph{The Thirty-ninth Annual Conference on Neural Information Processing Systems}, 2025.
\newblock URL \url{https://openreview.net/forum?id=wj4lM45xQR}.

\bibitem[Team et~al.(2024)]{team2024qwen2}
Qwen Team et~al.
\newblock Qwen2 technical report.
\newblock \emph{arXiv}, 2024.
\newblock URL \url{https://arxiv.org/abs/2407.10671}.

\bibitem[Turner et~al.(2023)Turner, Thiergart, Leech, Udell, Vazquez, Mini, and MacDiarmid]{azizi2025activation}
Alexander~Matt Turner, Lisa Thiergart, Gavin Leech, David Udell, Juan~J Vazquez, Ulisse Mini, and Monte MacDiarmid.
\newblock Steering language models with activation engineering.
\newblock \emph{arXiv}, 2023.
\newblock URL \url{https://arxiv.org/abs/2308.10248}.

\bibitem[Uchendu \& Le(2026)Uchendu and Le]{10.1145/3820356.3820362}
Adaku Uchendu and Thai Le.
\newblock Topological data analysis applications in natural language processing: A survey.
\newblock \emph{SIGKDD Explor. Newsl.}, 28\penalty0 (1):\penalty0 81–101, June 2026.
\newblock ISSN 1931-0145.
\newblock \doi{10.1145/3820356.3820362}.
\newblock URL \url{https://doi.org/10.1145/3820356.3820362}.

\bibitem[Van~Nguyen \& Le(2024)Van~Nguyen and Le]{van2024adapters}
Tuc Van~Nguyen and Thai Le.
\newblock Adapters mixup: Mixing parameter-efficient adapters to enhance the adversarial robustness of fine-tuned pre-trained text classifiers.
\newblock In \emph{Proceedings of the 2024 Conference on Empirical Methods in Natural Language Processing}, pp.\  21183--21203, 2024.
\newblock URL \url{https://aclanthology.org/2024.emnlp-main.1180.pdf}.

\bibitem[Xiao et~al.(2024)Xiao, Liu, Zhang, Muennighoff, Lian, and Nie]{10.1145/3626772.3657878}
Shitao Xiao, Zheng Liu, Peitian Zhang, Niklas Muennighoff, Defu Lian, and Jian-Yun Nie.
\newblock C-pack: Packed resources for general chinese embeddings.
\newblock In \emph{Proceedings of the 47th International ACM SIGIR Conference on Research and Development in Information Retrieval}, SIGIR '24, pp.\  641–649, New York, NY, USA, 2024. Association for Computing Machinery.
\newblock ISBN 9798400704314.
\newblock \doi{10.1145/3626772.3657878}.
\newblock URL \url{https://doi.org/10.1145/3626772.3657878}.

\bibitem[Yan et~al.(2025)Yan, Nguyen, Su, Lieffers, and Le]{yan2025sharechat}
Yueru Yan, Tuc Nguyen, Bo~Su, Melissa Lieffers, and Thai Le.
\newblock Sharechat: A dataset of chatbot conversations in the wild.
\newblock \emph{arXiv preprint arXiv:2512.17843}, 2025.
\newblock URL \url{https://arxiv.org/abs/2512.17843}.

\bibitem[Zou et~al.(2023)Zou, Phan, Liu, et~al.]{zou2023representation}
Andy Zou, Long Phan, Alisa Liu, et~al.
\newblock Representation engineering: A top-down approach to ai transparency.
\newblock \emph{arXiv preprint arXiv:2310.01405}, 2023.
\newblock URL \url{https://arxiv.org/abs/2310.01405}.

\end{thebibliography}
\bibliographystyle{iclr2027_conference}

\newpage
\tableofcontents
\appendix

\newpage
\section{Notation}
\label{sec:notation}
Table~\ref{tab:notation} collects the symbols used in the paper. A layer subscript $l$ is carried where several layers are compared and suppressed where a single fixed layer is understood, so $\tilde a_i^{\,c}$ and $\tilde a_l(x_i^{c})$ denote the same object. Four symbols
carry more than one meaning; in each case the reading is fixed by context and is noted here to
avoid ambiguity. The letter $d$ denotes the residual-stream width in
Sec.~\ref{sec:information_encoding_geometry}, a generic dissimilarity in the definition of the
Vietoris--Rips complex, and a death scale when subscripted as $d_i$. The letter $\sigma$ is the
scale parameter of the homothety family, while $\sigma_l$ is the vector of per-coordinate standard
deviations used for $Z$-scoring. The letter $\alpha$ is the steering strength of Sec.~\ref{sec:experiments_alignment}. In Appendix~\ref{sec_mst_vs_total_dist}, which is self-contained, $D$ denotes the ambient dimension of the synthetic clouds and $N$ the number of points, rather than the discriminability and the number of pairs used in the main text. Finally, $D_{\mathrm{int}}$ in Remark~\ref{rem:asym} is the intrinsic dimension of a displacement lobe, not an ambient dimension. In Appendix~\ref{app:anatomy}, $B_l$ and $W_l$ denote the quadratic forms evaluated at the \emph{unit} steering direction $\hat v$; they differ from the $B_l,W_l$ of Appendix~\ref{app:ln}, evaluated at $\mu_{\delta,l}$ itself, by the factor $\lVert\mu_{\delta,l}\rVert^{2}$, which cancels in $D_l$.

\begin{table}[h]
    \centering
    \renewcommand{\tabcolsep}{1pt}
    \footnotesize
    \caption{Notation Usage}
    \label{tab:notation}
    \begin{tabular}{ll}
        \toprule
        \textbf{Symbol} & \textbf{Meaning} \\
        \midrule
        \multicolumn{2}{l}{\emph{Model and data}}\\
        $l$, $L$ & layer index; number of layers \\
        $d$ & residual-stream width at the layer under consideration \\
        $N$ & number of contrastive pairs \\
        $x_i^{+},x_i^{-}$ & the positive and negative variant of the $i$-th prompt \\
        $a_l(x)$ & residual-stream activation at layer $l$, final token position \\
        $v_l^{\mathrm{raw}}$ & mean difference of raw activations \\
        $\alpha$, $K$ & steering strength; number of layers intervened on \\
        \midrule
        \multicolumn{2}{l}{\emph{Normalized activations and scatter}}\\
        $\tilde a_i^{\,c}$ & $Z$-scored activation of example $i$ in class $c\in\{+,-\}$ \\
        $\mu_l,\sigma_l$ & coordinate-wise mean and standard deviation over all $2N$ activations \\
        $\tilde\mu^{\,c}$ & class mean of the normalized activations \\
        $\varepsilon_l(x_i^{c})$ & within-class residual $\tilde a_i^{\,c}-\tilde\mu^{\,c}$ \\
        $\Sb,\Sw$ & between-class and within-class scatter matrices, Eqs.~\ref{eq:sb} and~\ref{eq:sw} \\
        $B,W$ & projected signal $\mu_\delta^{\top}\Sb\mu_\delta$ and dispersion $\mu_\delta^{\top}\Sw\mu_\delta$ \\
        $D_l$, $D_l(\sigma)$ & LayerNavigator discriminability; its value on the family ($B/(B+\sigma^{2}W)$) \\
        \midrule
        \multicolumn{2}{l}{\emph{Displacements}}\\
        $\delta_i$ & displacement $\tilde a_i^{+}-\tilde a_i^{-}$ of the $i$-th pair \\
        $\Delta$ & displacement cloud $\{\delta_i\}_{i=1}^{N}$ \\
        $\mu_\delta$, $\uhat$ & mean displacement $\bar\delta$ of the \emph{standardized} cloud; its unit vector \\
        $\xi_l^{i}$ & residual displacement $\varepsilon_l(x_i^{+})-\varepsilon_l(x_i^{-})$, so $\delta_i=\mu_\delta+\xi_l^{i}$ \\
        $\Xi$ & residual configuration $\{\xi_l^{i}\}_{i=1}^{N}$, i.e.\ $\Delta$ centered \\
        $\Delta(\sigma)$ & homothety family $\{\mu_\delta+\sigma\xi_l^{i}\}$, defined on the standardized cloud (Appendix~\ref{app:ln}) \\
        \midrule
        \multicolumn{2}{l}{\emph{Topology}}\\
        $\mathrm{VR}_\epsilon(X)$ & Vietoris--Rips complex of $X$ at scale $\epsilon$ \\
        $\mathrm{Dgm}(X)$, $B_0(X)$ & persistence diagram; $0$-dimensional barcode \\
        $d_i$ & death scale of the $i$-th $H_0$ class \\
        $\TotPers(X)$ & total $H_0$ persistence $\sum_i d_i$ (essential class discarded) \\
        $\MST(X)$, $\cL(X)$ & minimum spanning tree; its total edge weight (connectivity length) \\
        $b_l$ & bridge share $\max_{e\in\MST(\Delta)}w(e)/\cL(\Delta)$, Eq.~\eqref{eq:bridge} \\
        $W_1$ & $1$-Wasserstein distance on diagrams, $\ell^{1}$ ground metric \\
        $\mathrm{Diag}$ & diagonal $\{(t,t)\}$ of the persistence plane \\
        $\hat x$, $P_{\hat t}$ & unit vector $x/\lVert x\rVert$; orthogonal projector $I-\hat t\hat t^{\top}$ \\
        \bottomrule
    \end{tabular}
\end{table}

\section{Topological Data Analysis}\label{sec:tda}
Topological Data Analysis (TDA) provides a framework for studying the shape and structure of data by extracting topological features from discrete representations such as point clouds, graphs, or geometric datasets. In this work, we focus on point clouds obtained from neural network activations and use persistent homology to characterize their topological structure across different spatial scales.

\subsection{Simplicial complexes}
Simplicial complexes provide a convenient representation for studying the topology of discrete data. Given a set of vertices $V$, a simplicial complex $K$ is a collection of non-empty subsets of $V$ such that, whenever $\tau \in K$, all faces of $\tau$ are also contained in $K$. The elements of $K$ are called simplices. This representation allows geometric information from a point cloud to be converted into a combinatorial structure on which topological properties can be computed.

\subsection{Vietoris-Rips filtration}
Let $(X,d_X)$ be a metric space, where in our setting $d_X$ denotes the Euclidean distance. Given a scale parameter $\varepsilon \geq 0$, the Vietoris--Rips (VR) complex of a point cloud $X$ is defined as
\begin{align}
    \label{vr_formular}
    \mathrm{VR}_\varepsilon(X) = \left\{\emptyset \neq \tau \subseteq X \mid d(x_i, x_j) \leq \varepsilon,\; \forall x_i, x_j \in \tau \right\}
\end{align}

In other words, a simplex is included in the complex when all pairwise distances between its vertices are at most $\varepsilon$. As $\varepsilon$ increases, more simplices are added, resulting in a nested sequence
\[
\mathrm{VR}_{\varepsilon_{1}}(X)\subseteq\mathrm{VR}_{\varepsilon_{2}}(X),\,\,\,\,\,\,\,\,\,\,\varepsilon_{1}\leq\varepsilon_{2}
\]
This sequence is called the Vietoris--Rips filtration. 

\subsection{Persistent Homology}
Once a filtration has been constructed, homology can be used to identify topological structures at each scale. In a simplicial complex, homology distinguishes structures such as connected components, loops, and higher-dimensional cavities.

For a simplicial complex $K$, the $m$-th homology group can be expressed as
\[
H_m(K)
=
\frac{\ker \partial_m}
{\operatorname{im} \partial_{m+1}},
\]
where $\partial_m$ is the boundary operator. Intuitively, this quotient identifies structures with no boundary that are not themselves boundaries of higher-dimensional structures. More specifically, $H_0$ captures connected components, $H_1$ captures one-dimensional loops, and $H_2$ captures enclosed voids.

Persistent homology extends this analysis across the entire filtration. For two scales $\varepsilon_1 \le \varepsilon_2$, the inclusion
\[
\mathrm{VR}_{\varepsilon_1}(X) \hookrightarrow \mathrm{VR}_{\varepsilon_2}(X)
\]
induces a map between their homology groups,
\[
H_m(\mathrm{VR}_{\varepsilon_1}(X))\longrightarrow H_m(\mathrm{VR }_{\varepsilon_2}(X)).
\]

A topological feature can therefore be tracked as the scale parameter increases. Features that appear at some scale are said to be born, while features that subsequently disappear are said to die. Persistent homology records these birth and death scales, allowing topological features to be distinguished according to how long they persist across the filtration.

The resulting information can be represented by a barcode. For a fixed homological degree $m$, the barcode is a multiset of intervals
\[
B_m(X)
=
\left\{
[b_i,d_i)
\right\}_{i=1}^{|B_m(X)|},
\]
where each interval represents a topological feature, with $b_i$ and $d_i$ denoting its birth and death scales, respectively. The lifetime, or persistence, of the feature is $d_i-b_i$.

An alternative representation of the same information is given by persistence diagrams. For a fixed homological degree $m$, the persistence diagram $Dgm_m(X)$ is a locally finite multiset of points
\[
\mathrm{Dgm}_m(X)
=
\left\{
(b_i,d_i)
\right\}_{i=1}^{|\mathrm{Dgm}_m(X)|}
\subset
\left\{
(b,d)\in\mathbb{R}\times(\mathbb{R}\cup\{\infty\}) : b<d
\right\},
\]
together with the diagonal
\[
\mathrm{Diag}
=
\left\{
(t,t):t\in\mathbb{R}
\right\},
\]
which is considered with infinite multiplicity. Each interval $[b_i,d_i)$ in the barcode is represented by a point $(b_i,d_i)$ in the persistence diagram.

Finally, the total persistence in homological degree $m$ is defined as the sum of the lifetimes of all features:
\[
\operatorname{TotPers}_m(X)
=
\sum_{(b_i,d_i)\in B_m(X)}
(d_i-b_i).
\]

This quantity provides a compact measure of the overall amount of $m$-dimensional topological structure present in the point cloud across the considered range of scales. In degree $0$, the sum is taken over the finite bars only: the single essential class, which never dies, is discarded, so $\TotPers_0$ is finite. We write $\TotPers$ for $\TotPers_0$ throughout.

To compare persistence diagrams, one possible metric is the Wasserstein distance. It carries two independent indices, which we keep apart: the order $q\in[1,\infty)$ of the sum, and the ground metric $\ell^{r}$ used on the persistence plane. For two diagrams,
\[
W_q^{(r)}\bigl(\mathrm{Dgm}_m(X),\mathrm{Dgm}_m(X')\bigr)
=
\Biggl(
\inf_{\gamma}
\sum_{p\in \mathrm{Dgm}_m(X)\cup\mathrm{Diag}}
\|p-\gamma(p)\|_r^{\,q}
\Biggr)^{1/q},
\]
where the infimum is taken over all admissible bijections
\[
\gamma:
\mathrm{Dgm}_m(X)\cup\mathrm{Diag}
\longrightarrow
\mathrm{Dgm}_m(X')\cup\mathrm{Diag}.
\]
The use of the diagonal allows persistence features in one diagram to be matched to the diagonal, corresponding to features that are absent from the other diagram. Only the case $q=1$ with the $\ell^{1}$ ground metric is used in this paper; we abbreviate it $W_1:=W_1^{(1)}$, and it is the form appearing in Proposition~\ref{prop:stability}(a).

\section{$\MST$ characterization via persistent homology}
\label{app:mst}

A minimum spanning tree ($\MST$) of a weighted graph is a spanning tree that connects all vertices while minimizing the total sum of edge weights. In the setting of a point cloud $X$, we consider the complete graph whose vertices are the points in $X$ and whose edge weights are given by their pairwise Euclidean distances.

One standard way to construct an $\MST$ is Kruskal's algorithm. The algorithm considers the edges in non-decreasing order of their weights and adds an edge whenever it connects two previously disconnected components. Edges that would create a cycle are discarded. The process stops once all vertices are connected, resulting in an $\MST$.

Throughout this section, we focus exclusively on connectivity and therefore on homology in degree $0$. For simplicity of notation, we omit the homological degree from the notation for persistent homology quantities, with the convention that all such quantities refer to degree $0$.

\begin{proposition}[$\MST$ characterization]
    \label{prop:mst}
    Let $\MST(X)$ be a minimum spanning tree of the complete weighted graph on $X$, with edge
    weights $w(\{x_i,x_j\})=\|x_i-x_j\|$. The finite $H_0$ death scales are exactly the
    weights of the $n-1$ $\MST$ edges. Therefore,
    \begin{equation}
        \TotPers(X)
        =
        \sum_{e\in\MST(X)}w(e)
        =:
        \cL(X).
        \label{eq:mst_identity}
    \end{equation}
\end{proposition}
\begin{proof}
    
    Let $B_0(X)=\{(0,d_i)\}_{i=1}^{n-1}$ denote the finite part of the $0$-dimensional persistence diagram of the Vietoris-Rips filtration. In dimension $0$, connected components of $\mathrm{VR}_\varepsilon(X)$ correspond precisely to the connected components of the graph $G_\varepsilon = (X, E_\varepsilon)$, where
    \[
    E_\varepsilon = \{\{x_i, x_j\} \subset X \mid \|x_i-x_j\| \le \varepsilon\}.
    \]
    
    A birth-death pair $(0, d) \in B_0(X)$ is created if and only if two previously disjoint connected components merge at scale $\varepsilon = d$. This merge is induced by the addition of an edge $\{x_i, x_j\}$ of length $\|x_i-x_j\|$. Edges whose addition creates cycles in $G_\varepsilon$ do not alter the number of connected components and thus contribute no $H_0$ persistence classes. 
    
    By construction, this process mirrors the execution of Kruskal's algorithm on $X$: edges are inspected in non-decreasing order of weight, and an edge is added to the spanning forest if and only if it connects two distinct connected components. Consequently, the scale values $\{d_1, \dots, d_{n-1}\}$ at which $H_0$ classes die are precisely the weights of the $n-1$ edges selected by Kruskal's algorithm to form $\MST(X)$.
    \[
    \TotPers(X)\;=\;\sum_{i=1}^{n-1}d_i=\sum_{e\in\MST(X)}w(e)\;=:\;\cL(X).
    \]
    When some pairwise distances are tied, the minimum spanning tree need not be unique, but the multiset of its edge weights is; the identity is therefore unaffected by the tie-breaking rule.
\end{proof}

\begin{proposition}[Stability]\label{prop:stability}
    Let $X$ and $X'$ be finite subsets of $\mathbb{R}^d$ with $|X|=|X'|=n$. Then the following properties hold:
    
    \begin{enumerate}
        \item[(a)]
        As a functional of the persistence diagram, $\TotPers$ is $W_1$-Lipschitz (for an admissible bijection all but finitely many terms of the sum vanish). More precisely,
        \[
        \left|\TotPers(X)-\TotPers(X')\right|
        \leq
        W_1(\mathrm{Dgm}(X),\mathrm{Dgm}(X'))
        \]
        \item[(b)]
        As a functional of the point cloud, for any bijection
        $\phi:X\rightarrow X'$,
        \[
        \left|\cL(X)-\cL(X')\right|
        \leq
        2(n-1)
        \max_i\|x_i-\phi(x_i)\|.
        \]
        The constant grows linearly in $n$ and is attained at $n=2$; we make no claim of sharpness for larger $n$. The bound is therefore per-point and gives no control uniform in the number of points.
    \end{enumerate}
\end{proposition}

\begin{proof}
    (a) We prove that $\TotPers$ is $W_1$-Lipschitz. Let
    \[
    \gamma:\mathrm{Dgm}(X)\cup\mathrm{Diag}\longrightarrow \mathrm{Dgm}(X')\cup\mathrm{Diag}
    \]
    be any admissible bijection, where $\mathrm{Diag}=\{(t,t):t\ge 0\}\subset\mathbb{R}^2$ denotes the
    diagonal of the persistence plane. For a persistence point
    $p=(b,d)$, define its persistence by $\pers(p):=d-b$.
    
    For every matched pair $p=(b,d)$ and $q=(b',d')$, we have
    \[
    |\pers(p)-\pers(q)| =|(d-d')-(b-b')|\le |d-d'|+|b-b'|=\|(b-b',d-d')\|_1=\|p-q\|_1.
    \]

    Therefore,
    \[
    \begin{aligned}
        |\TotPers(X)-\TotPers(X')|
        &=
        \left|
        \sum_{p\in \mathrm{Dgm}(X)\cup\mathrm{Diag}}
        \bigl(\pers(p)-\pers(\gamma(p))\bigr)
        \right|\\
        &\le
        \sum_{p\in \mathrm{Dgm}(X)\cup\mathrm{Diag}}
        |\pers(p)-\pers(\gamma(p))|\\
        &\le
        \sum_{p\in \mathrm{Dgm}(X)\cup\mathrm{Diag}}
        \|p-\gamma(p)\|_1.
    \end{aligned}
    \]
    Since $\gamma$ was arbitrary, taking the infimum over all admissible bijections gives
    \[
    |\TotPers(X)-\TotPers(X')|
    \le
    W_1\!\left(\mathrm{Dgm}(X),\mathrm{Dgm}(X')\right).
    \]
    
    (b) Let $\varepsilon=\max_i\|x_i-\phi(x_i)\|$. For every edge $e=\{x_i,x_j\}$ of $\MST(X)$, let
    $\phi(e)=\{\phi(x_i),\phi(x_j)\}$ be the corresponding edge in $X'$.
    By the triangle inequality,
    \[
    \begin{aligned}
        w(\phi(e))
        &=\|\phi(x_i)-\phi(x_j)\|\\
        &\le
        \|\phi(x_i)-x_i\|
        +\|x_i-x_j\|
        +\|x_j-\phi(x_j)\|\\
        &\le
        w(e)+2\varepsilon.
    \end{aligned}
    \]
    
    The collection of edges $\{\phi(e):e\in\MST(X)\}$ forms a spanning tree of $X'$.
    Since $\MST(X')$ minimizes the total weight among all spanning trees of $X'$,
    \[
    \begin{aligned}
        \cL(X')\le\sum_{e\in\MST(X)} w(\phi(e))\le\sum_{e\in\MST(X)}\bigl(w(e)+2\varepsilon\bigr)=
        \cL(X)+2(n-1)\varepsilon,
    \end{aligned}
    \]
    because every spanning tree has exactly $n-1$ edges.
    
    Applying the same argument to the inverse bijection $\phi^{-1}:X'\to X$ yields
    \[
    \cL(X)\le \cL(X')+2(n-1)\varepsilon.
    \]
    Hence,
    \[
    |\cL(X)-\cL(X')|
    \le
    2(n-1)\varepsilon
    =
    2(n-1)\max_i\|x_i-\phi(x_i)\|.
    \]
    The constant is sharp: for $n=2$, moving both points apart by $\varepsilon$ each along the line
    through them lengthens the single edge by exactly $2\varepsilon=2(n-1)\varepsilon$.
\end{proof}

\begin{proposition}[Invariances]\label{prop:inv}
    For $t\in\R^d$ and $c>0$: $\cL(X+t)=\cL(X)$ and $\cL(cX)=c\,\cL(X)$.
\end{proposition}
\begin{proof}
    For every edge $e=\{x_i,x_j\}$ of $\MST(X)$, $t\in\mathbb{R}^d$ and $c>0$,
    \[
    \|(x_i+t)-(x_j+t)\|=\|x_i-x_j\|,\,\,\,\,\,\,\,\,
    \|cx_i-cx_j\|=c\|x_i-x_j\|,
    \]
    hence $\cL(X+t)=\cL(X)$ and $\cL(c\,X)=c\,\cL(X)$.
\end{proof}

\section{Derivation of the Discriminability Reduction}
\label{app:ln}
This appendix expands the algebra compressed in Sec.~\ref{sec:ln}, in the notation established
there.

We work at a fixed layer $l$ in the notation of Sec.~\ref{sec:setup}, writing
$\tilde a_{l}(x_{i}^{c})$ for the normalized activation of the $i$-th example under class label
$c\in\{+,-\}$ and $\tilde\mu_{l}^{c}$ for the corresponding class mean. We define the residual of each sample with respect to the mean of its class as
\[
\varepsilon_{l}(x_{i}^{c})
=\tilde{a}_{l}(x_{i}^{c})-\tilde{\mu}_{l}^{c},
\]
where $c\in \{+,-\}$ and $\tilde{\mu}_{l}^{c}$ for class $c$. By construction,
\[
\sum_{i=1}^{N}\varepsilon_{l}(x_{i}^{c})=0.
\]

Instead of working with the $2N$ normalized activations, we consider the $N$ paired differences. Specifically, we define
\[
\delta_{l}(x_{i})
=\tilde{a}_{l}(x_{i}^{+})-\tilde{a}_{l}(x_{i}^{-})
=(\tilde{\mu}_{l}^{+}-\tilde{\mu}_{l}^{-})
+\bigl(\varepsilon_{l}(x_{i}^{+})-\varepsilon_{l}(x_{i}^{-})\bigr)
:=\mu_{\delta, l} +\xi_{l}^{i},
\]
where $\mu_{\delta, l}:=\tilde{\mu}_{l}^{+}-\tilde{\mu}_{l}^{-}$ and $\xi_{l}^{i}$ denote the differences between the class means and between the class residuals, respectively.

We denote the centered residual difference $\xi_{l}^{i}$, and let $\Xi=\{\xi_{l}^{i}\}_{i=1}^{N}$ be the resulting cloud of residual differences. Since $\sum_i\varepsilon_l(x_i^{c})=0$ for each $c$, we have
$\sum_{i=1}^{N}\xi_l^{i}=0$: the residual configuration $\Xi$ is precisely the displacement cloud $\Delta$ with its mean removed, so $\cL(\Delta)=\cL(\Xi)$ by Proposition~\ref{prop:inv}.

To study the effect of varying the residual dispersion while keeping the class means fixed, we define the family directly on the standardized displacement cloud,
    \[
    \Delta(\sigma):=\{\mu_{\delta,l}+\sigma\xi_l^{i}\}_{i=1}^{N},
    \qquad \sigma>0,
    \]
    realized by the representative activations
    \[
    \tilde a_{l}(\sigma)(x_{i}^{c}):=\tilde\mu_{l}^{c}+\sigma\,\varepsilon_{l}(x_{i}^{c}),
    \qquad c\in\{+,-\},
    \]
    which keep the class means fixed and reproduce $\Delta(\sigma)$ exactly, since
    $\tilde a_{l}(\sigma)(x_{i}^{+})-\tilde a_{l}(\sigma)(x_{i}^{-})=\mu_{\delta,l}+\sigma\xi_{l}^{i}$.
    The standardization \eqref{eq:zscore} is held fixed along the family --- it is estimated once from
    the layer's own activations --- so that $\Sb$ is unchanged and $\Sw(\sigma)=\sigma^{2}\Sw$;
    Remark~\ref{rem:restd} treats the case in which it is re-estimated.

LayerNavigator's discriminability is
\[
D_{l}(\sigma)
=
\frac{\mu_{\delta,l}^\top S_{l}^{b}\mu_{\delta,l}}
{\mu_{\delta,l}^\top\left(S_{l}^{b}+S_{l}^{w}(\sigma)\right)\mu_{\delta,l}},
\]
where $S_l^{b}$ and $S_l^{w}$ denote the between-class and within-class covariance matrices, respectively,
\begin{align}
    S_l^{b}
    &=
    N
    \sum_{c\in\{+,-\}}
    \left(
    \tilde{\mu}_{l}^{c}
    -
    \tilde{\mu}_{l}
    \right)
    \left(
    \tilde{\mu}_{l}^{c}
    -
    \tilde{\mu}_{l}
    \right)^\top,
    \label{eq:sb}
    \\
    S_l^{w}(\sigma)
    &=
    \sum_{c\in\{+,-\}}
    \sum_{i=1}^{N}
    \left(
    \tilde{a}_{l}(\sigma)(x_{i}^{c})
    -
    \tilde{\mu}_{l}^{c}
    \right)
    \left(
    \tilde{a}_{l}(\sigma)(x_{i}^{c})
    -
    \tilde{\mu}_{l}^{c}
    \right)^\top.
    \label{eq:sw}
\end{align}

Using the notation introduced above, the expressions in \eqref{eq:sb} and \eqref{eq:sw} simplify to
\begin{align}
    S_l^{b}
    &=
    \frac{N}{2}
    \mu_{\delta,l} 
    (\mu_{\delta,l} )^\top
    \\
    S_l^{w}(\sigma)
    &=
    \sum_{c\in\{+,-\}}
    \sum_{i=1}^{N}
    \sigma
    \varepsilon_{l}(x_{i}^{c})
    \left(
    \sigma
    \varepsilon_{l}(x_{i}^{c})
    \right)^\top
    =
    \sigma^{2}S_l^{w}.
\end{align}

We define $B_l(v)=v^\top S_{l}^b\,v$ and $W_{l}(v)= v^\top S_{l}^w\,v$ such that
\begin{align}
    B_l
    &:=
    \mu_{\delta,l}^\top S_l^{b}\mu_{\delta,l}
    =
    \frac{N}{2}
    \|\mu_{\delta,l}\|^{4},
    \\
    W_l
    &:=
    \mu_{\delta,l}^\top S_l^{w}\mu_{\delta,l} = \sum_{c\in\{+,-\}} \sum_{i=1}^N \left( \mu_
    {\delta,l}^\top \varepsilon_l(x_i^c) \right)^2.
\end{align}

The within-class term $W_l$ is close to, but not equal to, the spread of the displacement cloud along the same direction, and the gap is exactly the paired covariance.

\begin{lemma}[What $W_l$ and the displacement spread differ by]\label{lem:paircov}
    With $a_i:=\mu_{\delta,l}^{\top}\varepsilon_l(x_i^{+})$ and
    $b_i:=\mu_{\delta,l}^{\top}\varepsilon_l(x_i^{-})$,
    \[
    \sum_{i=1}^{N}\bigl(\mu_{\delta,l}^{\top}\xi_l^{i}\bigr)^{2}
    \;=\;
    W_l\;-\;2\sum_{i=1}^{N}a_ib_i .
    \]
\end{lemma}
\begin{proof}
    $\mu_{\delta,l}^{\top}\xi_l^{i}=a_i-b_i$, so
    $\sum_i(a_i-b_i)^{2}=\sum_i a_i^{2}+\sum_i b_i^{2}-2\sum_i a_ib_i$, and the first two sums are
    $W_l$ by its definition above.
\end{proof}

Variation shared by the two members of a pair --- prompt template, topic, formatting, therefore enters $W_l$ but cancels in $\xi_l^{i}$. This is the structural reason the two criteria can come apart, and it is what the shared term $c_i$ isolates in the construction of
Appendix~\ref{app:toy}.

Substituting these expressions into the definition of the LayerNavigator's discriminability yields
\[
D_l(\sigma)
=
\frac{B_l}
{B_l+\sigma^{2}W_l}.
\]

\subsection*{Proof of Theorem~\ref{thm:decoupling}}
\begin{proof}
    (i) Applying Proposition~\ref{prop:mst} together with Proposition~\ref{prop:inv}, we obtain
    \[
    \TotPers\bigl(\Delta_{l}(\sigma)\bigr)
    =
    \mathcal{L}\bigl(\Delta_{l}(\sigma)\bigr)
    =
    \mathcal{L}\bigl(\{\sigma\xi_{l}^{i}\}_{i=1}^{N}\bigr)
    =
    \mathcal{L}(\sigma\Xi)
    =
    \sigma\mathcal{L}(\Xi).
    \]
    Therefore, $\TotPers(\Delta_{l}(\sigma))$ is strictly increasing since
    \[
    \frac{\partial\,\TotPers(\Delta_{l}(\sigma))}{\partial\sigma}
    =
    \mathcal{L}(\Xi)
    >
    0.
    \]
    
    (ii) Differentiating the discriminability operator $D_{l}(\sigma)$ yields
    \[
    \frac{\partial D_{l}(\sigma)}{\partial\sigma}
    =
    \frac{-2B_l\sigma W_l}{\left(B_l+\sigma^{2}W_l\right)^{2}}
    <
    0.
    \]
    where the strict inequality follows from the assumptions
    $B_l,W_l>0$ and $\sigma>0$. Hence, $D_{l}(\sigma)$ is strictly decreasing.
\end{proof}

\begin{remark}[The family under re-standardization]\label{rem:restd}
    If instead the family is generated at activation level and \eqref{eq:zscore} is re-applied, the coordinate-wise pooled variance becomes
    $\bigl(\mu^{\mathrm{raw}}_{\delta,k}\bigr)^{2}/4+\sigma^{2}s_k^{2}$ with $s_k:=\bar W_{l,k}^{1/2}$ the pooled within-class standard deviation of Proposition~\ref{prop:sepscale}, and the standardized displacements differ by
    \[
    \bigl(\delta^{i}(\sigma)-\delta^{j}(\sigma)\bigr)_k
    =
    \frac{\sigma\,(\xi^{\mathrm{raw},i}_k-\xi^{\mathrm{raw},j}_k)}
    {\sqrt{(\mu^{\mathrm{raw}}_{\delta,k})^{2}/4+\sigma^{2}s_k^{2}}} ,
    \]
    the mean displacement cancelling, where $\xi^{\mathrm{raw},i}$ is the raw residual displacement. Writing $m_k:=\mu^{\mathrm{raw}}_{\delta,k}/2$, the coordinate
    factor $g_k(\sigma)=\sigma(m_k^{2}+\sigma^{2}s_k^{2})^{-1/2}$ has
    \[
    g_k'(\sigma)=\frac{m_k^{2}}{\bigl(m_k^{2}+\sigma^{2}s_k^{2}\bigr)^{3/2}}\;\ge\;0 ,
    \]
    with equality exactly on the coordinates where $\mu^{\mathrm{raw}}_{\delta,k}=0$, on which $g_k\equiv
    1/s_k$. Every pairwise distance is therefore non-decreasing in $\sigma$, and since $\cL$ is non-decreasing in each entry of the distance matrix, so is $\TotPers$; it increases strictly as soon
    as every pair of residual displacements differs on some coordinate with
    $\mu^{\mathrm{raw}}_{\delta,k}\neq0$, which is the generic case. It is now bounded, with
    $\TotPers\to\cL(H_l)$ as $\sigma\to\infty$, the separation-free score of Proposition~\ref{prop:sepscale}, so the linear law
    $\TotPers(\Delta(\sigma))=\sigma\cL(\Xi_l)$ holds only for the family of the theorem.

    For $D_l$ the situation is different, because it is a ratio of quadratic forms and is therefore unchanged by a common rescaling of all coordinates. Whenever the re-standardization acts as a single
    scalar, $m_k^{2}+\sigma^{2}s_k^{2}$ independent of $k$, as in the isotropic model of Appendix~\ref{sec_mst_vs_total_dist}, the factor cancels from numerator and denominator and the reduction $D_l(\sigma)=B_l/(B_l+\sigma^{2}W_l)$ continues to hold \emph{exactly}, so Theorem~\ref{thm:decoupling} applies verbatim. When the rescaling is coordinate-dependent, it does not cancel; $D_l$ is no longer that ratio, and we do not prove monotonicity; in simulation it
    remains strictly decreasing. In either case the orderings remain opposite. This is the same coupling as Proposition~\ref{prop:sepscale}.
\end{remark}

\section{Formal Analysis of the Answer-Polarity Axis}
\label{app:anatomy}
This appendix expands and provides a formal analysis underlying Sec.~\ref{sec:polarity}.

\paragraph{Structural decomposition.}
In the persona format, each contrastive pair shares the same prompt--a statement describing the target behavior together with a question--and differs only in the appended response: the matching answer for $x^+$ and the non-matching answer for $x^-$. Motivated by this construction, we model each activation difference as the sum of a context term, a response-token term, and a behavior term:
\begin{equation}
    \delta_l^i
    \;=\;
    \underbrace{(c_l^i-c_l^i)}_{=\,0}
    \;+\;
    \underbrace{\pm\,t_l}_{\text{answer polarity}}
    \;+\;
    \underbrace{b_l^i}_{\text{behavior}} .
    \label{eq:anatomy}
\end{equation}
~\eqref{eq:anatomy} is an ansatz suggested by the contrastive construction, rather than a derived identity. All results below are therefore conditional on this decomposition. The sign of $t_l$ records which response token carries the matching answer for example $i$; it is thus a property of how the dataset is written, rather than of the behavior being steered.
We estimate the axis by the half-difference of the answer-conditioned displacement means,
\begin{equation}
    t_l := \tfrac12\bigl(\bar\delta_l^{\,\mathrm{yes}}-\bar\delta_l^{\,\mathrm{no}}\bigr),
    \qquad
    \hat t_l := t_l/\lVert t_l\rVert ,
    \label{eq:polarity_axis}
\end{equation}
Two assumptions enter here. The $\pm t_l$ terms cancel in $\mu_{\delta,l}$ only if the
\emph{Yes}/\emph{No} split is exactly balanced, which the dataset construction
of~\citep{sun2025layernavigator} enforces; and \eqref{eq:polarity_axis} is unbiased for $t_l$ only if the
behavior terms $b_l^i$ have equal means across the two answer groups.
For a balanced dataset, the mean displacement is therefore
\begin{equation}
    \mu_{\delta,l}
     =
    \frac{1}{N}\sum_{i=1}^{N}\delta_l^i
    =
    \frac{1}{N}\sum_{i=1}^{N}
    \left(\pm t_l+b_l^i\right)
    =
    \mu_{b,l}.
    \label{eq:displacement_mean}
\end{equation}
Thus, the answer-polarity component does not affect the mean displacement and consequently does not affect mean-displacement (MD) extraction.

We can remove this component using the orthogonal projector
\begin{equation}
P_{\hat{t}_l} = I - \hat{t}_l \hat{t}_l^\top,
\label{eq:polarity_projector}
\end{equation}
We then use this projection to study how removing the answer-polarity component affects the two modes considered below: discriminability, through the resulting change in the steering direction and its associated scatter, and connectivity, through the geometric change in the displacement cloud and its minimum spanning tree. The two effects are analyzed separately in the following sections

\paragraph{Discriminability under projection}

We now analyze how the discriminability score $D_l$ changes when the
answer-polarity axis is projected out. Recall that
\begin{equation}
D_l(v)
=
\frac{B_l(v)}
{B_l(v)+W_l(v)},
\qquad
B_l(v)=v^\top S_l^b v,
\qquad
W_l(v)=v^\top S_l^w v.
\end{equation}

Since $B_l$ and $W_l$ are quadratic forms, $D_l$ is invariant to
rescaling of its direction argument:
\begin{equation}
D_l(cv)=D_l(v),\qquad c\neq 0.
\end{equation}
We therefore evaluate $D_l$ on unit directions without loss of
generality.

Let
\begin{equation}
v=\mu_{\delta,l},
\qquad
v'=P_{\hat t_l}v,
\qquad
\hat v=\frac{v}{\|v\|},
\qquad
\hat v'=\frac{v'}{\|v'\|}.
\end{equation}

\begin{lemma}[Relation between a vector and its projection]\label{lm:relation_vector}
        Let $v$ be a nonzero vector and
				$v'=P_{\hat {t}_{l}}v$. Define $\varepsilon:=\langle\hat v,\hat {t}_{l}\rangle$ and assume that $|\varepsilon|<1$, Then,
                \begin{align}
                   \hat v=\varepsilon\hat {t}_{l}+\sqrt{1-\varepsilon^{2}}\,\hat v' 
                \end{align}
        \end{lemma}
        \begin{proof}
            $P_{\hat {t}_{l}}\hat v=\hat v-\varepsilon\hat {t}_{l}\neq0$ and
			$\lVert\hat v-\varepsilon\hat {t}_{l}\rVert^{2}=1-2\varepsilon^{2}+\varepsilon^{2}=1-\varepsilon^{2}$, so
			\begin{equation}
				\hat v'=\frac{\hat v-\varepsilon\hat {t}_{l}}{\sqrt{1-\varepsilon^{2}}},
				\qquad\text{equivalently}\qquad
				\hat v=\varepsilon\hat {t}_{l}+\sqrt{1-\varepsilon^{2}}\,\hat v' .
				\label{eq:frame}
			\end{equation}
			The pair $(\hat {t}_{l},\hat v')$ is orthonormal, because
			$\langle\hat {t}_{l},\hat v'\rangle=(\langle\hat {t}_{l},\hat v\rangle-\varepsilon)/\sqrt{1-\varepsilon^{2}}=0$.
			Note also $\langle\hat v,\hat v'\rangle=\sqrt{1-\varepsilon^{2}}$.
        \end{proof}

The key observation is that projecting the data does not change either the between-class or within-class scatter matrix, $S_{l}^{b}$ and $S_{l}^{w}$, when they are evaluated along $v'$. This gives the following identity.
        \begin{lemma}[Projection identity]\label{lem:projid}
			Let $D_{l}$ the discriminability and  $P_{\hat{t}_{l}}$ the orthogonal projector onto the subspace orthogonal to $\hat{t}_{l}$. For any direction $v$, let $v'=P_{\hat{t}_{l}}v$. Evaluating discriminability after projecting the activations along $v'$
 gives the same value as evaluating the original discriminability along $v'$. Hence, the shift $D_{l}(v')-D_{l}(v)$ is
			therefore a change of direction, not of data.
		\end{lemma}
		\begin{proof}
        Consider either the between-class or within-class scatter matrix $S\in \{S_{l}^{b}, S_{l}^{w}\}$, which are formed from the activations. Applying the projection $P_{\hat{t}_{l}}$ to the activations, the corresponding scatter matrix becomes $P_{\hat{t}_{l}}SP_{\hat{t}_{l}}$. Therefore, evaluating along $v'$ and applying $P_{\hat{t}_l}v' = v'$,
			\[
			(v')^{\top}(P_{\hat{t}_{l}}SP_{\hat{t}_{l}})(v')
			=
			(P_{\hat{t}_{l}}v')^{\top}S\,(P_{\hat{t}_{l}}v')
			=
			(v')^{\top}S\,v' .
			\]
			 Therefore, the projection does not require redefining the discriminability criterion on a new displacement cloud: its effect can be studied entirely as a change in steering direction from $v$ to $v'$.
		\end{proof}

        We now specialize these observations to the steering direction $v=\mu_{\delta,l}$.
For notational convenience, and at variance with Appendix~\ref{app:ln}, where the same symbols are evaluated at $\mu_{\delta,l}$ rather than at its unit vector (the two differ by the factor $\lVert\mu_{\delta,l}\rVert^{2}$, which cancels in $D_l$), let
\begin{equation}
B_l:=B_l(\hat v),
\qquad
W_l:=W_l(\hat v),
\qquad
B_l':=B_l(\hat v'),
\qquad
W_l':=W_l(\hat v').
\end{equation}
Similarly, write
\begin{equation}
D_l:=D_l(\hat v),
\qquad
D_l':=D_l(\hat v').
\end{equation}
		
		\begin{proposition}[Discriminability under projection]\label{prop:leak}
				Let
\[
v=\mu_{\delta,l},
\qquad
v'=P_{\hat t_l}v,
\]
and define
\[
\varepsilon:=\langle\hat v,\hat t_l\rangle,
\qquad
\lambda:=1-\frac{W_l'}{W_l}.
\]
Assume $B_l>0$, $W_l'>0$, and $|\varepsilon|<1$. Then:
				\begin{enumerate}
					\item[(i)] $B'_{l}=(1-\varepsilon^{2})\,B_{l}$.
					\item[(ii)] The shift in discriminability is
					\[
					D'_{l}-D_{l}
					=
					\frac{B_{l}\,W_{l}\,\bigl(\lambda-\varepsilon^{2}\bigr)}
					{(B_{l}+W_{l})\bigl[(1-\varepsilon^{2})B_{l}+(1-\lambda)W_{l}\bigr]} ,
					\]
					and is therefore positive if and only if $\lambda>\varepsilon^{2}$.
					\item[(iii)] $\lambda$ is not free: it is determined by $\varepsilon$ and $S^{w}_{l} $ through
					\[
					\lambda\,W_{l}
					\;=\;
					\varepsilon^{2}\bigl(W_{l}(\hat {t}_{l})-W'_{l}\bigr)
					\;+\;
					2\varepsilon\sqrt{1-\varepsilon^{2}}\;\hat {t}^{\top}_{l} S_{l}^{w}\,\hat v' .
					\]
					In particular, if $\varepsilon=0$ then $\hat v'=\hat v$, $\lambda=0$, and the shift vanishes
					identically.
					\item[(iv)] If $0<|\varepsilon|$ and $\hat{t}_{l}$ is an eigenvector of $S_{l}^{w}$, the cross term vanishes and, with
					$r:=W(\hat {t}_{l})/W'_{l}$,
					\[
					\lambda=\frac{\varepsilon^{2}(r-1)}{\varepsilon^{2}(r-1)+1},
					\qquad\text{and}\qquad
					D'_{l}-D_{l}>0
					\iff
					r>1+\frac{1}{1-\varepsilon^{2}} .
					\]
					As $\varepsilon\to0$ at fixed $r$,
					\[
					D'_{l}-D_{l}\;=\;\frac{B_{l}\,W_{l}\,(r-2)}{(B_{l}+W_{l})^{2}}\;\varepsilon^{2}\;+\;O(\varepsilon^{4}).
					\]
				\end{enumerate}
			\end{proposition}
		
		\begin{proof}
			\emph{(i).}
			By Appendix~\ref{app:ln}, $\Sb=\tfrac N2\mu_{\delta,l}\mu_{\delta,l}^{\top}= \tfrac N2 v\,v^{\top}$, so for a unit vector $w$
			\[
			B(w)=\tfrac N2\langle v,w\rangle^{2}
			=\tfrac N2\lVert v\rVert^{2}\langle\hat v,w\rangle^{2} .
			\]
			Taking $w=\hat v$ gives $B(\hat{v})=\tfrac N2\lVert v \rVert^{2}$, and taking
			$w=\hat v'$ with $\langle\hat v,\hat v'\rangle=\sqrt{1-\varepsilon^{2}}$ by Lemma \ref{lm:relation_vector} gives
				$B(\hat v')=(1-\varepsilon^{2})B_l$. This is an identity, not an approximation.
			
			\emph{(ii).}
			We have that  
			\[
			D'_{l}-D_{l}
			=\frac{B'_{l}W_{l}-B_{l}W'_{l}}{(B'_{l}+W'_{l})(B_{l}+W_{l})} .
			\]
			Substituting $B'_{l}=(1-\varepsilon^{2})B_{l}$ from (i) and $W'_{l}=(1-\lambda)W_{l}$ from the definition of
			$\lambda$, gives the stated expression. The denominator $(B'_{l}+W'_{l})(B_{l}+W_{l})$ is strictly positive under the
			standing assumptions, so the sign of the shift is the sign of $\lambda-\varepsilon^{2}$.
			
			\emph{(iii).}
			Expanding $W_{l}$ in the frame~\eqref{eq:frame} from Lemma \ref{lm:relation_vector},
			\[
			W_{l}
			=\varepsilon^{2}W(\hat {t}_{l})
			+2\varepsilon\sqrt{1-\varepsilon^{2}}\,\hat {t}_{l}^{\top}\Sw\hat v'
			+(1-\varepsilon^{2})W'_{l} .
			\]
			Subtracting $W'_{l}$ from both sides and using $\lambda W_{l}=W_{l}-W'_{l}$ yields the
			identity. If $\varepsilon=0$ then $\hat v'=\hat v$ by~\eqref{eq:frame}, so $\lambda=0$ and, by (ii),
			$D'_{l}=D_{l}$.
			
			\emph{(iv).}
			If $\Sw\hat {t}_{l}=W(\hat {t}_{l})\hat {t}_{l}$ then $\hat {t}_{l}^{\top}\Sw\hat v'=W(\hat {t}_{l})\langle\hat {t}_{l},\hat v'\rangle=0$,
			so the expansion in (iii) reduces to
			\[
			W_{l}=\varepsilon^{2}W(\hat {t}_{l})+(1-\varepsilon^{2})W'_{l},
			\qquad
			\lambda W_{l}=\varepsilon^{2}\bigl(W(\hat {t}_{l})-W'_{l}\bigr).
			\]
			Dividing both by $W'_{l}>0$ and writing $r=W(\hat {t}_{l})/W'_{l}$,
			\[
			\lambda
			=\frac{\varepsilon^{2}(r-1)}{\varepsilon^{2}r+1-\varepsilon^{2}}
			=\frac{\varepsilon^{2}(r-1)}{\varepsilon^{2}(r-1)+1} .
			\]
			Put $s:=\varepsilon^{2}(r-1)$, so $\lambda=s/(s+1)$ with $s>-1$. By (ii) the shift is positive iff
			$\lambda>\varepsilon^{2}$, i.e.\ iff $s>\varepsilon^{2}(s+1)$, i.e.\ iff
			$s(1-\varepsilon^{2})>\varepsilon^{2}$. Substituting $s$ and dividing by $\varepsilon^{2}>0$ gives
			$(r-1)(1-\varepsilon^{2})>1$, that is $r>1+(1-\varepsilon^{2})^{-1}$. Finally, substituting
			$\lambda=\varepsilon^{2}(r-1)+O(\varepsilon^{4})$ into the expression of (ii) and letting
			$\varepsilon\to0$ gives $\lambda-\varepsilon^{2}=\varepsilon^{2}(r-2)+O(\varepsilon^{4})$ and
			denominator $(B_{l}+W_{l})^{2}+O(\varepsilon^{2})$, hence the stated expansion.
		\end{proof}

        \paragraph{Connectivity under projection}

        We now analyze how projecting out the answer-polarity axis affects the connectivity of the displacement cloud. In contrast to discriminability, which depends on quadratic forms evaluated along a steering direction, connectivity depends directly on the Euclidean lengths of the edges of a minimum spanning tree ($\MST$) (Appendix \ref{app:mst}). Consequently, the effect of the projection is geometric: an edge loses precisely the component of its length along the projected-out direction.

        In our setting, we apply the projection with $u=\hat t_l$ to the displacement cloud $\Delta_{l}$, which gives the polarity-projected cloud
		\begin{equation}
			\Delta_l' := P_{\hat t_l}\Delta_l
			= \bigl\{\delta_l^i-\langle \delta_l^i,\hat t_l\rangle\,\hat t_l\bigr\}_{i=1}^{N}.
			\label{eq:projected_cloud}
		\end{equation}

        \begin{proposition}[Connectivity under projection]\label{prop:polinv}
				Let $u$ be a unit vector, $P_u:=I-uu^{\top}$ the orthogonal projector onto $u^{\perp}$, and $X\subset\mathbb{R}^d$ a point cloud of $N$ points. Then:
				\begin{enumerate}
					\item[(i)] $\cL(P_uX)\le\cL(X)$.
					\item[(ii)] Let $T'=\MST(P_uX)$ and lift each of its edges to the corresponding pair of
					points of $X$. Then
					\[
					\cL(X)-\cL(P_uX)\;\le\;\sum_{e\in T'}\frac{\langle e,u\rangle^{2}}{\lVert e\rVert}.
					\]
					In particular, if every lifted edge satisfies $\lvert\langle e,u\rangle\rvert\le\eta$ and
					$\lVert e\rVert\ge w_{\min}$, then $\cL(X)-\cL(P_uX)\le (N-1)\eta^{2}/w_{\min}$.
				\end{enumerate}
			\end{proposition}
            \begin{proof}
			(i) $P_u$ is $1$-Lipschitz and $P_u\MST(X)$ is a connected spanning subgraph of $P_uX$
			(possibly with coincident vertices), so
			\[
			\cL(P_uX)\;\le\;\sum_{e\in\MST(X)}\lVert P_ue\rVert\;\le\;\cL(X).
			\]
			
			(ii) The lift of $T'=\MST(P_uX)$ is a spanning tree of $X$, hence
			$\sum_{e\in T'}\lVert e\rVert\ge\cL(X)$. For each edge,
			$\lVert P_ue\rVert=\sqrt{\lVert e\rVert^{2}-\langle e,u\rangle^{2}}
			\ge\lVert e\rVert-\langle e,u\rangle^{2}/\lVert e\rVert$, so
			\[
			\cL(P_uX)=\sum_{e\in T'}\lVert P_ue\rVert
			\;\ge\;\sum_{e\in T'}\lVert e\rVert-\sum_{e\in T'}\frac{\langle e,u\rangle^{2}}{\lVert e\rVert}
			\;\ge\;\cL(X)-\sum_{e\in T'}\frac{\langle e,u\rangle^{2}}{\lVert e\rVert}.
			\]
			The final clause follows by bounding each of the $N-1$ summands by $\eta^{2}/w_{\min}$.
		\end{proof}
		
		\paragraph{Comparison between two criteria}
			The two propositions capture different effects of removing the answer-polarity axis. For connectivity, the projection acts directly on Euclidean distances and therefore on the MST. In a two-lobe cloud, the resulting change has two contributions: removing the component of the bridge along $\hat{t}_{l}$, and superimposing the two lobes after projection. The latter changes the MST from spanning two sets of approximately $N/2$ points to spanning $N$ points in the residual space.

            Discriminability responds differently because it depends on quadratic forms evaluated along the steering direction. By Proposition~\ref{prop:leak}, the effect of the projection is controlled by $\varepsilon=\left\langle\hat{\mu}_{\delta,l},\hat{t}_{l}\right\rangle.$ It vanishes when the answer-polarity axis is orthogonal to the mean-displacement direction, and becomes positive only when $\hat {t}_{l}$ carries more than about twice the within-class scatter of the residual steering direction. Thus, connectivity and discriminability need not change by comparable amounts under the same projection: the former is sensitive to the geometry of the entire cloud, whereas the latter is sensitive to how the projected direction interacts with the between- and within-class scatter.

		\begin{figure}[h]
			\centering
			\includegraphics[width=\textwidth]{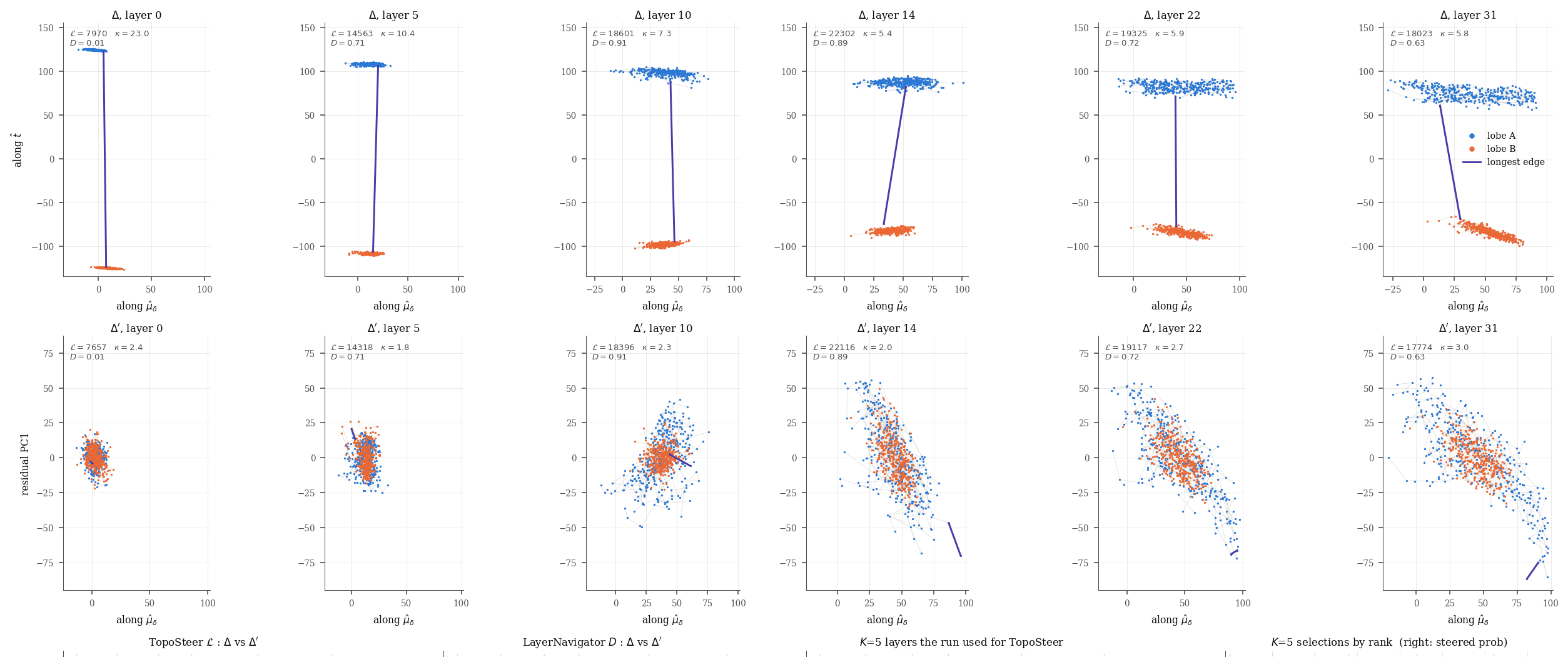}
			\caption{\textbf{Anatomy of a displacement cloud} (conscientiousness, Llama-8B, $700$ pairs).
				\emph{Top:} $\Delta_l$ at six depths, in the plane spanned by the mean displacement
				$\hat\mu_{\delta,l}$ and the answer-polarity axis $\hat {t}_{l}$; the cloud splits into two lobes and the
				longest spanning-tree edge (dark) bridges them. The lobes are point-like at layer $0$ and swell
				with depth, which is where $\cL$ grows.
				\emph{Bottom:} the same clouds after removing $\hat {t}_{l}$, plotted against the leading residual
				principal component; the residual behavior geometry is what survives.
				Panel annotations give $\cL$ and LayerNavigator's $D$.}
			\label{fig:anatomy}
		\end{figure}

            \begin{remark}[Effect of Lobe Superposition]\label{rem:asym}
                As mentioned above, one of the effects of projecting out $\hat{t}_{l}$ in the connectivity case is the change from spanning two sets of $N/2$ points to spanning a single set of $N$ points. To quantify this effect, let $D_{\mathrm{int}}$ denote the \emph{intrinsic dimension} of a lobe, i.e., the effective number of dimensions in which its points exhibit substantial geometric variation, rather than the ambient dimension of the activation space. By the Beardwood--Halton--Hammersley
			scaling $\cL\asymp n^{1-1/D_{\mathrm{int}}}$, the second part is a relative change of order
			\[
			1-2^{-1/D_{\mathrm{int}}}\;\approx\;\frac{\log 2}{D_{\mathrm{int}}},
			\]
            Thus, this contribution is small for high-dimensional lobes but can become substantial when the intrinsic dimension is low, and in neither case controlled by the bridge share.
			Appendix~\ref{app:toy} verifies the scaling.
            \end{remark}

\section{Two-Dimensional Illustrations}
\label{app:toy}
This appendix gives the constructions behind Figures~\ref{fig:toy2d} and~\ref{fig:bridge2d}. 

\subsection{The decoupling theorem in 2D}
\begin{figure}[h]
    \centering
    \includegraphics[width=0.92\textwidth]{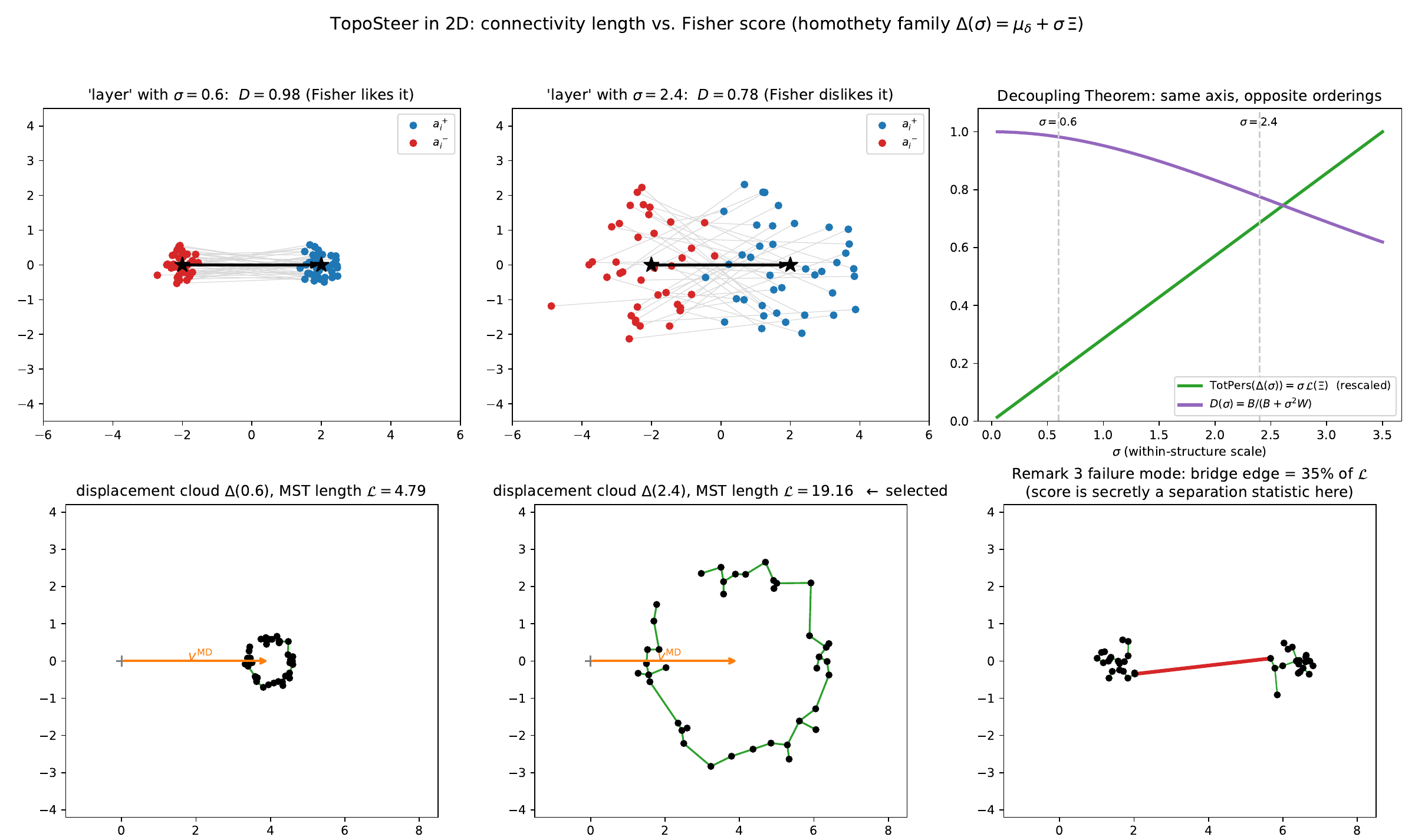}
    \caption{
        \textbf{Separability and displacement connectivity can induce opposite rankings.}
        The synthetic configurations share the class-mean difference $\mu_{\delta}=(4,0)$ and differ only
        in residual scale $\sigma$. Gray segments connect paired positive and negative activations. The
        displacement clouds keep the same mean direction while their connectivity length grows as
        $\TotPers(\Delta(\sigma))=\sigma\cL(\Xi)$, whereas discriminability decreases as
        $D(\sigma)=B/(B+\sigma^2W)$: the exact ordering reversal of Theorem~\ref{thm:decoupling}.
    }
    \label{fig:toy2d}
\end{figure}
Take $\mu_\delta=(4,0)$ and let $\Xi$ be $N=40$ points on a noisy unit ring, split into activations via $\tilde a_i^{\pm}=\pm\mu_\delta/2+\sigma(\pm\xi_i/2+c_i)$, where the shared pair noise $c_i$ inflates within-class scatter but cancels exactly in $\delta_i=\mu_\delta+\sigma\xi_i$. At $\sigma=0.6$ the classes look cleanly separated and Fisher rewards the layer ($D=0.98$) while the displacement MST is short ($\cL=4.8$); at $\sigma=2.4$ the classes visibly overlap, Fisher drops ($D=0.78$), yet the displacement ring has grown and $\cL=19.2$ --- \ourmethod{} prefers exactly the layer LayerNavigator penalizes (Figure~\ref{fig:toy2d}, left and middle columns). Sweeping $\sigma$ traces the two curves of Theorem~\ref{thm:decoupling}: $\TotPers$ is the exact line $\sigma\cL(\Xi)$ and $D=B/(B+\sigma^{2}W)$ decays, so the induced orderings are reversed at every pair of scales (right column). 

\subsection{Reading the MST of a displacement cloud}
\begin{figure}[t]
    \centering
    \includegraphics[width=\textwidth]{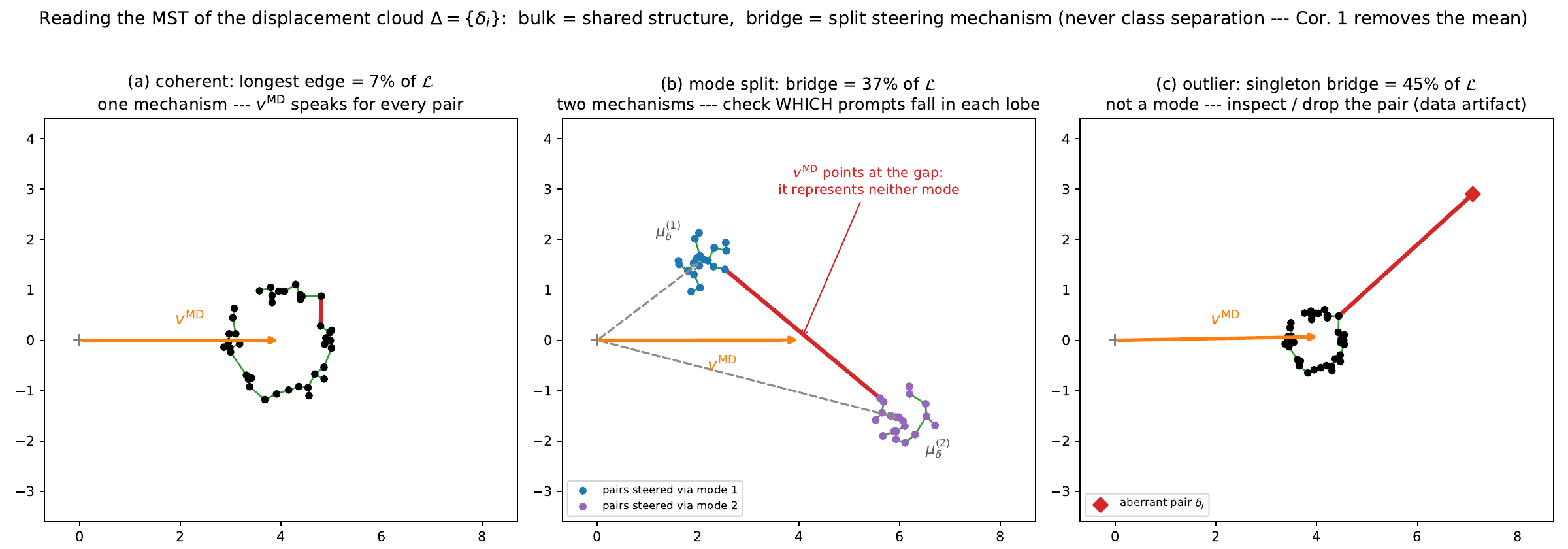}
    \caption{Reading the MST of the displacement cloud $\Delta$ (Example~\ref{ex:bridge}). (a) Coherent cloud: the longest edge is a small fraction of $\cL$ and $v^{\md}$ (orange) represents every pair. (b) Mode split: two subpopulations of pairs (colors) displace along different modes $\mu_\delta^{(1)},\mu_\delta^{(2)}$ (dashed); the bridge (red) carries a large share of $\cL$ and $v^{\md}$ points into the gap between the lobes, representing neither. (c) Outlier: the bridge joins a singleton --- a data artifact, not a mode. The bridge share~\eqref{eq:bridge} separates these regimes and must be reported at score-peak layers.}
    \label{fig:bridge2d}
\end{figure}

\begin{example}[what a bridge in $\Delta$ means]\label{ex:bridge}
    Because the cloud is the set of \emph{per-pair differences}, and Corollary~\ref{cor:inv} makes $\cL(\Delta)$ blind to its mean, a long bridge edge is \emph{never} the class separation $\lVert\mu_\delta\rVert$. It signals instead that the pairs split into subpopulations with distinct displacement modes $\mu_\delta^{(1)},\mu_\delta^{(2)}$: the model implements the behavioral contrast in (at least) two different ways depending on the example, and the bridge length is approximately the gap between the modes. This is bad news for mean-difference steering specifically: $v^{\md}$ averages the lobes and points into the empty region between them, representing neither mode --- a bridge-inflated score would select precisely the layers where the single-vector premise fails (Figure~\ref{fig:bridge2d}b). Two flavors must be distinguished by the sizes of the components the bridge joins: a bridge between two sizable clusters is a genuine mode split (inspect which prompts fall in each lobe --- answer polarity, phrasing templates, and topical subgroups are common culprits), whereas a bridge to a singleton is an outlier pair, i.e.\ a data or tokenization artifact to inspect or drop (Figure~\ref{fig:bridge2d}c). Finally, $Z$-scoring rescales coordinates by the total per-coordinate std, and the lobe-gap direction contributes within-class variance, so normalization can dampen or amplify a bridge; the bridge share must therefore be measured in the same normalization used for extraction and is cheap to report in both conventions. In this sense the bridge share is the topological counterpart of LayerNavigator's consistency score, but it is diagnostic rather than scalar: it identifies \emph{which} pairs disagree and by how much.
\end{example}

\subsection{The lobe-merging term of Remark~\ref{rem:asym}}
Remark~\ref{rem:asym} attributes part of the drop in $\cL$ under projection to the two lobes being superimposed rather than to the bridge. We check the predicted size. For each ambient dimension $D$, we draw two standard Gaussian lobes of $N/2=200$ points each, separated along $u$ by a gap large enough that the bridge is a single edge, remove the bridge from both sides of the comparison, and report the residual deficit as a fraction of the bulk length. The prediction is $1-2^{-1/D_{\mathrm{int}}}$ with $D_{\mathrm{int}}=D$. Table~\ref{tab:bhh} is produced by \texttt{bhh\_lobe\_merging.py}, provided in the supplement ($12$ trials per dimension, seed $11$).

\begin{table}[h]
    \centering
    \small
    \caption{Bulk deficit under projection, against the Beardwood--Halton--Hammersley prediction.}
    \label{tab:bhh}
    \begin{tabular}{rccc}
        \toprule
        $D$ & $1-2^{-1/D}$ & measured & s.d. \\
        \midrule
        2  & 0.293 & 0.279 & 0.014 \\
        3  & 0.206 & 0.193 & 0.011 \\
        5  & 0.129 & 0.123 & 0.008 \\
        10 & 0.067 & 0.070 & 0.005 \\
        20 & 0.034 & 0.042 & 0.002 \\
        \bottomrule
    \end{tabular}
\end{table}

The residual gap at $D=20$ is the finite-$N$ correction: the asymptotic scaling needs $N\gg 2^{D}$. The term is therefore negligible at the widths of the displacement clouds we score, and the bridge share~\eqref{eq:bridge} accounts for the rest.

\section{Why Mean-Difference Extraction}
\label{app:md}
We deliberately restrict attention to Mean-Difference (MD) extraction for three reasons. (i) \emph{Methodological alignment}: our layer-selection criterion is a diagnostic of precisely the premise underlying MD steering --- that a single vector, the mean of the displacement cloud, faithfully represents every pair's displacement (Corollary~\ref{cor:inv}, Example~\ref{ex:bridge}); probe- and subspace-based extractors rest on different premises and would require different diagnostics. (ii) \emph{Experimental control}: LayerNavigator \citep{sun2025layernavigator} likewise selects layers for MD-extracted vectors, so fixing the extractor isolates the variable under study --- layer selection --- from extraction effects. (iii) \emph{Standard practice}: MD is the many-pair generalization of activation addition, and the extractor used by contrastive activation addition is hyperparameter-free, and prior evidence indicates difference-in-means directions are causally effective for intervention relative to probe-derived directions \citep{marks2024geometry, arditi2024refusal}. We make no claim that MD is uniformly the best extractor; under a two-Gaussian model, the Bayes-optimal linear direction is the whitened difference $\Sigma^{-1}\mu_\delta$, and extending the topological diagnostic to whitened, subspace-, and probe-based extractors is left to future work.

\section{Computational Properties}
\label{sec:toposteer_complexity}
\ourmethod introduces no additional LLMs forward passes beyond those needed to extract the layerwise MD steering vectors. A forward pass for each contrastive input yields activations from all layers simultaneously, and these activations are cached for both vector extraction and layer scoring.

After activation extraction, coordinate-wise normalization and displacement construction require
\begin{equation}
    O(LNd)
\end{equation}
arithmetic operations. A dense pairwise-distance implementation requires
\begin{equation}
    O(LN^2d)
\end{equation}
time to construct all layerwise distance matrices. Using a dense MST routine such as Prim's algorithm adds
\begin{equation}
    O(LN^2)
\end{equation}
time. When layers are processed sequentially, the additional working memory is
\begin{equation}
    O(Nd+N^2),
\end{equation}
excluding the cached model activations. Sorting the $L$ layer scores requires
$O(L\log L)$ time.

The dominant cost is therefore an offline geometric computation over the extraction data, rather than additional LLMs inference. Once the ranking has been computed, deployment requires neither persistent-homology computation nor an auxiliary selector model for each new input. The only additional inference-time operation is the vector addition in ~\eqref{eq:multilayer_steering}, with cost proportional to the number of selected layers and intervention positions.

\section{Activation Steering: Methods and Limitations}
Activation steering controls LLM behavior by modifying internal representations at inference time without updating model parameters~\cite{li2023inference, nguyen2026beyond}. In contrast, parameter-efficient fine-tuning (PEFT) methods, such as Low-Rank Adaptation (LoRA)~\cite{hu2022lora} and adapters~\cite{houlsby2019parameter, van2024adapters, nguyen-le-2024-generalizability}, adapt pretrained models by learning a small subset of additional or modified parameters while keeping most model weights frozen. Although PEFT substantially reduces the computational and memory costs of full fine-tuning, it still requires task-specific optimization and parameter updates. Activation steering instead enables lightweight behavioral interventions without retraining, making it particularly attractive for flexible inference-time control.

Representative approaches include Representation Engineering (RepE)~\cite{zou2023representation}, Inference-Time Intervention (ITI)~\cite{li2023inference}, and Contrastive Activation Addition (CAA)~\cite{rimsky2024steering}. These methods typically construct steering directions from contrastive examples using mean differences, principal components, or linear classifiers, often yielding input-independent interventions. Despite their simplicity and efficiency, such approaches generally assume that the desired behavioral change can be approximated by a global linear direction in representation space. Subsequent work extends the expressivity and robustness of activation interventions through refined steering directions, geometric transformations such as optimal-transport alignment~\cite{rodriguez2025controlling}, norm-preserving transformations, and adaptive controllers based on auxiliary models or dynamical systems~\cite{kong2024aligning}.

However, the effectiveness of these interventions depends not only on how steering directions are constructed, but also on where they are applied within the network. Different transformer layers encode distinct representational properties, and applying the same steering mechanism at different depths can produce substantially different behavioral outcomes. This motivates systematic layer selection as a complementary problem to steering-vector construction. In particular, our work investigates whether the topology of contrastive activation displacements can identify effective intervention layers without additional validation-based steering evaluations.

\section{Experimental Details}
\label{app:expdetails}
\subsection{Setup}
\paragraph{Tasks.} We evaluate global layer selection on six binary behavioral steering tasks from the Anthropic Persona Dataset~\citep{perez-etal-2023-discovering}: Conscientiousness, Religion Following, Self-Awareness, Self-Improvement, Alliance Building, and Impact Maximization. Each example contains a question paired with two candidate responses, one aligned with the target behavior and one aligned with its opposite. Each behavior contains approximately $1000$ examples. We use approximately $700$ paired examples to extract the layerwise MD vectors and compute the layer scores, and reserve $100$ examples for evaluation.

\paragraph{Models.} We evaluate five instruction-tuned LLMs spanning two model families and multiple scales: \textsc{Llama-3-3B-Instruct}, \textsc{Llama-3-8B-Instruct}~\citep{grattafiori2024llama}, \textsc{Qwen2.5-3B-Instruct}, \textsc{Qwen2.5-7B-Instruct}, and \textsc{Qwen2.5-32B-Instruct}~\citep{team2024qwen2}. All interventions are applied at inference time without modifying model parameters.

\paragraph{Baselines.}
Our primary comparison is LayerNavigator~\citep{sun2025layernavigator}, the closest prior method for validation-free global layer selection. LayerNavigator scores each layer as
$s_\ell = d_\ell + c_\ell$, combining a discriminability term with a consistency term, and steers the top-$K$ layers. The discriminability term is a Fisher-style quotient, $d_\ell = (v_\ell^{\top} S_b\, v_\ell)/(v_\ell^{\top} S_t\, v_\ell)$, where $v_\ell$ is the unit-normalized steering direction and $S_b$, $S_t$ are the between-class and total scatter matrices of the standardized activations; it is large when the two classes separate along $v_\ell$. Since $d_\ell$ is the term most directly
comparable to $\cL$---both score a layer by how the contrastive pair geometry is arranged there---we also report \textbf{Disc}, which ranks layers by $d_\ell$ alone.

We include two further references. \textbf{Random} draws $K$ distinct layers uniformly without replacement from the middle third of the stack, $[\lfloor L/3\rfloor, \lfloor 2L/3\rfloor)$, under a fixed seed; the restricted pool makes this a competitive rather than a degenerate baseline. \textbf{Base} is the unsteered model, which bounds how much of any gain is attributable to steering at all. All methods use identical mean-difference steering vectors, the same per-layer strength $\alpha=1$, and the same budgets $K\in\{1,2,3\}$ applied simultaneously in one forward pass; they differ only in which layers they select.

\paragraph{Evaluation Metrics.} We measure behavioral steering using the average probability assigned to the response aligned with the target behavior. We additionally report generation perplexity to quantify the effect of steering on fluency. Higher aligned-answer probability and lower perplexity are preferred.

\paragraph{Implementation Details.} We extract residual-stream activations from the final prompt token at every model layer. The layerwise MD vectors are computed from the raw activations, while \ourmethod scores are computed from the standardized displacement clouds defined in ~\eqref{eq:displacement_cloud}. At inference time, we inject the layer-specific vector $v_l$ at each selected layer $l\in S_K$ and at all intervention token positions. Unless otherwise stated, we use steering strength $\alpha=1.0$ and evaluate $K\in\{1,2,3\}$ under both intervention regimes of Sec.~\ref{sec:toposteer_algorithm}; at $K=1$ the two regimes coincide, so it is reported once. All filtrations are Euclidean.

\paragraph{Computational Resources.}
All experiments were conducted using CUDA 12.8 on a single NVIDIA L40S and NVIDIA H100 GPU within a shared CPU Slurm cluster.

\paragraph{Codebase.}
Our implementation builds on the codebase of LayerNavigator~\citep{sun2025layernavigator}.

\section{Detailed Alignment Results}
\label{app:additional}
In this section, we provide detailed performance alignment under different variants of scoring metrics.

\subsection{Alignment performance under total persistence score}
Table~\ref{tab:alignment_layer_selection} reports per-task alignment probability and perplexity for every selector under simultaneous steering; Table~\ref{tab:sequential_detail} reports the sequential regime.
\begin{table*}[tb]
    \centering
    \tiny
    \renewcommand{\tabcolsep}{1pt}
    \renewcommand{\arraystretch}{0.82}
    \resizebox{\textwidth}{!}{
        \begin{tabular}{ccl*{12}{c}}
            \toprule
            Model & $K$ & Method
            & \multicolumn{2}{c}{Conscientiousness}
            & \multicolumn{2}{c}{Religion}
            & \multicolumn{2}{c}{Self-Improvement}
            & \multicolumn{2}{c}{Self-Aware}
            & \multicolumn{2}{c}{Alliance}
            & \multicolumn{2}{c}{Impact} \\
            \cmidrule(lr){4-5}
            \cmidrule(lr){6-7}
            \cmidrule(lr){8-9}
            \cmidrule(lr){10-11}
            \cmidrule(lr){12-13}
            \cmidrule(lr){14-15}
            &
            &
            &
            Prob ($\uparrow$) & PPL ($\downarrow$)
            & Prob ($\uparrow$) & PPL ($\downarrow$)
            & Prob ($\uparrow$) & PPL ($\downarrow$)
            & Prob ($\uparrow$) & PPL ($\downarrow$)
            & Prob ($\uparrow$) & PPL ($\downarrow$)
            & Prob ($\uparrow$) & PPL ($\downarrow$) \\
            \midrule
            \multirow{16}{*}{
                \rotatebox[origin=c]{90}{\bfseries Llama-3B}
            }
            & $K=0$ & Base & 72.51 & 26.82 & 55.68 & 21.40 & 70.96 & 25.13 & 59.77 & 30.25 & 75.56 & 26.38 & 65.89 & 25.00 \\
            \cmidrule(lr){2-15} 
            & \multirow{5}{*}{$K=1$} & D-only & 74.13 & 26.91 & 62.90 & 21.69 & 74.00 & 25.94 & 61.25 & 30.89 & 82.16 & 26.59 & 69.87 & 25.22 \\
            & & Random Selection & 73.42 & 26.80 & 62.82 & 21.55 & 71.51 & 25.06& 60.68 & 30.30 & 84.70 & 26.54 & 66.63& 24.98\\
            & & LayerNavigator & 77.16 & 26.89 & 64.90 & 21.56 & 76.00 & 25.34 & 66.61 & 30.29 & 87.08 & 26.59 & 71.01 & 25.18 \\
            & & \ourmethod{} $(l_2)$& 77.16 & 26.89 & 64.90 & 21.56 & 76.00 & 25.34 & 66.61 & 30.29 & 82.17 & 26.59 & 69.87 & 25.22 \\
            
            \cmidrule(lr){2-15}
            & \multirow{5}{*}{$K=2$} & D-only & 80.18 & 26.89 & 81.88 & 21.63 & 81.99 & 25.40 & 73.92 & 30.57 & 86.85 & 26.67 & 73.46 & 25.32 \\
            
            & & Random Selection & 82.06 & 26.88 & 67.61 & 21.59 & 82.34 & 25.44 & 60.84 & 30.18 & 89.56 & 26.57 & 70.00 & 25.32 \\
            & & LayerNavigator & 87.00 & 26.89 & 81.88 & 21.63 & 81.99 & 25.40 & 73.92 & 30.57 & 90.00 & 26.71 & 73.46 & 25.32 \\
            & & \ourmethod{} & 87.00 & 26.89 & 81.88 & 21.63 & 81.99 & 25.40 & 73.92 & 30.57 & 90.00 & 26.71 & 73.46 & 25.32 \\
            
            \cmidrule(lr){2-15} 
            & \multirow{5}{*}{$K=3$} & D-only & 84.88 & 26.90  & 83.64 & 21.74 & 64.51 & 25.58 & 74.21 & 30.69 & 83.41 & 26.98 & 71.51 & 25.95\\
            & & Random Selection & 83.20 & 26.89 & 82.10 & 21.64 & 61.25 & 26.51 & 72.29 & 30.45 &71.30 & 27.59 & 70.52&26.07\\
            & & LayerNavigator & 87.88 & 26.90 & 85.65 & 21.64 & 64.21 & 25.58 & 74.60 & 30.55 & 70.77 & 27.08 & 67.62 & 25.44 \\
            & & \ourmethod{} $(l_2)$& 86.70 & 26.90 & 85.87 & 21.62 & 64.21 & 25.58 & 74.60 & 30.55 & 85.41 & 26.84 & 67.62 & 25.44 \\
            \midrule
            
            \multirow{16}{*}{
                \rotatebox[origin=c]{90}{\textbf{Llama-8B}}
            }
            & $K=0$ & Base & 82.06 & 12.79 & 68.93 & 9.46 & 74.17 & 12.20 & 81.86 & 12.32 & 81.22 & 11.08 & 77.45 & 9.55 \\
            \cmidrule(lr){2-15} 
            & \multirow{5}{*}{$K=1$} & D-only & 83.33 & 12.79 & 78.64 & 9.41 & 73.05 & 12.23 & 82.37 & 12.30 & 81.70 & 11.58 & 79.30 & 9.53 \\
            & & Random Selection & 81.89& 12.74& 71.79 & 9.37 & 74.14 & 12.20 & 85.42 & 12.36 & 89.38 & 11.06 & 77.63 & 9.54\\
            & & LayerNavigator & 85.34 & 12.79 & 78.64 & 9.41 & 74.74 & 12.18 & 84.50 & 12.32 & 81.70 & 11.08 & 82.62 & 9.52 \\
            & & \ourmethod{} $(l_2)$& 87.09 & 12.78 & 71.79 & 9.37 & 80.11 & 12.13 & 88.34 & 12.40 & 86.23 & 11.06 & 82.62 & 9.52 \\
            
            \cmidrule(lr){2-15}
            & \multirow{5}{*}{$K=2$} & D-only & 86.92 & 12.78 & 78.98 & 9.40 & 74.18 & 12.18 & 85.99 & 12.34 & 84.30 & 11.06 & 79.85 & 9.53 \\
            & & Random Selection & 85.33 & 12.79 & 79.12 & 9.41 & 83.23 & 12.13 & 85.17 & 12.36 & 86.90 & 11.06 & 83.24 & 9.53 \\
            & & LayerNavigator & 90.54 & 12.73 & 84.05 & 9.40 & 74.18 & 12.18 & 91.60 & 12.43 & 84.30 & 11.06 & 84.54 & 9.53 \\
            & & \ourmethod{} $(l_2)$& 91.79 & 12.74 & 79.23 & 9.36 & 88.34 & 12.10 & 88.57 & 12.46 & 89.50 & 11.05 & 85.10 & 9.52 \\
            
            \cmidrule(lr){2-15} 
            & \multirow{5}{*}{$K=3$} & D-only & 86.36 & 12.78 & 79.57 & 9.38 & 83.00 & 12.07 & 85.81 & 12.33 & 84.44 & 11.06 & 85.47 & 9.52\\
            & & Random Selection & 85.35& 12.79& 82.65& 9.42 & 84.80 & 12.15 & 90.46 & 12.44& 89.02& 11.07 & 82.45 & 9.53\\
            & & LayerNavigator & 93.83 & 12.75 & 83.61 & 9.38 & 83.00 & 12.07 & 82.01 & 12.54 & 91.76 & 11.06 & 84.36 & 9.53 \\
            & & \ourmethod{} $(l_2)$& 93.83 & 12.75 & 77.80 & 9.37 & 90.18 & 12.08 & 73.29 & 12.72 & 88.94 & 11.27 & 84.36 & 9.53 \\
            
            \midrule
            \multirow{16}{*}{
                \rotatebox[origin=c]{90}{\bfseries Qwen-3B}
            }
            & $K=0$ & Base & 69.41 & 5.06 & 59.95 & 4.88 & 62.58 & 4.37 & 57.36 & 4.59 & 75.38 & 5.24 & 60.51 & 4.47 \\
            \cmidrule(lr){2-15} 
            & \multirow{5}{*}{$K=1$} & D-only & 60.89 & 5.26 & 54.73 & 4.96 & 59.41 & 5.17 & 57.09 & 4.76 & 78.36& 7.83 & 56.27 & 5.42\\
            & & Random Selection & 69.71 & 5.00& 58.52& 4.74& 62.76& 4.34& 55.98& 5.53 & 75.63& 5.19& 60.67& 4.56\\
            & & LayerNavigator & 61.27 & 5.34 & 54.38 & 5.10 & 65.13 & 5.47 & 64.64 & 8.60 & 78.41 & 7.73 & 62.00 & 6.33 \\
            & & \ourmethod{} $(l_2)$& 77.87 & 5.40 & 63.51 & 5.55 & 72.88 & 4.89 & 67.06 & 5.69 & 81.91 & 7.18 & 67.40 & 5.17 \\
            
            \cmidrule(lr){2-15}
            & \multirow{5}{*}{$K=2$} & D-only & 52.05 & 7.67 & 50.24 & 6.41 & 62.44 & 6.61 & 59.01 & 5.18 & 77.07 & 9.24 & 56.95 & 7.06 \\
            
            & & Random Selection & 60.91 & 5.33 & 57.90 & 4.95 & 62.50 & 4.41 & 53.31 & 5.05 & 75.35 & 8.62 & 61.76 & 6.07 \\
            & & LayerNavigator & 52.23 & 7.55 & 50.31 & 6.45 & 62.14 & 6.63 & 73.22 & 11.62 & 82.03 & 11.09 & 65.02 & 8.13 \\
            & & \ourmethod{} $(l_2)$& 82.35 & 6.47 & 66.32 & 6.15 & 81.24 & 5.50 & 76.00 & 7.00 & 85.98 & 9.14 & 73.88 & 8.63 \\
            
            \cmidrule(lr){2-15} 
            & \multirow{5}{*}{$K=3$} & D-only & 51.44 & 10.06 & 50.12 & 7.75 & 53.34 & 7.64 & 52.96 & 5.91 & 70.63 & 9.23 & 60.00 & 9.05\\
            & & Random Selection & 68.96 & 5.13 &55.03 &4.63 & 58.70 & 5.75 & 55.68 & 5.49& 74.93& 5.23& 56.02& 5.48\\
            & & LayerNavigator & 51.58 & 9.74 & 50.14 & 7.86 & 72.09 & 8.59 & 82.96 & 11.71 & 80.73 & 11.47 & 59.73 & 8.86 \\
            & & \ourmethod{} $(l_2)$& 84.86 & 7.61 & 68.83 & 6.97 & 81.49 & 7.49 & 81.69 & 9.49 & 87.36 & 10.89 & 74.02 & 9.18 \\

            \midrule
            \multirow{16}{*}{
                \rotatebox[origin=c]{90}{\bfseries Qwen-32B}
            }
            & $K=0$ & Base & 95.57 & 3.45 & 60.77 & 3.38 & 82.59 & 3.76 & 97.88 & 3.88 & 91.97 & 3.81 & 67.28 & 3.69 \\
            \cmidrule(lr){2-15} 
            & \multirow{5}{*}{$K=1$} & D-only & 94.72 & 3.51& 63.16 & 3.41& 81.41 & 3.88 & 97.97 & 4.32 & 92.51 &4.65 & 66.91 & 3.73 \\
            & & Random Selection & 95.51 & 3.43 & 62.44 & 3.51& 82.58 & 3.76 & 97.99 & 4.31 & 92.96 & 3.97& 67.13& 3.70\\
            & & LayerNavigator & 95.77 & 4.45 & 63.16 & 3.41 & 83.43 & 4.11 & 97.93 & 4.41 & 92.51 & 4.65 & 67.71 & 3.99 \\
            & & \ourmethod{} $(l_2)$& 95.74 & 4.26 & 60.95 & 3.40 & 83.36 & 4.20 & 98.19 & 4.68 & 93.17 & 4.51 & 67.96 & 4.31 \\
            
            \cmidrule(lr){2-15}
            & \multirow{5}{*}{$K=2$} & D-only & 95.38 & 3.84 & 61.85 & 3.61 & 81.43 & 3.94 & 97.16 & 4.57 & 93.08 & 5.63 & 66.14 & 3.82 \\
            & & Random Selection & 95.39 & 3.99 & 61.95 & 3.58 & 81.68 & 3.89 & 98.05 & 4.56 & 92.84 & 4.65 & 67.90 & 4.04 \\
            & & LayerNavigator & 95.90 & 7.91 & 61.82 & 3.61 & 82.96 & 5.05 & 98.04 & 5.83 & 92.93 & 7.57 & 66.83 & 3.96 \\
            & & \ourmethod{} $(l_2)$& 95.62 & 6.31 & 61.32 & 3.78 & 84.14 & 5.17 & 97.65 & 5.95 & 93.81 & 6.57 & 69.60 & 5.99 \\
            
            \cmidrule(lr){2-15} 
            & \multirow{5}{*}{$K=3$} & D-only & 94.71& 4.10& 61.01 &4.19 & 80.34 &4.28 & 97.23 & 5.25& 93.27 & 6.24 & 64.96& 4.01 \\
            & & Random Selection & 94.96& 3.94& 61.99 & 4.17 & 81.59& 4.30& 97.18& 4.50 & 92.29 & 4.20 & 66.64 & 3.88\\
            & & LayerNavigator & 95.94 & 11.03 & 61.01 & 4.19 & 82.97 & 5.23 & 97.77 & 9.72 & 86.88 & 10.12 & 65.90 & 4.08 \\
            & & \ourmethod{} $(l_2)$& 94.66 & 10.04 & 62.23 & 4.16 & 83.13 & 7.68 & 96.91 & 9.75 & 94.72 & 16.80 & 70.22 & 7.54 \\
            \bottomrule
        \end{tabular}
    }
    \caption{Per-task alignment probability (Prob, \%, $\uparrow$) and generation perplexity (PPL, $\downarrow$) for each layer-selection strategy under simultaneous steering, on the six persona tasks. \ourmethod{} here scores the unit-sphere projection of the displacement
    cloud, i.e.\ the \textbf{Ours}$_{\ell_2}$ column of Table~\ref{tab:main};
    Table~\ref{tab:alignment_layer_selection_without_l2_diff} gives the corresponding results for the standardized cloud $\Delta_l$. $K=0$ is the unsteered model. All filtrations are Euclidean and $\alpha=1.0$.}
    \label{tab:alignment_layer_selection}
\end{table*}

\begin{table*}[tb]
    \centering
    \tiny
    \renewcommand{\tabcolsep}{1pt}
    \renewcommand{\arraystretch}{0.82}
    \resizebox{\textwidth}{!}{
        \begin{tabular}{ccl*{12}{c}}
            \toprule
            Model & $K$ & Method
            & \multicolumn{2}{c}{Conscientiousness}
            & \multicolumn{2}{c}{Religion}
            & \multicolumn{2}{c}{Self-Improvement}
            & \multicolumn{2}{c}{Self-Aware}
            & \multicolumn{2}{c}{Alliance}
            & \multicolumn{2}{c}{Impact} \\
            \cmidrule(lr){4-5}
            \cmidrule(lr){6-7}
            \cmidrule(lr){8-9}
            \cmidrule(lr){10-11}
            \cmidrule(lr){12-13}
            \cmidrule(lr){14-15}
            &
            &
            &
            Prob ($\uparrow$) & PPL ($\downarrow$)
            & Prob ($\uparrow$) & PPL ($\downarrow$)
            & Prob ($\uparrow$) & PPL ($\downarrow$)
            & Prob ($\uparrow$) & PPL ($\downarrow$)
            & Prob ($\uparrow$) & PPL ($\downarrow$)
            & Prob ($\uparrow$) & PPL ($\downarrow$) \\
            \midrule
            \multirow{16}{*}{
                \rotatebox[origin=c]{90}{\bfseries Llama-3B}
            }
            & $K=0$ & Base & 72.51 & 26.82 & 55.68 & 21.40 & 70.96 & 25.13 & 59.77 & 30.25 & 75.56 & 26.38 & 65.89 & 25.00 \\
            \cmidrule(lr){2-15} 
            & \multirow{5}{*}{$K=1$} & D-only & 74.13 & 26.91 & 62.90 & 21.69 & 74.00 & 25.94 & 61.25 & 30.89 & 82.16 & 26.59 & 69.87 & 25.22 \\
            & & Random Selection & 73.42 & 26.80 & 62.82 & 21.55 & 71.51 & 25.06& 60.68 & 30.30 & 84.70 & 26.54 & 66.63& 24.98\\
            & & LayerNavigator & 77.16 & 26.89 & 64.90 & 21.56 & 76.00 & 25.34 & 66.61 & 30.29 & 87.08 & 26.59 & 71.01 & 25.18 \\
            & & \ourmethod{} & 77.16 & 26.89 & 64.90 & 21.56 & 76.00 & 25.34 & 66.61 & 30.29 & 82.17 & 26.59 & 69.87 & 25.22 \\
            
            \cmidrule(lr){2-15}
            & \multirow{5}{*}{$K=2$} & D-only & 80.18 & 26.89 & 81.88 & 21.63 & 81.99 & 25.40 & 73.92 & 30.57 & 86.85 & 26.67 & 73.46 & 25.32 \\
            
            & & Random Selection & 82.06 & 26.88 & 67.61 & 21.59 & 82.34 & 25.44 & 60.84 & 30.18 & 89.56 & 26.57 & 70.00 & 25.32 \\
            & & LayerNavigator & 87.00 & 26.89 & 81.88 & 21.63 & 81.99 & 25.40 & 73.92 & 30.57 & 90.00 & 26.71 & 73.46 & 25.32 \\
            & & \ourmethod{} & 87.00 & 26.89 & 81.88 & 21.63 & 81.99 & 25.40 & 73.92 & 30.57 & 90.00 & 26.71 & 73.46 & 25.32 \\
            
            \cmidrule(lr){2-15} 
            & \multirow{5}{*}{$K=3$} & D-only & 84.88 & 26.90  & 83.64 & 21.74 & 64.51 & 25.58 & 74.21 & 30.69 & 83.41 & 26.98 & 71.51 & 25.95\\
            & & Random Selection & 83.20 & 26.89 & 82.10 & 21.64 & 61.25 & 26.51 & 72.29 & 30.45 &71.30 & 27.59 & 70.52&26.07\\
            & & LayerNavigator & 87.88 & 26.90 & 85.65 & 21.64 & 64.21 & 25.58 & 74.60 & 30.55 & 70.77 & 27.08 & 67.62 & 25.44 \\
            & & \ourmethod{} & 86.70 & 26.90 & 85.87 & 21.62 & 76.52 & 25.54 & 74.68 & 31.10 & 85.41 & 26.84 & 73.78 & 25.41 \\
            \midrule
            
            \multirow{16}{*}{
                \rotatebox[origin=c]{90}{\textbf{Llama-8B}}
            }
            & $K=0$ & Base & 82.06 & 12.79 & 68.93 & 9.46 & 74.17 & 12.20 & 81.86 & 12.32 & 81.22 & 11.08 & 77.45 & 9.55 \\
            \cmidrule(lr){2-15} 
            & \multirow{5}{*}{$K=1$} & D-only & 83.33 & 12.79 & 78.64 & 9.41 & 73.05 & 12.23 & 82.37 & 12.30 & 81.70 & 11.58 & 79.30 & 9.53 \\
            & & Random Selection & 81.89& 12.74& 71.79 & 9.37 & 74.14 & 12.20 & 85.42 & 12.36 & 89.38 & 11.06 & 77.63 & 9.54\\
            & & LayerNavigator & 85.34 & 12.79 & 78.64 & 9.41 & 74.74 & 12.18 & 84.50 & 12.32 & 81.70 & 11.08 & 82.62 & 9.52 \\
            & & \ourmethod{} & 87.09 & 12.78 & 78.64 & 9.41 & 80.11 & 12.13 & 89.19 & 12.39 & 82.96 & 11.06 & 82.62 & 9.52 \\
            
            \cmidrule(lr){2-15}
            & \multirow{5}{*}{$K=2$} & D-only & 86.92 & 12.78 & 78.98 & 9.40 & 74.18 & 12.18 & 85.99 & 12.34 & 84.30 & 11.06 & 79.85 & 9.53 \\
            & & Random Selection & 85.33 & 12.79 & 79.12 & 9.41 & 83.23 & 12.13 & 85.17 & 12.36 & 86.90 & 11.06 & 83.24 & 9.53 \\
            & & LayerNavigator & 90.54 & 12.73 & 84.05 & 9.40 & 74.18 & 12.18 & 91.60 & 12.43 & 84.30 & 11.06 & 84.54 & 9.53 \\
            & & \ourmethod{} & 88.91 & 12.76 & 79.23 & 9.36 & 80.01 & 12.11 & 88.57 & 12.46 & 84.30 & 11.06 & 84.54 & 9.53 \\
            
            \cmidrule(lr){2-15} 
            & \multirow{5}{*}{$K=3$} & D-only & 86.36 & 12.78 & 79.57 & 9.38 & 83.00 & 12.07 & 85.81 & 12.33 & 84.44 & 11.06 & 85.47 & 9.52\\
            & & Random Selection & 85.35& 12.79& 82.65& 9.42 & 84.80 & 12.15 & 90.46 & 12.44& 89.02& 11.07 & 82.45 & 9.53\\
            & & LayerNavigator & 93.83 & 12.75 & 83.61 & 9.38 & 83.00 & 12.07 & 82.01 & 12.54 & 91.76 & 11.06 & 84.36 & 9.53 \\
            & & \ourmethod{} & 92.20 & 12.73 & 82.10 & 9.37 & 83.00 & 12.07 & 82.01 & 12.54 & 91.76 & 11.06 & 84.36 & 9.53 \\
            
            \midrule
            \multirow{16}{*}{
                \rotatebox[origin=c]{90}{\bfseries Qwen-3B}
            }
            & $K=0$ & Base & 69.41 & 5.06 & 59.95 & 4.88 & 62.58 & 4.37 & 57.36 & 4.59 & 75.38 & 5.24 & 60.51 & 4.47 \\
            \cmidrule(lr){2-15} 
            & \multirow{5}{*}{$K=1$} & D-only & 60.89 & 5.26 & 54.73 & 4.96 & 59.41 & 5.17 & 57.09 & 4.76 & 78.36& 7.83 & 56.27 & 5.42\\
            & & Random Selection & 69.71 & 5.00& 58.52& 4.74& 62.76& 4.34& 55.98& 5.53 & 75.63& 5.19& 60.67& 4.56\\
            & & LayerNavigator & 61.27 & 5.34 & 54.38 & 5.10 & 65.13 & 5.47 & 64.64 & 8.60 & 78.41 & 7.73 & 62.00 & 6.33 \\
            & & \ourmethod{} & 73.56 & 6.54 & 63.51 & 5.55 & 59.21 & 5.32 & 62.54 & 5.29 & 74.42 & 6.13 & 55.90 & 5.50 \\
            
            \cmidrule(lr){2-15}
            & \multirow{5}{*}{$K=2$} & D-only & 52.05 & 7.67 & 50.24 & 6.41 & 62.44 & 6.61 & 59.01 & 5.18 & 77.07 & 9.24 & 56.95 & 7.06 \\
            
            & & Random Selection & 60.91 & 5.33 & 57.90 & 4.95 & 62.50 & 4.41 & 53.31 & 5.05 & 75.35 & 8.62 & 61.76 & 6.07 \\
            & & LayerNavigator & 52.23 & 7.55 & 50.31 & 6.45 & 62.14 & 6.63 & 73.22 & 11.62 & 82.03 & 11.09 & 65.02 & 8.13 \\
            & & \ourmethod{} & 62.26 & 7.90 & 66.32 & 6.15 & 53.57 & 6.55 & 60.43 & 6.67 & 65.99 & 6.64 & 56.81 & 6.87 \\
            
            \cmidrule(lr){2-15} 
            & \multirow{5}{*}{$K=3$} & D-only & 51.44 & 10.06 & 50.12 & 7.75 & 53.34 & 7.64 & 52.96 & 5.91 & 70.63 & 9.23 & 60.00 & 9.05\\
            & & Random Selection & 68.96 & 5.13 &55.03 &4.63 & 58.70 & 5.75 & 55.68 & 5.49& 74.93& 5.23& 56.02& 5.48\\
            & & LayerNavigator & 51.58 & 9.74 & 50.14 & 7.86 & 72.09 & 8.59 & 82.96 & 11.71 & 80.73 & 11.47 & 59.73 & 8.86 \\
            & & \ourmethod{} & 51.58 & 9.74 & 68.83 & 6.97 & 53.33 & 7.67 & 72.12 & 9.19 & 70.32 & 9.33 & 50.27 & 7.99 \\
            
            \midrule
            \multirow{16}{*}{
                \rotatebox[origin=c]{90}{\bfseries Qwen-32B}
            }
            & $K=0$ & Base & 95.57 & 3.45 & 60.77 & 3.38 & 82.59 & 3.76 & 97.88 & 3.88 & 91.97 & 3.81 & 67.28 & 3.69 \\
            \cmidrule(lr){2-15} 
            & \multirow{5}{*}{$K=1$} & D-only & 94.72 & 3.51& 63.16 & 3.41& 81.41 & 3.88 & 97.97 & 4.32 & 92.51 &4.65 & 66.91 & 3.73 \\
            & & C-only & 95.77 & 4.45 & 60.77 & 3.38 & 83.08 & 4.33 & 97.93 & 4.41& 92.84& 4.70 & 67.71 & 3.99 \\
            & & Random Selection & 95.51 & 3.43 & 62.44 & 3.51& 82.58 & 3.76 & 97.99 & 4.31 & 92.96 & 3.97& 67.13& 3.70\\
            & & LayerNavigator & 95.77 & 4.45 & 63.16 & 3.41 & 83.43 & 4.11 & 97.93 & 4.41 & 92.51 & 4.65 & 67.71 & 3.99 \\
            & & \ourmethod{} & 95.74 & 4.26 & 61.09 & 3.48 & 83.36 & 4.20 & 98.19 & 4.33 & 93.16 & 4.48 & 67.91 & 3.91 \\
            
            \cmidrule(lr){2-15}
            & \multirow{5}{*}{$K=2$} & D-only & 95.38 & 3.84 & 61.85 & 3.61 & 81.43 & 3.94 & 97.16 & 4.57 & 93.08 & 5.63 & 66.14 & 3.82 \\
            & & C-only & 95.56 & 6.52 & 61.85 & 3.61 & 83.41 & 5.57 & 98.01 & 5.91 & 93.39 & 7.34 & 67.98 & 4.60 \\
            & & Random Selection & 95.39 & 3.99 & 61.95 & 3.58 & 81.68 & 3.89 & 98.05 & 4.56 & 92.84 & 4.65 & 67.90 & 4.04 \\
            & & LayerNavigator & 95.90 & 7.91 & 61.82 & 3.61 & 82.96 & 5.05 & 98.04 & 5.83 & 92.93 & 7.57 & 66.83 & 3.96 \\
            & & \ourmethod{} & 95.62 & 6.31 & 61.35 & 3.57 & 84.14 & 5.17 & 97.97 & 6.00 & 93.81 & 6.57 & 70.59 & 5.19 \\
            
            \cmidrule(lr){2-15} 
            & \multirow{5}{*}{$K=3$} & D-only & 94.71& 4.10& 61.01 &4.19 & 80.34 &4.28 & 97.23 & 5.25& 93.27 & 6.24 & 64.96& 4.01 \\
            & & C-only & 94.28& 10.78 & 61.01 & 4.19 & 83.84 & 9.35 & 97.95& 12.02& 93.84& 16.92 & 68.42& 5.80\\
            & & Random Selection & 94.96& 3.94& 61.99 & 4.17 & 81.59& 4.30& 97.18& 4.50 & 92.29 & 4.20 & 66.64 & 3.88\\
            & & LayerNavigator & 95.94 & 11.03 & 61.01 & 4.19 & 82.97 & 5.23 & 97.77 & 9.72 & 86.88 & 10.12 & 65.90 & 4.08 \\
            & & \ourmethod{} & 94.66 & 10.04 & 61.62 & 3.79 & 83.95 & 8.63 & 97.05 & 9.47 & 94.04 & 9.33 & 70.65 & 6.34 \\
            \bottomrule
        \end{tabular}
    }
    \caption{Per-task alignment probability (Prob, \%, $\uparrow$) and generation perplexity (PPL, $\downarrow$) for each layer-selection strategy under simultaneous steering, on the six persona tasks. \ourmethod{} here scores the standardized displacement cloud $\Delta_l$
    of~\eqref{eq:displacement_cloud}, i.e.\ the \textbf{Ours} column of
    Table~\ref{tab:main}; Table~\ref{tab:alignment_layer_selection} gives the corresponding results for the unit-sphere variant \ourmethod{}$_{\ell_2}$. $K=0$ is the unsteered model. All filtrations are Euclidean and $\alpha=1.0$.}
    \label{tab:alignment_layer_selection_without_l2_diff}
\end{table*}


\subsection{Sequential Steering with Distinct Layers}
Table~\ref{tab:sequential_detail} reports sequential steering results with distinct layers enforced at each selection step.

\begin{table*}[tb]
    \centering
    \tiny
    \renewcommand{\tabcolsep}{1pt}
    \renewcommand{\arraystretch}{0.82}
    \resizebox{\textwidth}{!}{
        \begin{tabular}{ccl*{12}{c}}
            \toprule
            Model & $K$ & Method
            & \multicolumn{2}{c}{Conscientiousness}
            & \multicolumn{2}{c}{Religion}
            & \multicolumn{2}{c}{Self-Improvement}
            & \multicolumn{2}{c}{Self-Aware}
            & \multicolumn{2}{c}{Alliance}
            & \multicolumn{2}{c}{Impact} \\
            \cmidrule(lr){4-5}
            \cmidrule(lr){6-7}
            \cmidrule(lr){8-9}
            \cmidrule(lr){10-11}
            \cmidrule(lr){12-13}
            \cmidrule(lr){14-15}
            &
            &
            &
            Prob ($\uparrow$) & PPL ($\downarrow$)
            & Prob ($\uparrow$) & PPL ($\downarrow$)
            & Prob ($\uparrow$) & PPL ($\downarrow$)
            & Prob ($\uparrow$) & PPL ($\downarrow$)
            & Prob ($\uparrow$) & PPL ($\downarrow$)
            & Prob ($\uparrow$) & PPL ($\downarrow$) \\
            \midrule
            \multirow{11}{*}{
                \rotatebox[origin=c]{90}{\bfseries Llama-3B}
            }
            & $K=0$ & Base & 72.51 & 26.82 & 55.68 & 21.40 & 70.96 & 25.13 & 59.77 & 30.25 & 75.56 & 26.38 & 65.89 & 24.99 \\
            \cmidrule(lr){2-15}
            & \multirow{5}{*}{$K=2$} & D-only & 80.18 & 26.89 & 81.17 & 21.63 & 77.99 & 25.42 & 73.03 & 30.58 & 86.85 & 26.67 & 73.23 & 25.30 \\
            & & Random Selection & 78.65 & 26.89 & 60.62 & 21.54 & 80.80 & 25.38 & 61.78 & 30.32 & 89.55 & 26.59 & 67.79 & 25.23 \\
            & & LayerNavigator & 84.68 & 26.89 & 81.17 & 21.63 & 81.34 & 25.41 & 73.03 & 30.58 & 90.00 & 26.71 & 73.46 & 25.32 \\
            & & \ourmethod{} & 84.68 & 26.89 & 81.17 & 21.63 & 81.34 & 25.41 & 73.03 & 30.58 & 88.99 & 26.68 & 73.23 & 25.31 \\
            & & \ourmethod{}$_{\ell_2}$ & 84.68 & 26.89 & 81.17 & 21.63 & 81.34 & 25.41 & 73.03 & 30.58 & 88.99 & 26.68 & 73.23 & 25.30 \\
            \cmidrule(lr){2-15}
            & \multirow{5}{*}{$K=3$} & D-only & 86.62 & 26.89 & 84.64 & 21.64 & 65.28 & 25.60 & 71.26 & 30.88 & 85.75 & 26.81 & 69.18 & 25.45 \\
            & & Random Selection & 80.29 & 26.86 & 66.14 & 21.64 & 84.82 & 25.44 & 62.61 & 30.30 & 90.37 & 26.63 & 71.85 & 25.29 \\
            & & LayerNavigator & 81.50 & 26.92 & 84.64 & 21.64 & 67.22 & 25.61 & 74.48 & 31.13 & 75.02 & 26.95 & 68.14 & 25.45 \\
            & & \ourmethod{} & 81.50 & 26.92 & 75.44 & 21.68 & 77.88 & 25.52 & 74.48 & 31.13 & 85.48 & 26.84 & 74.02 & 25.40 \\
            & & \ourmethod{}$_{\ell_2}$ & 81.50 & 26.92 & 75.44 & 21.68 & 67.22 & 25.61 & 71.26 & 30.88 & 73.79 & 26.98 & 69.18 & 25.45 \\
            \midrule
            \multirow{11}{*}{
                \rotatebox[origin=c]{90}{\bfseries Llama-8B}
            }
            & $K=0$ & Base & 82.07 & 12.79 & 68.93 & 9.46 & 74.17 & 12.20 & 81.86 & 12.32 & 81.23 & 11.08 & 77.45 & 9.55 \\
            \cmidrule(lr){2-15}
            & \multirow{5}{*}{$K=2$} & D-only & 82.52 & 12.79 & 78.98 & 9.40 & 73.59 & 12.19 & 81.91 & 12.30 & 83.77 & 11.06 & 79.85 & 9.53 \\
            & & Random Selection & 88.21 & 12.78 & 71.44 & 9.42 & 79.04 & 12.12 & 85.23 & 12.36 & 87.33 & 11.06 & 77.70 & 9.54 \\
            & & LayerNavigator & 88.76 & 12.74 & 84.05 & 9.40 & 74.18 & 12.18 & 90.41 & 12.39 & 83.77 & 11.06 & 84.54 & 9.53 \\
            & & \ourmethod{} & 88.91 & 12.76 & 78.15 & 9.37 & 80.01 & 12.11 & 88.72 & 12.44 & 84.30 & 11.06 & 84.54 & 9.53 \\
            & & \ourmethod{}$_{\ell_2}$ & 89.99 & 12.73 & 79.23 & 9.36 & 86.46 & 12.10 & 88.57 & 12.46 & 89.50 & 11.05 & 85.17 & 9.52 \\
            \cmidrule(lr){2-15}
            & \multirow{5}{*}{$K=3$} & D-only & 84.63 & 12.78 & 78.98 & 9.37 & 73.26 & 12.19 & 84.23 & 12.28 & 83.82 & 11.06 & 80.07 & 9.53 \\
            & & Random Selection & 87.99 & 12.78 & 74.77 & 9.42 & 81.42 & 12.10 & 85.49 & 12.35 & 87.25 & 11.06 & 81.67 & 9.53 \\
            & & LayerNavigator & 90.24 & 12.73 & 82.94 & 9.37 & 74.10 & 12.08 & 90.94 & 12.44 & 87.59 & 11.06 & 85.61 & 9.52 \\
            & & \ourmethod{} & 91.70 & 12.73 & 80.99 & 9.38 & 82.73 & 12.06 & 84.18 & 12.49 & 88.29 & 11.05 & 85.61 & 9.52 \\
            & & \ourmethod{}$_{\ell_2}$ & 92.96 & 12.75 & 76.64 & 9.38 & 87.70 & 12.09 & 79.84 & 12.52 & 92.12 & 11.07 & 84.63 & 9.51 \\
            \midrule
            \multirow{11}{*}{
                \rotatebox[origin=c]{90}{\bfseries Qwen-3B}
            }
            & $K=0$ & Base & 70.20 & 4.95 & 59.74 & 4.91 & 62.49 & 4.42 & 57.02 & 4.60 & 75.50 & 5.42 & 60.60 & 4.35 \\
            \cmidrule(lr){2-15}
            & \multirow{5}{*}{$K=2$} & D-only & 51.72 & 7.10 & 50.24 & 6.41 & 53.91 & 6.52 & 59.01 & 5.18 & 77.07 & 9.24 & 50.77 & 6.74 \\
            & & Random Selection & 70.55 & 6.87 & 53.41 & 6.15 & 61.50 & 4.44 & 58.75 & 4.84 & 75.99 & 4.89 & 60.66 & 4.49 \\
            & & LayerNavigator & 51.72 & 7.10 & 52.23 & 9.06 & 69.04 & 7.82 & 72.89 & 11.61 & 80.81 & 11.08 & 62.58 & 8.19 \\
            & & \ourmethod{} & 72.76 & 11.41 & 66.32 & 6.15 & 53.57 & 6.55 & 60.43 & 6.67 & 65.99 & 6.64 & 50.73 & 6.75 \\
            & & \ourmethod{}$_{\ell_2}$ & 80.83 & 6.61 & 65.92 & 6.14 & 80.53 & 5.53 & 75.57 & 7.22 & 85.84 & 9.13 & 71.65 & 5.71 \\
            & \multirow{5}{*}{$K=3$} & D-only & 53.43 & 9.82 & 50.10 & 6.77 & 54.39 & 9.25 & 52.87 & 5.79 & 79.26 & 11.30 & 51.08 & 8.46 \\
            & & Random Selection & 70.66 & 6.84 & 53.31 & 6.16 & 61.62 & 4.41 & 58.56 & 4.89 & 76.02 & 4.93 & 60.42 & 4.70 \\
            & & LayerNavigator & 53.43 & 9.82 & 50.91 & 10.59 & 77.52 & 9.92 & 67.42 & 12.17 & 83.64 & 11.28 & 65.98 & 9.21 \\
            & & \ourmethod{} & 62.98 & 12.65 & 68.83 & 6.97 & 54.29 & 9.40 & 68.31 & 9.63 & 62.53 & 8.29 & 50.35 & 8.40 \\
            & & \ourmethod{}$_{\ell_2}$ & 83.27 & 7.87 & 68.57 & 7.14 & 81.32 & 7.52 & 80.87 & 10.03 & 87.37 & 10.82 & 73.75 & 6.83 \\
            \midrule
            \multirow{11}{*}{
                \rotatebox[origin=c]{90}{\bfseries Qwen-7B}
            }
            & $K=0$ & Base & 88.08 & 15.30 & 84.39 & 9.80 & 83.28 & 10.94 & 94.43 & 11.82 & 87.43 & 10.61 & 73.80 & 9.50 \\
            \cmidrule(lr){2-15}
            & \multirow{5}{*}{$K=2$} & D-only & 84.84 & 17.29 & 82.63 & 7.38 & 85.59 & 26.35 & 95.94 & 19.86 & 89.75 & 24.64 & 76.40 & 9.30 \\
            & & Random Selection & 85.73 & 16.93 & 85.56 & 8.11 & 84.31 & 10.49 & 95.14 & 12.97 & 89.44 & 12.79 & 74.63 & 9.15 \\
            & & LayerNavigator & 83.24 & 19.99 & 83.91 & 14.76 & 86.55 & 25.14 & 95.65 & 21.75 & 89.75 & 24.64 & 78.02 & 13.11 \\
            & & \ourmethod{} & 84.23 & 18.51 & 84.57 & 11.72 & 84.76 & 23.70 & 95.19 & 24.46 & 90.48 & 23.04 & 75.45 & 16.78 \\
            & & \ourmethod{}$_{\ell_2}$ & 84.97 & 19.53 & 85.04 & 10.83 & 84.67 & 22.01 & 94.87 & 22.21 & 89.38 & 22.59 & 74.14 & 15.46 \\
            \cmidrule(lr){2-15}
            & \multirow{5}{*}{$K=3$} & D-only & 84.50 & 16.70 & 85.98 & 9.46 & 87.55 & 27.53 & 96.18 & 27.97 & 82.83 & 25.05 & 77.46 & 12.37 \\
            & & Random Selection & 85.71 & 16.82 & 87.76 & 12.06 & 84.83 & 14.10 & 95.09 & 13.19 & 90.71 & 15.53 & 76.82 & 9.47 \\
            & & LayerNavigator & 87.63 & 20.38 & 83.09 & 16.11 & 87.37 & 26.95 & 96.38 & 27.77 & 82.83 & 25.05 & 80.52 & 18.42 \\
            & & \ourmethod{} & 83.41 & 19.60 & 84.76 & 11.74 & 84.79 & 26.36 & 95.03 & 27.19 & 90.67 & 24.30 & 75.32 & 19.64 \\
            & & \ourmethod{}$_{\ell_2}$ & 84.31 & 21.05 & 84.83 & 11.82 & 84.63 & 26.26 & 94.80 & 27.49 & 89.67 & 24.21 & 74.04 & 19.24 \\
            \midrule
            \multirow{11}{*}{
                \rotatebox[origin=c]{90}{\bfseries Qwen-32B}
            }
            & $K=0$ & Base & 95.57 & 3.45 & 60.77 & 3.38 & 82.59 & 3.76 & 97.89 & 3.88 & 91.97 & 3.81 & 67.28 & 3.69 \\
            \cmidrule(lr){2-15}
            & \multirow{5}{*}{$K=2$} & D-only & 95.34 & 3.69 & 61.87 & 3.63 & 80.69 & 3.91 & 97.46 & 5.35 & 93.05 & 5.81 & 66.53 & 3.83 \\
            & & Random Selection & 92.97 & 3.99 & 65.27 & 3.82 & 81.12 & 3.86 & 98.17 & 4.02 & 92.72 & 4.40 & 66.93 & 3.72 \\
            & & LayerNavigator & 95.53 & 6.47 & 61.87 & 3.63 & 82.71 & 4.88 & 97.80 & 6.86 & 91.91 & 8.05 & 67.87 & 4.42 \\
            & & \ourmethod{} & 95.62 & 6.31 & 61.32 & 3.58 & 84.14 & 5.17 & 97.97 & 6.00 & 93.78 & 6.59 & 68.33 & 4.41 \\
            & & \ourmethod{}$_{\ell_2}$ & 95.62 & 6.31 & 61.37 & 3.73 & 84.14 & 5.17 & 97.59 & 5.52 & 93.81 & 6.57 & 68.70 & 4.89 \\
            \cmidrule(lr){2-15}
            & \multirow{5}{*}{$K=3$} & D-only & 95.00 & 4.00 & 62.15 & 3.99 & 79.73 & 4.12 & 96.05 & 5.52 & 93.27 & 6.24 & 65.58 & 3.98 \\ 
            & & Random Selection & 92.88 & 3.88 & 64.37 & 3.71 & 81.00 & 3.88 & 97.95 & 4.13 & 92.70 & 4.51 & 67.48 & 3.73 \\
            & & LayerNavigator & 95.85 & 9.31 & 62.15 & 3.99 & 82.43 & 6.31 & 97.72 & 11.30 & 90.97 & 12.22 & 68.10 & 5.17 \\
            & & \ourmethod{} & 94.66 & 10.04 & 61.76 & 3.78 & 83.95 & 8.63 & 97.30 & 9.83 & 94.07 & 10.59 & 68.24 & 5.23 \\
            & & \ourmethod{}$_{\ell_2}$ & 94.89 & 10.01 & 62.28 & 4.11 & 83.13 & 7.68 & 97.13 & 7.49 & 94.25 & 16.09 & 69.30 & 5.47 \\
            \bottomrule
        \end{tabular}
    }
    \caption{Per-task alignment probability (Prob, \%) and perplexity (PPL, $\downarrow$) for greedy sequential layer selection without layer repetition under the Euclidean filtration: each step adds a layer not chosen in an earlier step and re-extracts the steering vectors. $K=1$ coincides with simultaneous steering and is omitted. Averaging a row over the six tasks gives the corresponding entry of the Sequential block in Table~\ref{tab:main}.}
    \label{tab:sequential_detail}
\end{table*}

\section{Separation-Free Standardization}
\label{app:sepablation}
\begin{proof}[Proof of Proposition~\ref{prop:sepscale}]
    The variance identity is the law of total variance for a balanced two-component mixture whose component means are $\pm\mu^{\mathrm{raw}}_{\delta,l}/2$ about the pooled mean, with $\bar W_{l,k}$ and $\sigma_{l,k}^{2}$ both taken with the same (biased) denominator. Since \eqref{eq:zscore} rescales coordinate $k$ by $\sigma_{l,k}^{-1}$ and the displacement is formed afterwards, $\xi^{i}_{l,k}=\xi^{\mathrm{raw},i}_{l,k}/\sigma_{l,k}
    =\eta^{i}_{l,k}(1+\rho_{l,k})^{-1/2}$, and Corollary~\ref{cor:inv} gives
    $\cL(\Delta_l)=\cL(\Xi_l)=\cL(H_l\Lambda_l)$. Increasing $\rho_{l,k}$ multiplies coordinate $k$ of every point by a factor in $(0,1)$, hence does not increase any pairwise distance; $\cL$ is non-decreasing in each entry of the distance matrix, so $\cL(\Delta_l)$ does not increase, strictly so when coordinate $k$ separates at least two points (some spanning-tree edge must then straddle two distinct values of that coordinate). The isotropic case is the scalar
    $\Lambda_l=(1+\rho_l)^{-1/2}I$ together with Proposition~\ref{prop:inv}.
\end{proof}

Proposition~\ref{prop:sepscale} shows that the pooled standardization~\eqref{eq:zscore} makes $\cL(\Delta_l)$ decrease in the coordinate-wise separation-to-noise ratios $\rho_{l,k}$, so a layer can be ranked lower than another purely because its classes are further apart. $D_l$ is invariant to a common scalar rescaling but not to this coordinate-wise one, so both criteria move under the shared standardization; the proposition isolates the effect on $\cL$, which is the one we can remove.

The effect can be separated from displacement geometry by dividing each coordinate by the pooled \emph{within-class} standard deviation $\bar W_{l,k}^{1/2}$ rather than by $\sigma_{l,k}$, which by Proposition~\ref{prop:sepscale} gives the separation-free score $\cL(H_l)$; this reuses the cached activations and adds no forward passes. All results in this paper use the pooled standardization~\eqref{eq:zscore}, which is also what LayerNavigator's scoring applies, so that the two criteria are compared on the same representation; comparing the rankings under the two standardizations is left to future work.

\section{Detail Layer Selection}
\label{additional_layer_selection}
The layers selected by each strategy on Qwen-3B, for every task and budget $K$, are given in Table~\ref{tab:layers_qwen3b}.

\begin{table*}[t]
    \centering
    \footnotesize
    \setlength{\aboverulesep}{0pt}
    \setlength{\belowrulesep}{0pt}
    \setlength{\extrarowheight}{0ex}
    \renewcommand{\tabcolsep}{1pt}
    \begin{tabular}{c l cccccc}
        \toprule
        $K$ & Method & Conscientiousness & Religion & Self-Improvement
            & Self-Aware & Alliance & Impact \\
        \midrule
        \multirow{6}{*}{$1$}
            & D-only & 21 & 22 & 22 & 20 & 23 & 22 \\
            & C-only & 26 & 23 & 34 & 25 & 34 & 25 \\
            & Random Selection & 12 & 17 & 12 & 23 & 18 & 12 \\
            & LayerNavigator & 21 & 22 & 23 & 25 & 23 & 23 \\
            & \ourmethod{} (diff) & 23 & 35 & 22 & 30 & 22 & 22 \\
            & \ourmethod{} ($\ell_2$) & 35 & 35 & 35 & 35 & 35 & 35 \\
        \midrule
        \multirow{6}{*}{$2$}
            & D-only & 21,\,22 & 22,\,21 & 22,\,23 & 20,\,19 & 23,\,22 & 22,\,23 \\
            & C-only & 26,\,21 & 23,\,24 & 34,\,33 & 25,\,34 & 34,\,35 & 25,\,23 \\
            & Random Selection & 13,\,21 & 13,\,16 & 15,\,18 & 12,\,21 & 21,\,23 & 17,\,23 \\
            & LayerNavigator & 21,\,22 & 22,\,21 & 23,\,22 & 25,\,24 & 23,\,24 & 23,\,24 \\
            & \ourmethod{} (diff) & 23,\,22 & 35,\,34 & 22,\,21 & 30,\,23 & 22,\,21 & 22,\,23 \\
            & \ourmethod{} ($\ell_2$) & 35,\,34 & 35,\,34 & 35,\,34 & 35,\,34 & 35,\,34 & 35,\,26 \\
        \midrule
        \multirow{6}{*}{$3$}
            & D-only & 21,\,22,\,23 & 22,\,21,\,23 & 22,\,23,\,21 & 20,\,19,\,21 & 23,\,22,\,21 & 22,\,23,\,24 \\
            & C-only & 26,\,21,\,25 & 23,\,24,\,25 & 34,\,33,\,35 & 25,\,34,\,33 & 34,\,35,\,33 & 25,\,23,\,26 \\
            & Random Selection & 12,\,13,\,20 & 14,\,17,\,21 & 16,\,18,\,22 & 15,\,17,\,23 & 12,\,14,\,17 & 13,\,15,\,22 \\
            & LayerNavigator & 21,\,22,\,23 & 22,\,21,\,23 & 23,\,22,\,24 & 25,\,24,\,26 & 23,\,24,\,22 & 23,\,24,\,22 \\
            & \ourmethod{} (diff) & 23,\,22,\,21 & 35,\,34,\,33 & 22,\,21,\,23 & 30,\,23,\,35 & 22,\,21,\,23 & 22,\,23,\,21 \\
            & \ourmethod{} ($\ell_2$) & 35,\,34,\,33 & 35,\,34,\,33 & 35,\,34,\,30 & 35,\,34,\,33 & 35,\,34,\,33 & 35,\,26,\,30 \\
        \bottomrule
    \end{tabular}
    \caption{Layers selected by each strategy on Qwen-3B-Instruct (36 layers, indexed $0$--$35$), for every persona task and layer budget $K$. Layers are listed in selection order, highest-scoring first; for Random Selection they are the sampled set. \ourmethod{} is shown under both scoring clouds: the standardized difference cloud (diff) and its unit-sphere projection ($\ell_2$). All baselines concentrate in the middle of the stack, whereas \ourmethod{}~($\ell_2$) consistently selects the
    final layers.}
    \label{tab:layers_qwen3b}
\end{table*}

\section{Ablation Study with Different Topological Metrics}
\label{app:tda_ablation}
To assess how sensitive topology-guided layer selection is to its design choices, we sweep the space of descriptors used to rank layers and compare, for every model and behaviour, the best-performing configuration against LayerNavigator (Table~\ref{tab:best_tda_per_dataset}). The sweep covers (i) persistence statistics including mean, maximum and total persistence of $H_0$, $H_1$ and $H_2$, persistence entropy, bar counts and dominance ratios. Those metrics are computed on the union of
positive and negative activations, on the per-example displacement cloud, and on its $\ell_2$-normalised counterpart; (ii) composite HOLE-style scores (topological separability, purity, separability) and min--max-normalised combinations of the above; (iii) seven distances for the Vietoris--Rips filtration (Euclidean, cosine, Mahalanobis, geodesic, and density-normalised Euclidean, cosine and Mahalanobis); and (iv) steering budgets $k \in \{1, 2, 3, 5\}$. In every case, the mean-difference steering vector is added at unit
strength to the top-$k$ layers under the descriptor, and we report the probability of the behaviour-consistent answer on the test split. 
Selected post hoc as an oracle upper bound, the best topological configuration matches or exceeds LayerNavigator in 29 of the 30 model--behaviour pairs, raising the steered probability by 4.0 points on average, with the largest gains on Qwen2.5-3B, where it lifts Conscientiousness to $0.923$ while LayerNavigator falls below the unsteered model ($0.694 \rightarrow 0.610$), although no single descriptor or distance is best across all pairs.

\begin{table*}[t]
  \centering
  \footnotesize
  \caption{Best TDA layer-selection score for each model and dataset, searched over all
  scores, distance metrics and numbers of steered layers $k$. Values are the probability
  of the target-behaviour answer. Base is the unsteered model; LayerNavigator is shown at
  its own best $k$ (in brackets), measured against the same unsteered baseline.}
  \label{tab:best_tda_per_dataset}
  \resizebox{\linewidth}{!}{%
  \begin{tabular}{llllcccc}
    \toprule
    Model & Dataset & TDA score & Distance & $k$ & Base & TDA $\uparrow$ & LayerNavigator $\uparrow$ \\
    \midrule
    \multirow{6}{*}{\rotatebox[origin=c]{90}{\shortstack{Llama-3B}}} & Conscientiousness$^{*}$ & Mean pers.\ $H_0$ (diff) & Euclidean & 2 & 0.7251 & 0.8700 & \textbf{0.8788}  \\
     & Religion$^{*}$ & Mean pers.\ $H_0$ (diff) & Euclidean & 3 & 0.5568 & \textbf{0.8587} & 0.8565  \\
     & Self-Aware$^{*}$ & Mean pers.\ $H_0$ (diff) & Euclidean & 3 & 0.5977 & \textbf{0.7468} & 0.7460  \\
     & Self-Improvement$^{*}$ & Mean pers.\ $H_0$ (diff) & Euclidean & 2 & 0.7096 & \textbf{0.8199} & \textbf{0.8199} \\
     & Alliance$^{*}$ & Mean pers.\ $H_0$ (union) & Euclidean & 2 & 0.7556 & \textbf{0.9000} & \textbf{0.9000}  \\
     & Impact$^{*}$ & Mean pers.\ $H_0$ (diff) & Euclidean & 3 & 0.6589 & \textbf{0.7378} & 0.7346 \\
    \midrule
    \multirow{6}{*}{\rotatebox[origin=c]{90}{\shortstack{Llama-8B}}} & Conscientiousness & Total pers.\ $H_0$ (union) & Geodesic & 3 & 0.8207 & \textbf{0.9419} & 0.9383 \\
     & Religion & TSS with $\beta_1$-AUC penalty & Cosine & 5 & 0.6893 & \textbf{0.8803} & 0.8405  \\
     & Self-Aware & Max $H_0$ (union) $+$ inv.\ max $H_0$ (diff) & Euclidean & 5 & 0.8186 & \textbf{0.9304} & 0.9160  \\
     & Self-Improvement & Purity $+$ separability & Mahalanobis & 5 & 0.7417 & \textbf{0.9160} & 0.8628  \\
     & Alliance & $H_1$ dominance (max/mean pers.) & Geodesic & 5 & 0.8123 & \textbf{0.9455} & 0.9369  \\
     & Impact & Max pers.\ $H_1$ (union) & Euclidean & 5 & 0.7745 & \textbf{0.8720} & 0.8454  \\
    \midrule
    \multirow{6}{*}{\rotatebox[origin=c]{90}{\shortstack{Qwen-3B}}} & Conscientiousness & Mean pers.\ $H_1$ (union) & Euclidean & 5 & 0.6940 & \textbf{0.9229} & 0.6100  \\
     & Religion & Mean pers.\ $H_1$ (union) & Euclidean & 5 & 0.6004 & \textbf{0.7974} & 0.5475  \\
     & Self-Aware & Total pers.\ $H_1$ (union) & Euclidean & 5 & 0.5702 & \textbf{0.9155} & 0.8311  \\
     & Self-Improvement & Total pers.\ $H_0{+}H_1{+}H_2$ (union) & Cosine & 3 & 0.6272 & \textbf{0.8697} & 0.8244  \\
     & Alliance  & Inverse strong-loop count ($H_1$) & Mahalanobis & 5 & 0.7552 & \textbf{0.9019} & 0.8837  \\
     & Impact & Mean pers.\ $H_1$ (union) & Euclidean & 5 & 0.6060 & \textbf{0.7637} & 0.6502  \\
    \midrule
    \multirow{6}{*}{\rotatebox[origin=c]{90}{\shortstack{Qwen-7B}}} & Conscientiousness & Mean $H_0$ $\times$ $H_1$ fragility & Euclidean & 5 & 0.8785 & \textbf{0.9435} & 0.9315  \\
     & Religion & $H_0$ dominance (max/mean pers.) & Geodesic & 5 & 0.8445 & \textbf{0.8709} & 0.8574  \\
     & Self-Aware & $H_1$ bar count & Mahalanobis & 5 & 0.9435 & \textbf{0.9639} & 0.9638  \\
     & Self-Improvement & Max pers.\ $H_1$ (union) & Dens.-norm.\ Euclidean & 5 & 0.8352 & \textbf{0.8752} & 0.8723  \\
     & Alliance  & $H_1$ dominance (max/mean pers.) & Geodesic & 3 & 0.8705 & \textbf{0.9539} & 0.9114  \\
     & Impact & Inverse $H_1$ entropy & Dens.-norm.\ cosine & 5 & 0.7334 & \textbf{0.8153} & 0.8077  \\
    \midrule
    \multirow{6}{*}{\rotatebox[origin=c]{90}{\shortstack{Qwen-32B}}} & Conscientiousness$^{*}$ & Mean pers.\ $H_1$ (diff-$\ell_2$) & Euclidean & 5 & 0.9557 & \textbf{0.9685} & 0.9594  \\
     & Religion $^{*}$ & Max $H_0$ (union) $+$ inv.\ max $H_0$ (diff) & Euclidean & 5 & 0.6077 & \textbf{0.7292} & 0.6920  \\
     & Self-Aware & Mean pers.\ $H_0$ (union) & Euclidean & 5 & 0.9789 & \textbf{0.9852} & 0.9804  \\
     & Self-Improvement$^{*}$ & Max $H_0$ (union) $+$ inv.\ max $H_0$ (diff) & Euclidean & 5 & 0.8259 & \textbf{0.8564} & 0.8343  \\
     & Alliance  & Mean pers.\ $H_0$ (union) & Mahalanobis & 5 & 0.9198 & \textbf{0.9631} & 0.9242  \\
     & Impact $^{*}$ & Mean pers.\ $H_0$ (diff) & Euclidean & 5 & 0.6728 & \textbf{0.7292} & 0.6771  \\
    \bottomrule
  \end{tabular}}
  \par\smallskip
  \parbox{\linewidth}{\footnotesize
  union: persistence of the pooled positive and negative activations; diff: persistence of the
  per-example (positive $-$ negative) displacements; inv.: min--max normalised and inverted;
  $+$: sum of min--max normalised scores; TSS: HOLE topological separability score.
  $^{*}$Only Euclidean distance was run for this model and dataset.
  Where several scores select the same layers, one is shown.}
\end{table*}

\section{Additional Analysis}
\label{mmlu_utility}

Table~\ref{tab:mmlu_utility_k12} reports detailed MMLU utility results for
$K\in\{1,2\}$, while Table~\ref{tab:mmlu_utility_k3} reports the corresponding
results for $K=3$.
\begin{table*}[t]
    \centering
    \setlength{\aboverulesep}{0pt}
    \setlength{\belowrulesep}{0pt}
    \setlength{\extrarowheight}{.75ex}
    \renewcommand{\tabcolsep}{1pt}
    \footnotesize
    \begin{tabular}{l c ccc ccc}
        \toprule
        &
        &
        \multicolumn{3}{c}{$K=1$} &
        \multicolumn{3}{c}{$K=2$}
        \\
        \cmidrule(lr){3-5}
        \cmidrule(lr){6-8}

        Model & Base
        & LN & Ours & Ours$_{\ell_2}$
        & LN & Ours & Ours$_{\ell_2}$
        \\
        \midrule

        Llama-3B
        & 59.56
        & 60.00 {\scriptsize$(+0.44)$}
        & 60.00 {\scriptsize$(+0.44)$}
        & 60.00 {\scriptsize$(+0.44)$}
        & 58.86 {\scriptsize$(-0.70)$}
        & 58.86 {\scriptsize$(-0.70)$}
        & 58.86 {\scriptsize$(-0.70)$}
        \\

        Llama-8B
        & 67.89
        & 67.46 {\scriptsize$(-0.43)$}
        & 67.37 {\scriptsize$(-0.52)$}
        & 67.37 {\scriptsize$(-0.52)$}
        & 66.75 {\scriptsize$(-1.14)$}
        & 67.81 {\scriptsize$(-0.08)$}
        & 66.23 {\scriptsize$(-1.66)$}
        \\

        Qwen-3B
        & 67.19
        & 67.21 {\scriptsize$(+0.02)$}
        & 66.93 {\scriptsize$(-0.26)$}
        & 67.28 {\scriptsize$(+0.09)$}
        & 66.84 {\scriptsize$(-0.35)$}
        & 66.75 {\scriptsize$(-0.44)$}
        & 67.54 {\scriptsize$(+0.35)$}
        \\

        Qwen-7B
        & 74.39
        & 73.95 {\scriptsize$(-0.44)$}
        & 74.21 {\scriptsize$(-0.18)$}
        & 74.39 {\scriptsize$(+0.00)$}
        & 73.68 {\scriptsize$(-0.71)$}
        & 74.56 {\scriptsize$(+0.17)$}
        & 74.30 {\scriptsize$(-0.09)$}
        \\

        Qwen-32B
        & 83.86
        & 83.60 {\scriptsize$(-0.26)$}
        & 83.95 {\scriptsize$(+0.09)$}
        & 83.95 {\scriptsize$(+0.09)$}
        & 83.77 {\scriptsize$(-0.09)$}
        & 83.60 {\scriptsize$(-0.26)$}
        & 83.60 {\scriptsize$(-0.26)$}
        \\

        \bottomrule
    \end{tabular}

    \caption{
        MMLU accuracy (\%, $\uparrow$) under simultaneous persona steering for
        $K\in\{1,2\}$. Each entry reports 5-shot MMLU accuracy, with the absolute
        change from the unsteered Base model in parentheses (percentage points).
        Models are steered on \texttt{conscientiousness} at $\alpha=1.0$.
    }
    \label{tab:mmlu_utility_k12}
\end{table*}

\begin{table*}[t]
    \centering
    \setlength{\aboverulesep}{0pt}
    \setlength{\belowrulesep}{0pt}
    \setlength{\extrarowheight}{.75ex}
    \renewcommand{\tabcolsep}{4pt}
    \footnotesize
    \begin{tabular}{l c ccc}
        \toprule
        &
        &
        \multicolumn{3}{c}{$K=3$}
        \\
        \cmidrule(lr){3-5}

        Model & Base
        & LN & Ours & Ours$_{\ell_2}$
        \\
        \midrule

        Llama-3B
        & 59.56
        & 56.84 {\scriptsize$(-2.72)$}
        & 56.49 {\scriptsize$(-3.07)$}
        & 56.49 {\scriptsize$(-3.07)$}
        \\

        Llama-8B
        & 67.89
        & 66.14 {\scriptsize$(-1.75)$}
        & 66.58 {\scriptsize$(-1.31)$}
        & 66.14 {\scriptsize$(-1.75)$}
        \\

        Qwen-3B
        & 67.19
        & 67.02 {\scriptsize$(-0.17)$}
        & 67.02 {\scriptsize$(-0.17)$}
        & 67.19 {\scriptsize$(+0.00)$}
        \\

        Qwen-7B
        & 74.39
        & 73.60 {\scriptsize$(-0.79)$}
        & 74.47 {\scriptsize$(+0.08)$}
        & 74.39 {\scriptsize$(+0.00)$}
        \\

        Qwen-32B
        & 83.86
        & 83.25 {\scriptsize$(-0.61)$}
        & 83.77 {\scriptsize$(-0.09)$}
        & 83.77 {\scriptsize$(-0.09)$}
        \\

        \bottomrule
    \end{tabular}

    \caption{
        MMLU accuracy (\%, $\uparrow$) under simultaneous persona steering for
        $K=3$. Parenthetical values denote the absolute change from the unsteered
        Base model in percentage points. Models are steered on
        \texttt{conscientiousness} at $\alpha=1.0$.
    }
    \label{tab:mmlu_utility_k3}
\end{table*}

\section{Detail Performance Evaluation with Evaluation-Based Selection Method}
\label{evaluation_based_selection_method}

\paragraph{Per-instance baseline (W2S-Multi).}
We compare TopoSteer against W2S-Multi, the deployable per-instance multi-layer steering method of \citet{nelwan2026deployable}. The steering setup, prompts, data splits, and metrics are exactly those used for \ourmethod{} and LayerNavigator: the contrastive (CAA) steering vectors are extracted from the 700 training pairs of each task, a steering vector is added with $\alpha{=}1$ to the last-token residual stream at every selected layer, and alignment (Prob) and perplexity (PPL) are measured on the same 100 test prompts per task. In addition, W2S-Multi chooses layers per input rather than per task. On the 200 validation prompts of each task, we first compute the single-layer effect of every layer: the change in the restricted Yes/No probability of the gold answer when that layer alone is steered. This costs $(L{+}1)$ forward passes per prompt (one unsteered pass plus one per layer) and requires the gold labels. Each prompt is then embedded with a sentence encoder (\texttt{bge-base-en-v1.5} \citep{10.1145/3626772.3657878}) and reduced to 25 dimensions by PCA fitted on the validation split. A one-hidden-layer MLP ($25{\to}64{\to}L$, ReLU, dropout 0.3) is trained to reproduce the distribution of single-layer effects over layers, using a listwise KL loss with the effects expressed in percentage points, optimized with full-batch Adam (learning rate $10^{-3}$, weight decay $10^{-4}$, 300 epochs).
At test time, each input is steered at the $K$ layers ranked highest by the MLP, so different inputs can receive different layer sets; we report $K\in\{1,2,3\}$ to match \ourmethod.

\paragraph{Comparison with Evaluation-Based Selection.}
Table~\ref{tab:toposteer_vs_w2s_avg} reports the detailed comparison with the
per-instance W2S-Multi baseline. W2S-Multi achieves higher alignment probability on average ($81.25\%$ vs.\ $78.90\%$), but requires substantially more computation, with a total runtime of $15.9$ minutes compared with $8.4$ minutes for \ourmethod{}. This difference reflects the additional inference and label-dependent training required by W2S-Multi, whereas \ourmethod{}
computes layer scores directly from the extraction data without additional steering evaluations. The full model-level results show the same trade-off across model families, with W2S-Multi generally achieving higher alignment at a higher perplexity and computational cost.

\section{Comparing Connectivity Length vs.\ Sum of Distance to the Mean\label{sec_mst_vs_total_dist}}

A natural way to ask ``how good is the mean as a representative of the individual points'' is to measure the typical distance from a point to some reference --- either the mean itself, or, as a proxy for overall compactness, the connectivity length $\cL$ captured by the minimum spanning tree. Intuitively, a single mean should represent a unimodal cloud reasonably well (points are dispersed around one center) but should represent a bimodal cloud poorly (no single point, including the mean, sits close to both modes). A good quality measure should therefore be \emph{sensitive to modality}, scoring the unimodal case as better represented and the bimodal case as worse represented.

The relevant standardization here is the one \ourmethod{} actually uses: the pooled $2N$ activations $\{a_i^{+}\}\cup\{a_i^{-}\}$ are standardized once, coordinate-wise, using a single mean and std computed over all $2N$ points (Sec.~\ref{sec:setup}, Eq.~\eqref{eq:zscore}{; Proposition~\ref{prop:sepscale} records what that standardization couples the score to}; this is also exactly what LayerNavigator's own scoring code performs on \texttt{all\_acts} before differencing). The displacement cloud $\Delta=\{\delta_i\}$ is formed only afterward, $\delta_i=\tilde a_i^{+}-\tilde a_i^{-}$, with no further rescaling. This standardization scale is therefore driven by class separation and any pair-shared noise that cancels in the difference but not in the individual activations --- quantities the displacement cloud itself cannot see, and this shapes which of the two candidate statistics below remain reliable.

\subsection{Model}

We extend the homothety family of Sec.~\ref{sec:ln} {(Appendix~\ref{app:ln})} to allow the residual shape to be unimodal or bimodal, dropping the layer subscript $l$ as in that example:
\[
  a_i^{+} = \tfrac{\mu_\delta}{2} + \sigma\left(\tfrac{\xi_i}{2}+c_i\right), \qquad
  a_i^{-} = -\tfrac{\mu_\delta}{2} + \sigma\left(-\tfrac{\xi_i}{2}+c_i\right),
\]
so that $\delta_i = a_i^{+} - a_i^{-} = \mu_\delta + \sigma\xi_i$ exactly, with the pair-shared noise $c_i$ canceling as before. Isotropically:
\begin{itemize}
  \item $\mu_\delta = (\mu_0/\sqrt D)\mathbf 1$, so $\|\mu_\delta\|=\mu_0$ (class separation, spread evenly across coordinates);
  \item $c_i$ i.i.d.\ uniform on the sphere of radius $\gamma$ (shared pair-level noise);
  \item $\xi_i$ \textbf{unimodal}: i.i.d.\ uniform on the sphere of radius $R_u$; or \textbf{bimodal}: $N/2$ points uniform on the sphere of radius $R_w$ centered at $c_\xi=(\mu_\xi/\sqrt D)\mathbf1$, and $N/2$ centered at $-c_\xi$, the mode-separation spread evenly across coordinates.
\end{itemize}
Write $V_\xi := R_u^2$ (unimodal) or $V_\xi := R_w^2+\mu_\xi^2$ (bimodal, by the usual law-of-total-variance decomposition).

\subsection{Lemma: sum of norms vs.\ connectivity length}

\begin{lemma}[Sum of norms vs.\ connectivity length under pooled-activation standardization]
\label{lem:main}
Under the model above, as $D\to\infty$ with $N$ fixed, let $\Delta$ denote the displacement cloud obtained by pooled standardization of the $2N$ activations followed by differencing, and let
\[
  \tau := \sigma^2\!\left(\frac{V_\xi}{4}+\gamma^2\right) + \frac{\mu_0^2}{4}.
\]

\textbf{Part A (sum of norms).} For either shape,
\[
  \mathrm{SN}(\Delta) := \sum_i\|\delta_i-\bar\delta\| = N\sigma\sqrt{V_\xi}\,\sqrt{D/\tau} + o(\sqrt D),
\]
so
\[
  \frac{\mathrm{SN}(\Delta_{\mathrm{bi}})}{\mathrm{SN}(\Delta_{\mathrm{uni}})} \;\to\; \sqrt{\frac{V_{\xi,\mathrm{bi}}}{V_{\xi,\mathrm{uni}}}}\cdot\sqrt{\frac{\tau_{\mathrm{uni}}}{\tau_{\mathrm{bi}}}}.
\]

\textbf{Part B (connectivity length).}
\[
  \cL(\Delta_{\mathrm{uni}}) = (N-1)\sigma\sqrt2\,R_u\sqrt{D/\tau_{\mathrm{uni}}} + o(\sqrt D),
\]
\[
  \cL(\Delta_{\mathrm{bi}}) = \left[(N-2)\sigma\sqrt2\,R_w + \sigma\sqrt{4\mu_\xi^2+2R_w^2}\right]\sqrt{D/\tau_{\mathrm{bi}}} + o(\sqrt D),
\]
so
\[
  \frac{\cL(\Delta_{\mathrm{bi}})}{\cL(\Delta_{\mathrm{uni}})} \;\to\;
  \underbrace{\frac{(N-2)R_w+\sqrt{2\mu_\xi^2+R_w^2}}{(N-1)R_u}}_{\text{raw shape factor}}
  \times
  \underbrace{\sqrt{\tau_{\mathrm{uni}}/\tau_{\mathrm{bi}}}}_{\text{standardization dilution}}.
\]

\textbf{Part C (bridge share).} Algorithm~\ref{alg:toposteer} reports, alongside $\cL$, the bridge share $b_l := \max_{e\in\MST(\Delta)}w(e)/\cL(\Delta)$ (Eq.~\eqref{eq:bridge}). Being a ratio of a single edge to the total, $b_l$ is invariant to any positive scalar multiplying the whole cloud, hence to $\tau$ entirely:
\[
  b(\Delta_{\mathrm{uni}}) \to \frac{1}{N-1}, \qquad
  b(\Delta_{\mathrm{bi}}) \to \frac{\sqrt{2\mu_\xi^2+R_w^2}}{(N-2)R_w+\sqrt{2\mu_\xi^2+R_w^2}},
\]
regardless of $\mu_0,\gamma,\sigma$, with $b(\Delta_{\mathrm{bi}})>b(\Delta_{\mathrm{uni}})$ whenever $\mu_\xi>0$.
\end{lemma}

\subsection{Proof of Lemma~\ref{lem:main}}

\textbf{Step 0 (translation invariance, exact).} By Proposition~\ref{prop:inv}, $\cL(\Delta)=\cL(\Delta-v^{\md})$ for any $D,N$, with no asymptotics needed. Centering $\Delta$ at its own empirical mean for Part A removes the same constant $v^{\md}=\mu_\delta$ in the standardized scale. So for both statistics, the effect of $\mu_\delta$ is confined entirely to how it inflates the standardization scale $\tau$ below --- never to a residual shift of the centered cloud.

\textbf{Step 1 (raw, unstandardized distances concentrate to constants).} For $\xi_i,\xi_j$ i.i.d.\ uniform on a sphere of radius $R$ in $\R^D$,
\[
  \|\xi_i-\xi_j\|^2 = 2R^2-2\langle\xi_i,\xi_j\rangle.
\]
Writing $\xi_i=R\,u_i$ for $u_i$ uniform on the unit sphere, each coordinate of $u_i$ behaves for large $D$ like an independent $N(0,1/D)$ variable, so $\langle u_i,u_j\rangle=\sum_k u_{i,k}u_{j,k}$ is a sum of $D$ independent mean-zero terms of variance $O(1/D^2)$ each, giving $\langle u_i,u_j\rangle=O_p(1/\sqrt D)\to0$. Hence $\|\xi_i-\xi_j\|\xrightarrow{p}\sqrt2\,R$ --- and this holds for \emph{every} pair $(i,j)$, not just in expectation, so in the limit \emph{all} pairwise distances within a radius-$R$ spherical cloud become equal to the same constant. The same computation, applied to $\xi_i=c_\xi+v_i$, $\xi_j=-c_\xi+v_j$ in different modes of the bimodal cloud, gives $\|\xi_i-\xi_j\|\xrightarrow{p}\sqrt{4\mu_\xi^2+2R_w^2}$; applied within one mode it gives $\sqrt2\,R_w$, as in the unimodal case.

\textbf{Step 2 (the pooled standardization is a single shared scalar).} The pooled set $\{a_i^+\}\cup\{a_i^-\}$ is, per coordinate $k$, a 50/50 mixture of the class-$+$ and class-$-$ distributions. By the law of total variance,
\[
  s_k^2 = \underbrace{\tfrac12\mathrm{Var}(a_{i,k}^+\mid{+}) + \tfrac12\mathrm{Var}(a_{i,k}^-\mid{-})}_{\text{within-class}} + \underbrace{\mathrm{Var}\bigl(\{\mu_{\delta,k}/2,-\mu_{\delta,k}/2\}\bigr)}_{\text{between-class}}.
\]
From $a_i^{+}=\mu_\delta/2+\sigma(\xi_i/2+c_i)$ and independence of $\xi_i,c_i$: within-class variance is $\sigma^2\bigl(\mathrm{Var}(\xi_k)/4+\mathrm{Var}(c_k)\bigr)$ (identical for class $-$ by the symmetric construction). The between-class term is the variance of a two-point $\pm a$ distribution, which is $a^2$; here $a=\mu_{\delta,k}/2$, giving $\mu_{\delta,k}^2/4$. So
\[
  s_k^2 = \sigma^2\left(\mathrm{Var}(\xi_k)+4\,\mathrm{Var}(c_k)\right)/4 + \mu_{\delta,k}^2/4.
\]
Substituting the isotropic per-coordinate values $\mathrm{Var}(\xi_k)=V_\xi/D$, $\mathrm{Var}(c_k)=\gamma^2/D$, $\mu_{\delta,k}^2=\mu_0^2/D$ gives $s_k^2=\tau/D$ for \emph{every} coordinate $k$ --- the same number regardless of $k$, because every quantity entering it ($V_\xi,\gamma,\mu_0$) was assumed spread evenly across coordinates. A coordinate-wise standardization that divides every coordinate by the same number is, by definition, a single global scalar division: $\delta_i-\bar\delta = (\mu_\delta+\sigma\xi_i)/s - \mu_\delta/s = \sigma\xi_i/s = \sigma\xi_i\sqrt{D/\tau}$.

\textbf{Step 3 (Part A).} For the unimodal cloud, $\|\xi_i\|=R_u$ exactly for every $i$ (points lie exactly on the sphere), so $\|\delta_i-\bar\delta\|=\sigma R_u\sqrt{D/\tau_{\mathrm{uni}}}$ exactly, and summing over $N$ points gives $N\sigma R_u\sqrt{D/\tau_{\mathrm{uni}}}$ --- matching the claimed formula since $R_u=\sqrt{V_\xi}$ in the unimodal case. For the bimodal cloud, $\|\xi_i\|^2=\mu_\xi^2+R_w^2\pm2\langle c_\xi,v_i\rangle$ with the cross term $O_p(\mu_\xi R_w/\sqrt D)\to0$ by the Step~1 argument, so $\|\xi_i\|\to\sqrt{\mu_\xi^2+R_w^2}=\sqrt{V_\xi}$ for both modes; the $O_p(1)$ fluctuations have mean zero across the two equally sized modes and cancel in the sum, giving $N\sigma\sqrt{V_\xi}\sqrt{D/\tau_{\mathrm{bi}}}+o(\sqrt D)$.

\textbf{Step 4 (Part B).} Step~1 shows every pairwise distance within the unimodal cloud converges to the \emph{same} constant $\sqrt2\,\sigma R_u\sqrt{D/\tau_{\mathrm{uni}}}$ (after applying the Step~2 scalar); when all edges of a complete graph have equal weight, every spanning tree --- in particular the minimum one --- has the same total weight, namely $(N-1)$ times that constant. This gives the stated formula for $\cL(\Delta_{\mathrm{uni}})$. For the bimodal cloud, the same argument applies separately within each mode (giving $(N/2-1)$ edges of length $\sqrt2\,\sigma R_w\sqrt{D/\tau_{\mathrm{bi}}}$ per mode, $(N-2)$ in total across both modes) and across modes (every inter-mode pair converges to the same distance, so it does not matter which single edge the minimum spanning tree uses to join the two modes; exactly one such edge is needed to connect the forest into a tree). Summing the intra-mode and the one inter-mode edge gives the stated formula.

\textbf{Step 5 (Part C).} By Step~2, standardization multiplies every edge of $\MST(\Delta)$ by the same scalar $\sigma\sqrt{D/\tau}$, which therefore cancels exactly from the ratio $b_l=\max_e w(e)/\cL$ --- this ratio depends only on the \emph{relative} edge lengths established in Step~4. For the unimodal cloud all $(N-1)$ edges converge to the same length $e$, so the single longest edge is also length $e$ and $b(\Delta_{\mathrm{uni}})\to e/((N-1)e)=1/(N-1)$. For the bimodal cloud, by Step~4, $\cL(\Delta_{\mathrm{bi}})$ is the sum of $(N-2)$ bulk edges of length $\sqrt2\,R_w$ and one bridge edge of length $\sqrt{4\mu_\xi^2+2R_w^2}=\sqrt2\sqrt{2\mu_\xi^2+R_w^2}$, and since $\mu_\xi>0$ makes the bridge strictly the longest edge, $b(\Delta_{\mathrm{bi}})\to\sqrt2\sqrt{2\mu_\xi^2+R_w^2}\big/\bigl[(N-2)\sqrt2\,R_w+\sqrt2\sqrt{2\mu_\xi^2+R_w^2}\bigr]$, which simplifies to the stated expression after canceling the common factor of $\sqrt2$. $\blacksquare$

\subsection{Two regimes}

The comparison in the Lemma is abstract; the following two concrete cases make it operational, and in both cases we track sum of norms alongside connectivity length rather than one at a time. The two cases differ only in what is held fixed between the unimodal and bimodal $\xi$: total scatter $V_\xi$ in the first, the within-mode noise level $R_w$ in the second.

\textbf{Matched-scatter comparison} ($R_u=\sqrt{R_w^2+\mu_\xi^2}$, unimodal and bimodal $\xi$ built to have identical aggregate scatter): $\tau_{\mathrm{uni}}=\tau_{\mathrm{bi}}$ exactly, so the dilution factor is $1$ for \emph{both} statistics, and each ratio reduces to its own shape factor alone. For connectivity length this leaves $R_w/\sqrt{R_w^2+\mu_\xi^2}<1$ as $N\to\infty$: the bimodal cloud is correctly scored lower. For sum of norms, Part A's shape factor is $\sqrt{V_{\xi,\mathrm{bi}}/V_{\xi,\mathrm{uni}}}=1$ by construction (that is exactly what ``matched scatter'' means), so its ratio is exactly $1$ --- not wrong, but uninformative: it cannot distinguish the two clouds \emph{by construction}, regardless of how differently their MSTs are organized. This is unaffected by $\mu_0,\gamma$, since these cancel identically from $\tau_{\mathrm{uni}}$ and $\tau_{\mathrm{bi}}$ in either statistic once $V_\xi$ is matched.

\textbf{Matched-noise comparison} ($R_u=R_w$, unimodal and bimodal share the same within-mode spread, and bimodal simply adds extra between-mode separation on top): now $\tau_{\mathrm{bi}}>\tau_{\mathrm{uni}}$, so \emph{both} statistics' dilution factors drop below $1$, but the two shape factors respond to this in opposite ways. Connectivity length's shape factor is itself $>1$ for finite $N$ (with $N=40,R_w=1,\mu_\xi=1.5$: $[(N-2)+\kappa_{\mathrm{br}}]/(N-1)=(38+2.345)/39\approx1.034$, where $\kappa_{\mathrm{br}}:=\sqrt{1+2\mu_\xi^2/R_w^2}$ is the bridge-to-bulk edge-length ratio, since one bridge edge $\kappa_{\mathrm{br}}\approx2.35\times$ longer than typical outweighs saving one edge count out of 39), so the dilution factor is fighting against a shape factor that already points the wrong way; sum of norms' shape factor is $\sqrt{(R_w^2+\mu_\xi^2)/R_w^2}\approx1.803>1$, so its dilution factor is fighting against an even larger shape factor pointing the same wrong way. At $\mu_0=\gamma=0$, connectivity length's dilution factor is small enough ($\approx0.555$) to win, giving a product $\approx0.573<1$ --- correctly ranking bimodal below unimodal, if only barely and by cancellation rather than by design. Sum of norms' dilution factor at $\mu_0=\gamma=0$ happens to invert to \emph{exactly} $1/1.803$ (pooled standardization degenerates to displacement-cloud self-standardization at zero nuisance), so its product is also exactly $1$ here --- coincidentally blind rather than coincidentally correct.

As $\mu_0$ (class separation) or $\gamma$ (shared pair-noise) grow, both dilution factors rise toward $1$, and each statistic's ratio drifts toward its own fixed shape factor: connectivity length's ratio \textbf{crosses above 1} near $\mu_0\approx5$ in our simulation, meaning it now ranks the mechanistically-split (bimodal) cloud \emph{above} the coherent one; sum of norms' ratio climbs toward $\approx1.80$, moving even further from useful. So in the matched-noise comparison the two statistics fail in the \emph{same direction} (both drift toward, and connectivity length crosses, the wrong ranking). For the \emph{same reason} (dilution washing out as separation dominates $\tau$) --- connectivity length simply starts from a better position (a shape factor close to $1$) and takes longer to fail. In comparison, sum of norms starts from a worse position (a shape factor of $1.8$) and is wrong for nearly the entire sweep. Neither statistic's failure here is a coincidence of this specific $R_w,\mu_\xi$ choice: both trace back to the same dilution mechanism identified in Step~2 of the proof, applied through two different shape factors. This identifies a risk to check for at well-separated layers specifically --- it is a claim about the standardization pipeline, not a diagnosis of any particular row of Table~\ref{tab:alignment_layer_selection}, since that table does not report the separation, bridge share, or $\cL$ values needed to confirm which mechanism, if any, is responsible for a given failure.

Bridge share $b_l$, by contrast, is unaffected by any of this in either regime: it depends only on $\tau_{\mathrm{uni}}=\tau_{\mathrm{bi}}$ or $\tau_{\mathrm{uni}}<\tau_{\mathrm{bi}}$ through a factor that cancels identically (Step~5), so it stays fixed at its two predicted values, $1/(N-1)\approx0.026$ for unimodal and $\approx0.058$ for bimodal (with $N=40, R_w=1, \mu_\xi=1.5$), regardless of $\mu_0,\gamma$, or which comparison is used.

\subsection{Simulation}

We simulate the model above at $D=500$, $N=40$, $R_w=1$, $\mu_\xi=1.5$, $\sigma=1$, $\gamma=1$ (Prim's algorithm MST, mode-separation spread isotropically across all $D$ coordinates, averaged over 8 trials per point), sweeping the class separation $\mu_0$ from 0 to 12, under both the matched-scatter and matched-noise comparisons.

\begin{figure}[h]
\centering
\includegraphics[width=\textwidth]{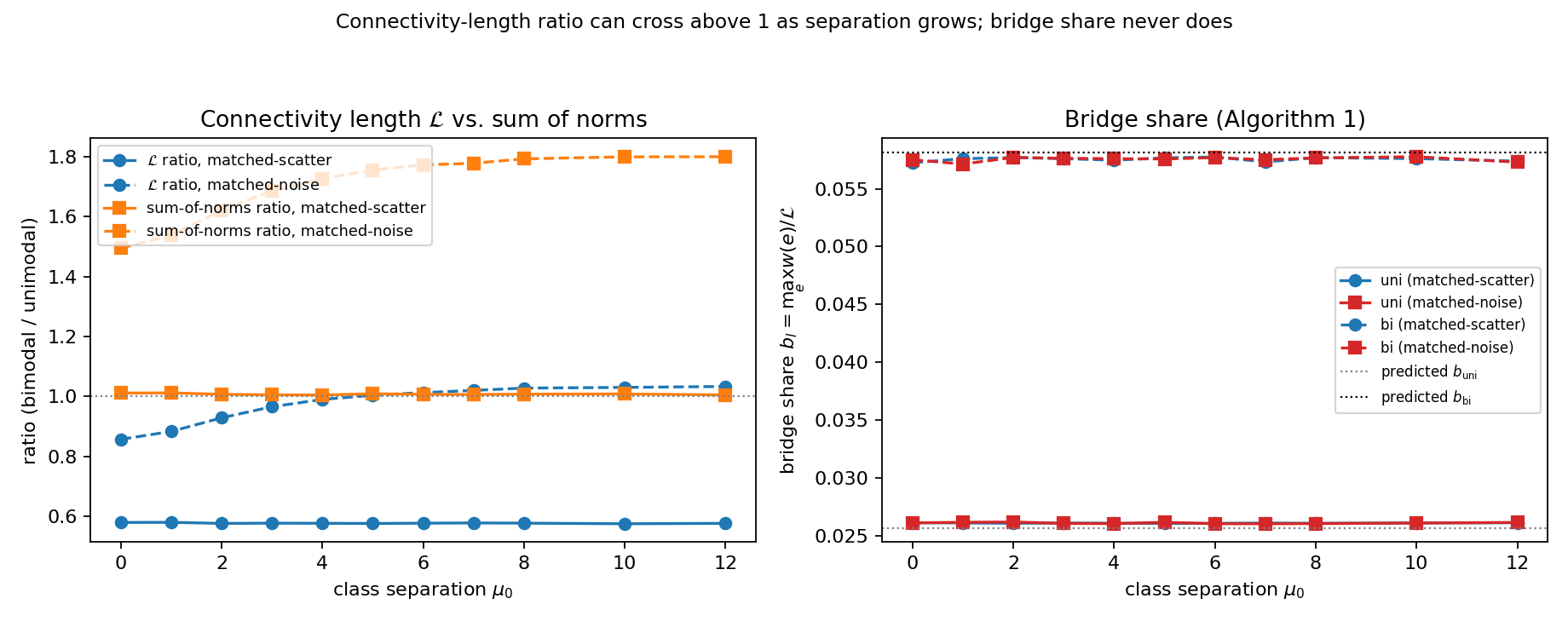}
\caption{Left: connectivity-length ratio $\cL(\Delta_{\mathrm{bi}})/\cL(\Delta_{\mathrm{uni}})$ and sum-of-norms ratio, overlaid for direct comparison. The connectivity-length ratio (blue) is stable at $\approx0.58$ under the matched-scatter comparison (solid) across all $\mu_0$, but under the matched-noise comparison (dashed) it rises with $\mu_0$ and crosses above~1 near $\mu_0\approx5$, meaning the bimodal (mechanistically split) cloud is ranked \emph{above} the unimodal one once class separation is large enough. The sum-of-norms ratio (orange) stays near 1 under matched-scatter (solid) but climbs toward $\approx1.8$ under the matched-noise comparison (dashed), becoming strongly shape-sensitive for a reason unrelated to representativeness. Right: the bridge share $b_l=\max_e w(e)/\cL$ that Algorithm~\ref{alg:toposteer} already reports is essentially unchanged across every condition --- $\approx0.026$ for unimodal, $\approx0.058$ for bimodal, matching the closed-form predictions of Part~C exactly, regardless of $\mu_0$ or comparison regime.}
\label{fig:dilution}
\end{figure}

Table~\ref{tab:sim_snapshot} reports selected values from the sweep underlying Figure~\ref{fig:dilution}.

\begin{table}[h]
\footnotesize
\centering
\caption{Selected points from the $\mu_0$-sweep at $D=500$, $\gamma=1$. The connectivity-length ratio crosses above 1 in the matched-noise comparison as $\mu_0$ grows, while the matched-scatter ratio and the bridge share (both regimes) stay essentially fixed at their predicted values.}
\begin{tabular}{r|rr|rr|rr}
\toprule
$\mu_0$ & $\cL$ ratio (m.s.) & $\cL$ ratio (m.n.) & sn ratio (m.s.) & sn ratio (m.n.) & $b_{\mathrm{uni}}$ (m.n.) & $b_{\mathrm{bi}}$ (m.n.) \\
\midrule
0  & 0.581 & 0.857 & 1.012 & 1.494 & 0.0261 & 0.0575 \\
2  & 0.577 & 0.929 & 1.007 & 1.621 & 0.0262 & 0.0577 \\
5  & 0.577 & 1.004 & 1.009 & 1.755 & 0.0262 & 0.0576 \\
8  & 0.578 & 1.028 & 1.008 & 1.792 & 0.0261 & 0.0577 \\
12 & 0.578 & 1.034 & 1.005 & 1.799 & 0.0262 & 0.0573 \\
\bottomrule
\end{tabular}
\label{tab:sim_snapshot}
\end{table}

\subsection{Summary}

Sum-of-norms is uninformative or actively misleading in every regime tested: it either cannot distinguish a coherent displacement cloud from a mechanistically split one, or does so only by leaking in class separation and shared noise that have nothing to do with representativeness. Connectivity length $\cL$ is a genuine but conditional signal: it correctly favors the coherent layer when class separation is modest, but the same pooled standardization that \ourmethod{} relies on can let separation dominate the normalizing scale at well-separated layers, at which point $\cL$ can rank a mechanistically split displacement cloud \emph{above} a coherent one. Whether this occurs for any specific layer in Table~\ref{tab:alignment_layer_selection} is an open empirical question --- the table does not report the diagnostics (separation magnitude, bridge share) needed to check it, so we state this as a risk the theory identifies, not as an explanation already confirmed for any reported result. The bridge share $b_l$ that Algorithm~\ref{alg:toposteer} already computes is unaffected by this: being a ratio of a single edge to the total, it is invariant to whatever scale the standardization imposes, and it separates unimodal from bimodal clouds by a fixed, predictable margin in every condition tested here. The practical recommendation is therefore to rank layers by $\cL$ but to treat that ranking as valid only where $b_l$ confirms it --- not to read $\cL$ magnitude alone as sufficient evidence of steerability.

\end{document}